\documentclass{article}
\usepackage{iclr2027_conference,times}
\setcitestyle{yysep={,}}
\iclrfinalcopy

\newcommand{\R}{\mathbb{R}}
\newcommand{\E}{\mathbb{E}}

\usepackage[utf8]{inputenc}
\usepackage[T1]{fontenc}
\usepackage{url}
\usepackage{booktabs}
\usepackage{array}
\usepackage{longtable}
\usepackage{graphicx}
\usepackage{amsmath,amssymb,amsfonts}
\usepackage{microtype}
\usepackage{xcolor}
\usepackage{enumitem}
\usepackage[hidelinks]{hyperref}
\usepackage{multirow}
\usepackage{colortbl}
\hypersetup{
  pdftitle={WorldToken: Time-First Sequence Modeling for Robotic Imitation Learning},
  pdfauthor={Chunkai Yang; Andong Yang; Di Huang; Chao Gao; Guyue Zhou}
}



\title{WorldToken: Time-First Sequence Modeling for Robotic Imitation Learning}

\author{%
\makebox[\dimexpr\textwidth-2\tabcolsep\relax][c]{%
Chunkai Yang\textsuperscript{1,}\thanks{These authors contributed equally to this work.}\kern0.5em\quad
Andong Yang\textsuperscript{2,}\footnotemark[1]\kern0.5em\quad
Di Huang\textsuperscript{1,}\thanks{Corresponding authors.}\kern0.5em\quad
Chao Gao\textsuperscript{2,}\footnotemark[2]\kern0.5em\quad
Guyue Zhou\textsuperscript{2}}\\[0.5ex]
\makebox[\dimexpr\textwidth-2\tabcolsep\relax][c]{%
\textsuperscript{1}Wuhan University\qquad
\textsuperscript{2}Tsinghua University}%
}

\graphicspath{{figures/}}

\begin{document}

\maketitle
\fancyhead{}
\renewcommand{\headrulewidth}{0pt}
\setcounter{footnote}{0}

\begin{abstract}
Vision-language-action policies inherit both capabilities and input representations from pretrained vision-language models, showing great potential for robotic manipulation across diverse industrial settings. As these policies increasingly use interaction history, organizing the representation of historical observations determines how experiences enter temporal context and how relationships across time are modeled, which is a generally ignored challenge in previous works. In this work, we believe solving this challenge requires an architectural reconstruction and propose \textbf{WorldToken}. WorldToken encodes each timestep's observations into one world token, processes the resulting history with a causal Transformer, and generates action chunks with a diffusion action head. Unified token enables long horizon tasks while relieving the memory requirement of the temporal backbone, hence improving performance. This design also offers high interpretability and allows advances in language modeling, such as pre-training and scaling, to be transferred to robot interaction policies. In RoboCasa experiments, WorldToken successfully handles most tasks with 85M parameters and achieves 59.4\% mean closed-loop success close to $\pi_{0.5}$ with 3.35B parameters. On the memory benchmark RMBench Blocks Ranking, WorldToken can reach the context of two minutes and achieves a success rate of 95\%. In addition, WorldToken’s holdout action RMSE is well described by power-law fits, and its closed-loop success rate improves consistently with increasing training data size in a study involving approximately 350,000 closed-loop evaluation episodes across 50 trained policies on RoboCasa, showing its scaling potential. We also conduct experiments to analyze information preservation and history use in WorldToken, providing empirical grounding for future work.

Code, model checkpoints, experiment records, and demonstration videos
are available at {\urlstyle{same}\url{https://me271828.github.io/WorldToken-project-page/}}.
\end{abstract}

\section{Introduction}
\label{sec:intro}

Vision-language-action models (VLA) inherit both capabilities and input representations from pretrained vision-language models (VLM) \citep{brohan2023rt2,kim2025openvla,black2025pi0}, showing great potential in applications like manipulation. Early VLMs such as LLaVA used an input image as context for understanding and generating text \citep{liu2023visual}. Inherently, VLAs predict actions from the current observation and perform effectively on a range of manipulation tasks \citep{kim2025openvla}. However, as tasks become complex, policies have increasingly incorporated interaction history, and observations must also provide context for decisions across time \citep{koo2026hamlet,torne2026mem}. The memory mechanism, which summarizes historical information with memory modules, extends the context of VLAs and gains popularity \citep{koo2026hamlet}. However, these memory modules generally retain the representational structure of VLMs, which can entangle within-timestep multimodal processing with cross-timestep history modeling. This makes it less straightforward to organize robot interaction histories around a unified sequence interface analogous to that used in language modeling. This raises a question: should the representational granularity inherited from vision-language models be carried over to history-conditioned robot policies?

\begin{figure}[t]
\centering
\includegraphics[width=\linewidth]{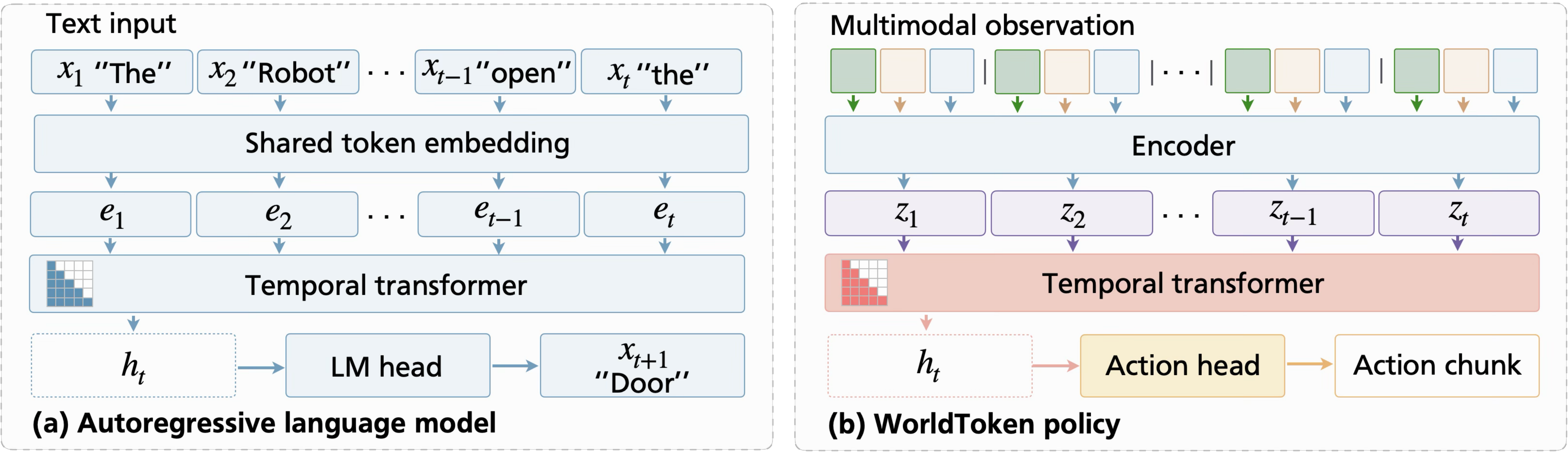}
\caption{Structure of a language model and WorldToken. (a) An LM head maps the language model's contextual representation to a distribution over the next text token, which is appended to the sequence. (b) WorldToken encodes each policy timestep's multimodal observations into one world token and uses a diffusion action head to generate an action chunk from the history-conditioned state. Executed actions affect the environment, whose next observation supplies the next world token.}
\label{fig:analogy}
\end{figure}

To answer this question, we propose WorldToken, an architecture-level way to reconsider the organization of historical observations for VLAs. WorldToken formulates a time-first organization, separating multimodal encoding within each timestep from history modeling across timesteps. As shown in Figure~\ref{fig:analogy}, an encoder encodes each policy timestep's multimodal observations into one learned world token and processes the resulting history with a causal Transformer. An LM head converts the language model's contextual representation into a distribution over text tokens. A DiT action head plays the corresponding role by generating an action chunk from the history-conditioned state. The encoder, temporal backbone, and action head have clear primary roles in perception, history modeling, and action generation. This design relieves the study of time relations for the temporal backbone, allowing VLA models to benefit from LLM-style scaling and pretraining without modification. Also, the design makes it easier to study component changes and interactions.

We evaluate WorldToken through extensive experiments. Across over 400,000 closed-loop evaluation episodes with more than 60 trained policies on RoboCasa \citep{nasiriany2024robocasa}, WorldToken supports effective control with an 85M model size. Experiments across data and model sizes show consistent gains from additional demonstrations, showing the potential for scaling. We also vary the number of tokens per timestep and the available history to examine their effects on control performance and the computation required for temporal modeling to analyze if the information compression sacrifices performance. On RMBench Blocks Ranking \citep{chen2026rmbench}, WorldToken reaches a context of two minutes, showing long horizon context ability. These results demonstrate that WorldToken can efficiently tokenize historical observations and also support policy timesteps as a useful basis for robot sequence modeling.

\section{WorldToken policy architecture}
\label{sec:method}

\begin{figure}[t]
\centering
\includegraphics[width=\linewidth]{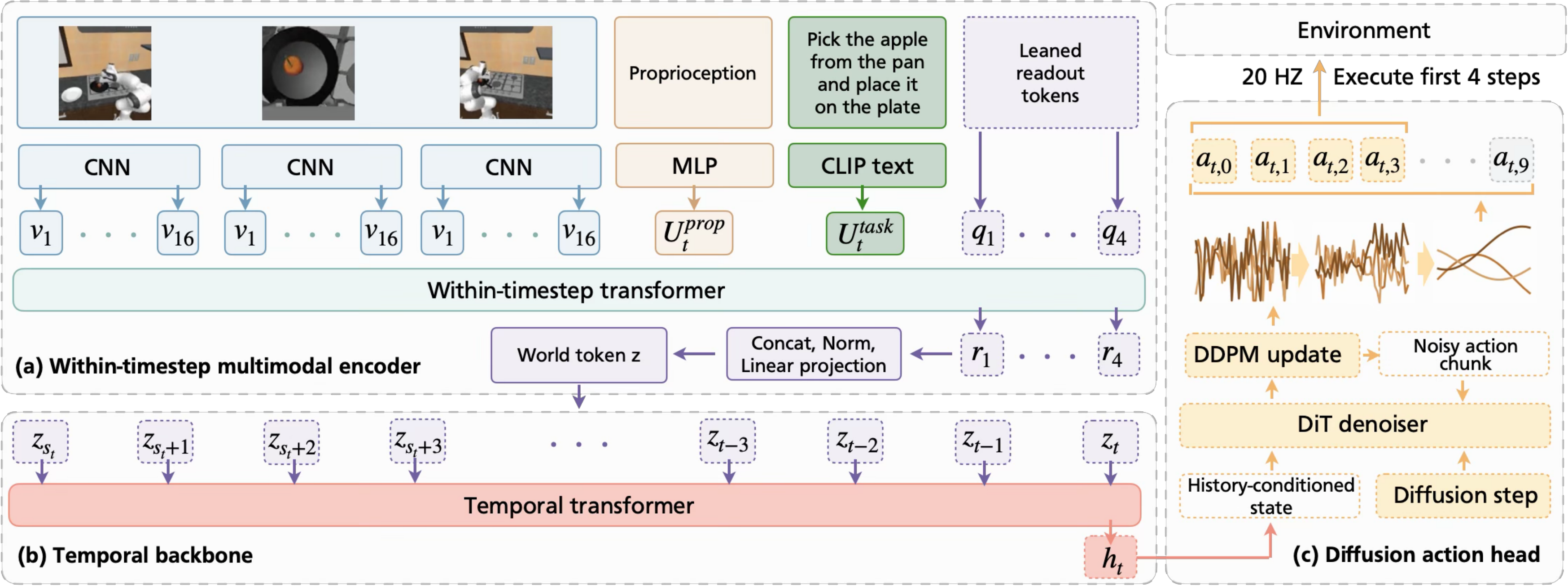}
\caption{WorldToken architecture. The encoder maps each observation to one world token.}
\label{fig:architecture}
\end{figure}

We study how to organize multimodal observation histories for sequence modeling. We define a policy timestep as one cycle in which the robot observes, generates an action chunk, and executes part of it before observing again. We use this cycle as the basic sequence unit, separating multimodal encoding within each timestep from history modeling across timesteps. This design leads to three modules (Figure~\ref{fig:architecture}): a within-timestep encoder $E_\theta$, a temporal backbone $T_\phi$, and a DiT action head $\mathcal{D}_\psi$~\citep{peebles2023dit}. We factorize the policy as
\begin{equation}
z_t = E_\theta(o_t), \qquad
h_t = \left[T_\phi(z_{\le t})\right]_{\mathrm{last}}, \qquad
\hat A_t \sim \mathcal{D}_\psi(\cdot\mid h_t).
\label{eq:method-decomp}
\end{equation}
At each policy timestep $t$, WorldToken takes the current observation $o_t$ as input and outputs an action chunk $\hat A_t$. Previously encoded world tokens can be reused without re-encoding past observations.

\paragraph{Within-timestep multimodal encoding.}
\label{sec:method-encoder}

The within-timestep encoder $E_\theta$ takes the multimodal observation $o_t$ as input and produces one world token $z_t$. We define
\begin{equation}
o_t=\{I_t^{1:K_{\mathrm{view}}},p_t,c\},
\label{eq:method-observation}
\end{equation}
where $I_t^{1:K_{\mathrm{view}}}$ denotes the images from $K_{\mathrm{view}}$ cameras, $p_t$ is proprioception, and $c$ is the task condition. A CNN visual stem maps each camera image $I_t^i$ to visual tokens $V_t^i$. Proprioception and the task condition are projected into tokens $u_t^{\mathrm{prop}}$ and $u_t^{\mathrm{task}}$. The \emph{observation tokens} are:
\begin{equation}
X_t=\bigl[V_t^1;\ldots;V_t^{K_{\mathrm{view}}};u_t^{\mathrm{prop}};u_t^{\mathrm{task}}\bigr].
\label{eq:method-observation-tokens}
\end{equation}
We append four learned readout tokens to $X_t$ and process the combined sequence with a Transformer. Observation tokens attend only to observation tokens. Readout tokens attend to both observation and readout tokens. The output representations of the four readout tokens are concatenated, normalized, and linearly projected into $z_t$, which is passed to the temporal backbone.

\paragraph{Temporal backbone.}
\label{sec:method-temporal}

The temporal backbone $T_\phi$ processes world tokens (without KV caching). We use a maximum context length of $C$ policy timesteps. At timestep $t$, the input is $z_{s_t:t}$, with
\begin{equation}
h_t=\left[T_\phi(z_{s_t:t})\right]_{\mathrm{last}}, \qquad
s_t=\max(1,t-C+1).
\label{eq:decomp}
\end{equation}
We use $K=1$ world token per policy timestep, so the input contains at most $C$ tokens. A causal Transformer processes these tokens in time order, allowing each token to attend to itself and earlier tokens. Its last output $h_t$ integrates the observation history $o_{s_t:t}$ and is passed to the DiT action head.

\paragraph{DiT action head.}
\label{sec:method-actions}

The DiT action head $\mathcal{D}_\psi$ takes $h_t$ from the temporal backbone as input and generates an $H$-step action chunk $\hat A_t\in\mathbb{R}^{H\times d_a}$, where $d_a$ is the action dimension. At inference, action generation starts from Gaussian noise. At each diffusion step, the DiT denoiser $\epsilon_\psi$ receives the noisy action chunk, the diffusion step, and $h_t$, and predicts the noise. Stochastic DDPM sampling~\citep{ho2020ddpm} iteratively denoises the action chunk to produce $\hat A_t$. The generated actions are unnormalized before execution. The robot executes the first $H_{\mathrm{exec}}$ actions before observing again at timestep $t+1$.

During training, we normalize an expert action chunk $A_t$ using statistics from the training demonstrations. At diffusion step $k$, we add Gaussian noise $\epsilon$ to obtain $A_t^{(k)}$. The DiT denoiser $\epsilon_\psi$ takes $A_t^{(k)}$, $k$, and $h_t$ as inputs and predicts the injected noise. The default training objective is
\begin{equation}
\mathcal{L}_{\mathrm{act}}
= \mathbb{E}_{t,k,\epsilon}
\left[
\left\|\epsilon - \epsilon_\psi\bigl(A_t^{(k)},k,h_t\bigr)\right\|_2^2
\right].
\label{eq:method-action-loss}
\end{equation}
All three modules are trained jointly with the action-generation objective, without auxiliary losses.%

\section{Evaluation framework}
\label{sec:evaluation-framework}

\paragraph{Evaluation measures.}
For RoboCasa, we use two complementary measures. Holdout action root-mean-square error (RMSE) assesses how well the policy imitates expert behavior. Given observations and histories from held-out demonstrations, we compare sampled action chunks with expert action chunks. We compute RMSE in the original, unnormalized action space and average the taskwise values across the 23 tasks. Closed-loop success rate (SR) assesses task completion when the policy's own actions determine subsequent observations. RMSE provides a continuous measure for fitting scaling trends and shows less variability than SR across training seeds in our experiments. The two measures broadly agree across our configuration grid, but meaningful gaps remain between action imitation and closed-loop task success. We therefore interpret them jointly (Appendix~\ref{app:rmse-sr}).

\paragraph{Data and model configurations.}
Performance at a single data scale cannot fully characterize a model's capabilities. The benefit of an architectural choice may also depend on the amount of training data. We therefore evaluate WorldToken at five data scales and five model sizes to study performance trends and compare design choices across data scales. For RoboCasa, we use $D \in \{50,100,300,1000,2900\}$ training demonstrations per task, denoted D50, D100, D300, D1000, and D2900, respectively. The five model configurations (Appendix~\ref{app:robocasa-training}) are N1 (44.3M parameters), N2 (85.3M), N3 (218.8M), N4 (648.9M), and N5 (1.49B). Parameter counts include the visual stems and policy modules, excluding the frozen CLIP text encoder.

\paragraph{Training and evaluation setup.}
We use causal attention and compute the action prediction loss at every valid timestep, following the sequence training structure of autoregressive language models. At position $i$ in a training window, the model predicts an action chunk from the first $i$ observations. Thus, $C_{\mathrm{train}}$ denotes the maximum history length, and $C_{\mathrm{train}}=10$ includes training with histories of length 1 through 10. For RoboCasa, all evaluations use the final checkpoint from each training run, without selecting checkpoints based on evaluation performance. At these checkpoints, RMSE is near its observed plateau across the scaling sweep  (Appendix~\ref{app:optimization-audit}). $C_{\mathrm{test}}$ denotes the visible history length at evaluation. Unless otherwise stated, we evaluate with $C_{\mathrm{test}}=C_{\mathrm{train}}=10$ and report mean SR over three complete evaluations of the same checkpoint under identical test settings.

We evaluate WorldToken through four questions: whether it supports effective multitask control and how its performance scales with data and model capacity (Section~\ref{sec:robocasa}); how the number of tokens per timestep affects performance and temporal computation (Section~\ref{sec:token-allocation}); whether trained policies rely on recent history (Section~\ref{sec:history}); and whether WorldToken can use longer histories (Section~\ref{sec:ranking}).

\section{Can WorldToken achieve effective multitask control and scale systematically?}
\label{sec:robocasa}

We first examine whether WorldToken supports effective multitask control and how performance scales with training data and model capacity. To this end, we evaluate WorldToken across the $5\times5$ grid defined in Section~\ref{sec:evaluation-framework}, training two policies with different random seeds for each configuration.

\subsection{Effective multitask control using RoboCasa data}
\label{sec:robocasa-scale}

\begin{figure}[t]
\centering
\includegraphics[width=\linewidth]{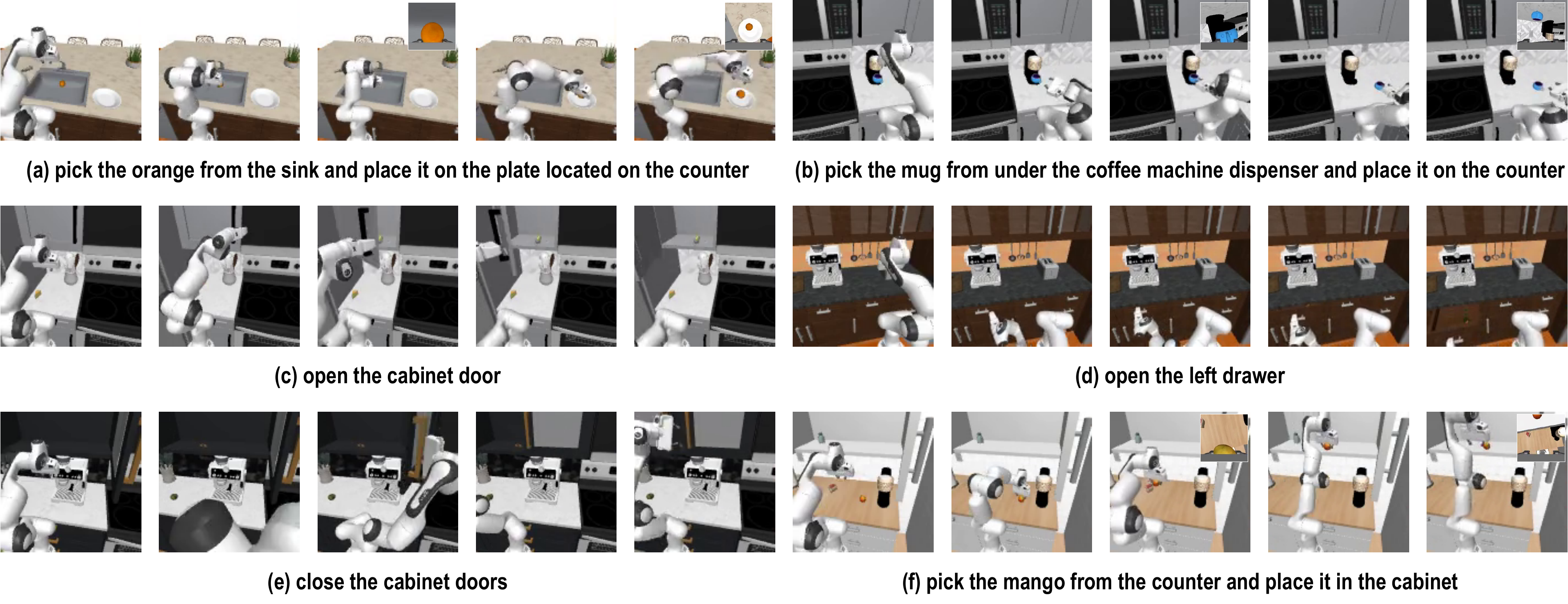}
\caption{Examples of manipulation process.}
\label{fig:robocasademo}
\end{figure}

WorldToken achieves effective multitask control with an 85.3M-parameter policy. All trainable policy modules are initialized from scratch, while the pretrained CLIP text encoder remains frozen. At D300, WorldToken reaches a mean SR of 46.83\% across two training seeds, compared with 31.28\% for our reproduction of BC-Transformer using its official code and the same task set, data scale, and evaluation episodes (Appendix~\ref{app:local-bc-transformer}). Increasing the training data to D2900 raises WorldToken's mean SR to 59.45\% (Table~\ref{tab:sweep}). For context, the 3.35B-parameter $\pi_{0.5}$ model is reported to achieve 62.1\% SR on RoboCasa Kitchen~\citep{kim2026a_rldx} . Manipulation examples are shown in Figure \ref{fig:robocasademo}.

\subsection{Scaling with data and model capacity}
\label{sec:robocasa-sweep}
\label{sec:data-scaling}
\label{sec:capacity-scaling}

Increasing the amount of training data consistently improves both RMSE and SR across all five model sizes (Table~\ref{tab:sweep}). From D50 to D2900, RMSE decreases by 47.0--56.8\% and SR increases by 32.7--39.3 percentage points, after averaging the two training seeds. Additional trajectories may improve coverage of scene and object configurations, helping the policy generalize across task states.

\begin{table}[t]
\caption{Performance of WorldToken across dataset sizes and model capacities on RoboCasa.}
\label{tab:sweep}
\centering
\definecolor{wtgreen}{HTML}{3E8F86}
\definecolor{wtblue}{HTML}{3B84A5}
\small
\setlength{\tabcolsep}{4pt}
\renewcommand{\arraystretch}{1.0}
\begin{tabular}{@{}cl ccccc|ccccc@{}}
\toprule
& & \multicolumn{5}{c|}{\textbf{Training seed 0}} & \multicolumn{5}{c}{\textbf{Training seed 1}} \\
\cmidrule(lr){3-7} \cmidrule(lr){8-12}
& & D50 & D100 & D300 & D1000 & D2900 & D50 & D100 & D300 & D1000 & D2900 \\
\midrule
\multirow{5}{*}{\rotatebox[origin=c]{90}{\textbf{SR (\%) $\uparrow$}}}
& N1  & \cellcolor{wtgreen!6}14.1 & \cellcolor{wtgreen!26}30.0 & \cellcolor{wtgreen!40}41.0 & \cellcolor{wtgreen!50}49.1 & \cellcolor{wtgreen!56}54.4 & \cellcolor{wtgreen!11}18.1 & \cellcolor{wtgreen!23}27.4 & \cellcolor{wtgreen!40}41.4 & \cellcolor{wtgreen!51}49.9 & \cellcolor{wtgreen!57}54.7 \\
& N2  & \cellcolor{wtgreen!15}21.4 & \cellcolor{wtgreen!27}31.2 & \cellcolor{wtgreen!48}47.5 & \cellcolor{wtgreen!58}55.7 & \cellcolor{wtgreen!62}59.1 & \cellcolor{wtgreen!12}18.9 & \cellcolor{wtgreen!29}32.4 & \cellcolor{wtgreen!46}46.2 & \cellcolor{wtgreen!54}52.2 & \cellcolor{wtgreen!63}59.8 \\
& N3 & \cellcolor{wtgreen!16}22.0 & \cellcolor{wtgreen!33}35.8 & \cellcolor{wtgreen!49}48.1 & \cellcolor{wtgreen!62}\textbf{58.5} & \cellcolor{wtgreen!63}59.3 & \cellcolor{wtgreen!17}22.7 & \cellcolor{wtgreen!30}33.2 & \cellcolor{wtgreen!49}48.4 & \cellcolor{wtgreen!61}\textbf{57.8} & \cellcolor{wtgreen!65}\textbf{61.2} \\
& N4 & \cellcolor{wtgreen!22}\textbf{27.1} & \cellcolor{wtgreen!34}\textbf{36.5} & \cellcolor{wtgreen!52}\textbf{51.2} & \cellcolor{wtgreen!61}57.7 & \cellcolor{wtgreen!62}58.8 & \cellcolor{wtgreen!18}\textbf{24.0} & \cellcolor{wtgreen!34}\textbf{36.3} & \cellcolor{wtgreen!53}\textbf{51.4} & \cellcolor{wtgreen!59}56.7 & \cellcolor{wtgreen!61}57.8 \\
& N5  & \cellcolor{wtgreen!17}23.1 & \cellcolor{wtgreen!31}33.7 & \cellcolor{wtgreen!51}49.8 & \cellcolor{wtgreen!59}56.2 & \cellcolor{wtgreen!63}\textbf{59.7} & \cellcolor{wtgreen!16}21.7 & \cellcolor{wtgreen!32}34.6 & \cellcolor{wtgreen!51}49.9 & \cellcolor{wtgreen!59}56.4 & \cellcolor{wtgreen!63}59.3 \\
\midrule
\multirow{5}{*}{\rotatebox[origin=c]{90}{\textbf{RMSE $\downarrow$}}}
& N1  & \cellcolor{wtblue!6}0.208 & \cellcolor{wtblue!22}0.175 & \cellcolor{wtblue!32}0.153 & \cellcolor{wtblue!44}0.129 & \cellcolor{wtblue!53}0.109 & \cellcolor{wtblue!8}0.203 & \cellcolor{wtblue!21}0.177 & \cellcolor{wtblue!33}0.152 & \cellcolor{wtblue!44}0.129 & \cellcolor{wtblue!53}0.109 \\
& N2  & \cellcolor{wtblue!12}0.195 & \cellcolor{wtblue!23}0.172 & \cellcolor{wtblue!40}0.137 & \cellcolor{wtblue!53}0.109 & \cellcolor{wtblue!62}0.091 & \cellcolor{wtblue!11}0.197 & \cellcolor{wtblue!25}0.169 & \cellcolor{wtblue!41}0.135 & \cellcolor{wtblue!53}0.110 & \cellcolor{wtblue!62}0.091 \\
& N3 & \cellcolor{wtblue!12}0.195 & \cellcolor{wtblue!26}\textbf{0.166} & \cellcolor{wtblue!42}\textbf{0.132} & \cellcolor{wtblue!58}\textbf{0.099} & \cellcolor{wtblue!65}\textbf{0.084} & \cellcolor{wtblue!13}0.194 & \cellcolor{wtblue!26}\textbf{0.165} & \cellcolor{wtblue!42}\textbf{0.133} & \cellcolor{wtblue!57}0.101 & \cellcolor{wtblue!65}\textbf{0.084} \\
& N4 & \cellcolor{wtblue!13}0.193 & \cellcolor{wtblue!25}0.169 & \cellcolor{wtblue!41}0.135 & \cellcolor{wtblue!57}0.101 & \cellcolor{wtblue!65}\textbf{0.084} & \cellcolor{wtblue!12}0.195 & \cellcolor{wtblue!24}0.170 & \cellcolor{wtblue!41}0.135 & \cellcolor{wtblue!57}\textbf{0.100} & \cellcolor{wtblue!65}\textbf{0.084} \\
& N5  & \cellcolor{wtblue!15}\textbf{0.189} & \cellcolor{wtblue!24}0.171 & \cellcolor{wtblue!36}0.144 & \cellcolor{wtblue!53}0.109 & \cellcolor{wtblue!64}0.087 & \cellcolor{wtblue!16}\textbf{0.186} & \cellcolor{wtblue!24}0.171 & \cellcolor{wtblue!36}0.144 & \cellcolor{wtblue!54}0.108 & \cellcolor{wtblue!64}0.086 \\
\bottomrule
\end{tabular}
\end{table}

\begin{figure}[!t]
\centering
\includegraphics[width=\linewidth]{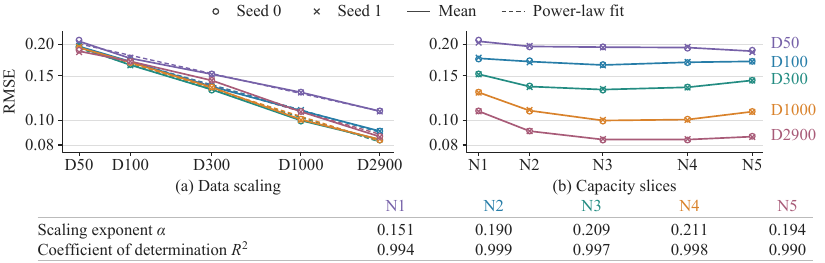}
\caption{On RoboCasa, WorldToken's RMSE decreases approximately as a power law with dataset size, while gains from increasing model capacity diminish. (a) RMSE versus dataset size for each model capacity. (b) RMSE versus model capacity for each dataset size. Both panels use log axes. Markers show individual training seeds, and solid lines connect the means across the two seeds. Dashed lines in (a) show fits of $\mathrm{RMSE}\propto D^{-\alpha}$ to these means. The table reports the fitted scaling exponent $\alpha$ and the coefficient of determination $R^2$ for each model, with $R^2$ computed in log--log space.}
\label{fig:scaling}
\end{figure}

The dependence of holdout RMSE on dataset size also varies with model capacity (Figure~\ref{fig:scaling}(a)). Across D50--D2900, RMSE decreases approximately as a power law. After averaging the two training seeds at each model capacity, we fit
\begin{equation}
\mathrm{RMSE}_{P}(D) \approx a_{P}D^{-\alpha_{P}},
\label{eq:power-law}
\end{equation}
where $P$ indexes model capacity. The fitted exponents range from 0.151 to 0.211, with $R^2$ values between 0.990 and 0.999. The exponent increases from 0.151 for N1 to approximately 0.21 for N3 and N4, indicating that these higher-capacity models make more effective use of additional data. This interaction between data and capacity resembles the coordinated scaling observed in language modeling \citep{kaplan2020scaling,hoffmann2022chinchilla}.

Figure~\ref{fig:scaling}(b) shows that the relative RMSE improvement from N1 to N3 increases with dataset size. Further increases to N4 and N5 yield no consistent reduction. These patterns suggest that the benefits of additional model capacity may be constrained by available data. However, the SR gap between smaller and larger models does not show the same pattern (Table~\ref{tab:sweep}). Larger models therefore gain a clearer advantage in action fitting without a corresponding increase in their advantage in task completion. This may reflect limits in data quantity and quality, particularly behavioral diversity. RoboCasa generates trajectories by transforming a fixed set of human demonstrations \citep{nasiriany2024robocasa}. Additional data may therefore reinforce similar behavior patterns, allowing larger models to fit demonstrated actions more accurately without comparable gains in closed-loop execution.

\section{Do more tokens per timestep improve performance?}
\label{sec:token-allocation}

The encoder processes each observation using multiple tokens, while the temporal backbone receives only one token per timestep. This raises a central question: how does passing more tokens per timestep to the temporal backbone affect control performance and computational cost?

We compare $K\in\{1,4,50\}$ across all five data settings, using the N2 policies from the scaling grid in Section~\ref{sec:robocasa} as the $K=1$ references. The $K=1$ and $K=4$ variants use the same encoder and differ only in the final projection. The $K=50$ variant passes all 50 fused observation tokens directly to the temporal backbone without compression. All variants share the training data and evaluation episodes at each data setting, and comparisons use seed 0.

\begin{table*}[t]
\caption{Token-interface comparisons with a context of $C=10$ policy timesteps. $K$ is the number of temporal tokens contributed by each timestep. Relative FLOPs (RelPs) are the leading analytic full-prefix temporal-backbone counts (Appendix~\ref{app:temporal-flops}), normalized to the N2, $K=1$ reference. Encoder and action-head costs are excluded. Each SR entry is the mean over three complete evaluations.}
\label{tab:token-allocation-scale}
\centering
\small
\setlength{\tabcolsep}{4pt}
\renewcommand{\arraystretch}{1.0}
\begin{tabular}{@{}rclrccccc|ccccc@{}}
\toprule
& & & & \multicolumn{5}{c|}{\textbf{SR (\%) $\uparrow$}} & \multicolumn{5}{c}{\textbf{RMSE $\downarrow$}} \\
\cmidrule(lr){5-9} \cmidrule(lr){10-14}
$K$ & Seed & Params & RelPs & D50 & D100 & D300 & D1000 & D2900 & D50 & D100 & D300 & D1000 & D2900 \\
\midrule
1  & 0 & 85.3M & $1.0\times$  & 21.4 & 31.2 & 47.5 & 55.7 & 59.1 & 0.195 & 0.172 & 0.137 & 0.109 & 0.091 \\
1  & 1 & 85.3M & $1.0\times$  & 18.9 & 32.4 & 46.2 & 52.2 & 59.8 & 0.197 & 0.169 & 0.135 & 0.110 & 0.091 \\
4  & 0 & 92.4M & $4.0\times$  & 20.3 & \textbf{38.1} & 48.8 & 55.6 & 59.9 & 0.202 & 0.166 & 0.138 & 0.110 & 0.090 \\
50 & 0 & 83.0M & $55.3\times$ & \textbf{26.5} & 37.6 & \textbf{49.6} & \textbf{57.7} & \textbf{60.7} & \textbf{0.191} & \textbf{0.165} & \textbf{0.132} & \textbf{0.107} & \textbf{0.089} \\
\bottomrule
\end{tabular}
\end{table*}

Increasing $K$ from 1 to 4 yields no consistent RMSE improvement across data scales, while SR improves at some settings, most clearly at D100 (Table~\ref{tab:token-allocation-scale}). The two variants use the same encoder apart from the final projection, so adding tokens per timestep brings no clear overall benefit.

Passing all 50 fused observation tokens directly to the temporal backbone reduces RMSE by 1.7--4.2\% and improves SR by 1.6--6.4 percentage points across the five data scales (Table~\ref{tab:token-allocation-scale}). These modest gains suggest that a single token already preserves the vast majority of the information needed for action prediction and control. However, the SR advantage of $K=50$ over $K=1$ is larger at D50 and D100 than at D300 and above, suggesting that learning this compact representation may be harder with limited training data. Retaining all 50 tokens avoids this compression, while more training data may help the encoder learn which information to retain in a single token.

Table~\ref{tab:token-allocation-scale} shows that, at $C=10$, $K=4$ and $K=50$ require 4.03 and 55.31 times the temporal-backbone FLOPs of $K=1$, respectively. With KV caching, the rate at which temporal-backbone FLOPs increase with history length is proportional to $K^2$. Although CNN visual stems account for roughly 80--90\% of total policy FLOPs at $C=10$, using more tokens per timestep causes temporal computation to grow rapidly as history becomes longer (Appendix~\ref{app:temporal-flops}--\ref{app:history-growth}). Using one token per timestep therefore balances control performance and the cost of modeling long histories. These results support the policy timestep as a suitable sequence unit for modeling robot interaction histories.

\section{Does WorldToken use recent history?}
\label{sec:history}

WorldToken predicts actions using current and past observations. To examine how much trained policies rely on recent history, we evaluate each of the 50 policies from Section~\ref{sec:robocasa} once at each shorter evaluation history length $C_{\mathrm{test}}\in{1,2,5}$, keeping all other evaluation settings fixed and using the reported results at $C_{\mathrm{test}}=10$ as the reference. Table~\ref{tab:truncation} shows that reducing the evaluation history from ten to one or two policy timesteps lowers SR for all 50 policies, while using five timesteps retains most of their performance. These policies therefore make substantial use of recent observations, which may help them track motion.

\begin{table}[t]
\caption{SR changes when the same policies are evaluated with shorter histories. Each cell reports $\Delta\mathrm{SR}=\mathrm{SR}(C_{\mathrm{test}})-\mathrm{SR}(10)$ in percentage points for one, two, and five visible policy timesteps.}
\label{tab:truncation}
\centering
\small

\definecolor{wtblue}{HTML}{3B84A5}
\definecolor{wtorange}{HTML}{BF7A4A}
\newcommand{\WTB}[2]{\cellcolor{wtblue!#1}$#2$}
\newcommand{\WTO}[2]{\cellcolor{wtorange!#1}$#2$}

\setlength{\tabcolsep}{1pt}
\renewcommand{\arraystretch}{1.0}

\begin{tabular}{@{}clccccc|ccccc|ccccc@{}}
\toprule
& & \multicolumn{5}{c|}{$C_{\mathrm{test}}=1$}
    & \multicolumn{5}{c|}{$C_{\mathrm{test}}=2$}
    & \multicolumn{5}{c}{$C_{\mathrm{test}}=5$} \\
\cmidrule(lr){3-7}
\cmidrule(lr){8-12}
\cmidrule(lr){13-17}
& & D50 & D100 & D300 & D1000 & D2900
  & D50 & D100 & D300 & D1000 & D2900
  & D50 & D100 & D300 & D1000 & D2900 \\
\midrule

\multirow{5}{*}{\rotatebox[origin=c]{90}{\textbf{Seed 0}}}
& N1
  & \WTB{10}{-3.5} & \WTB{30}{-10.1} & \WTB{53}{-18.1} & \WTB{44}{-14.9} & \WTB{38}{-13.0}
  & \WTB{12}{-4.1} & \WTB{10}{-3.3} & \WTB{32}{-11.0} & \WTB{25}{-8.6} & \WTB{20}{-6.8}
  & \WTB{13}{-0.9} & \WTO{6}{+0.4} & \WTB{7}{-0.5} & \WTB{1}{-0.1} & \WTB{9}{-0.6} \\

& N2
  & \WTB{21}{-7.1} & \WTB{26}{-8.7} & \WTB{53}{-17.8} & \WTB{53}{-18.0} & \WTB{45}{-15.3}
  & \WTB{16}{-5.3} & \WTB{20}{-6.7} & \WTB{36}{-12.3} & \WTB{19}{-6.5} & \WTB{18}{-6.2}
  & \WTO{4}{+0.3} & \WTB{1}{-0.1} & \WTB{16}{-1.1} & \WTO{12}{+0.8} & \WTO{22}{+1.5} \\

& N3
  & \WTB{22}{-7.3} & \WTB{33}{-11.1} & \WTB{60}{-20.2} & \WTB{60}{-20.3} & \WTB{56}{-18.9}
  & \WTB{13}{-4.4} & \WTB{29}{-9.9} & \WTB{39}{-13.2} & \WTB{35}{-11.7} & \WTB{17}{-5.6}
  & \WTO{4}{+0.3} & \WTB{4}{-0.3} & \WTB{16}{-1.1} & \WTB{9}{-0.6} & $0.0$ \\

& N4
  & \WTB{31}{-10.5} & \WTB{44}{-15.0} & \WTB{63}{-21.3} & \WTB{56}{-18.9} & \WTB{65}{-22.2}
  & \WTB{26}{-8.9} & \WTB{33}{-11.3} & \WTB{35}{-11.9} & \WTB{30}{-10.1} & \WTB{27}{-9.2}
  & \WTB{20}{-1.4} & \WTB{27}{-1.9} & \WTB{19}{-1.3} & \WTB{9}{-0.6} & \WTB{14}{-1.0} \\

& N5
  & \WTB{19}{-6.4} & \WTB{40}{-13.6} & \WTB{58}{-19.7} & \WTB{51}{-17.3} & \WTB{55}{-18.6}
  & \WTB{17}{-5.7} & \WTB{25}{-8.5} & \WTB{38}{-12.7} & \WTB{28}{-9.4} & \WTB{26}{-8.8}
  & \WTO{7}{+0.5} & \WTB{6}{-0.4} & \WTB{22}{-1.5} & \WTO{22}{+1.5} & \WTB{14}{-1.0} \\

\midrule

\multirow{5}{*}{\rotatebox[origin=c]{90}{\textbf{Seed 1}}}
& N1
  & \WTB{17}{-5.6} & \WTB{17}{-5.7} & \WTB{31}{-10.4} & \WTB{41}{-14.0} & \WTB{42}{-14.1}
  & \WTB{8}{-2.6} & \WTB{14}{-4.8} & \WTB{4}{-1.4} & \WTB{22}{-7.4} & \WTB{17}{-5.6}
  & \WTO{9}{+0.6} & \WTB{3}{-0.2} & \WTO{12}{+0.8} & \WTB{1}{-0.1} & \WTO{9}{+0.6} \\

& N2
  & \WTB{24}{-8.0} & \WTB{32}{-10.7} & \WTB{49}{-16.6} & \WTB{53}{-17.8} & \WTB{51}{-17.2}
  & \WTB{16}{-5.4} & \WTB{22}{-7.4} & \WTB{29}{-9.7} & \WTB{20}{-6.6} & \WTB{20}{-6.6}
  & \WTB{12}{-0.8} & $0.0$ & \WTO{3}{+0.2} & \WTO{7}{+0.5} & \WTB{16}{-1.1} \\

& N3
  & \WTB{25}{-8.4} & \WTB{34}{-11.5} & \WTB{61}{-20.6} & \WTB{60}{-20.2} & \WTB{57}{-19.2}
  & \WTB{22}{-7.6} & \WTB{22}{-7.3} & \WTB{37}{-12.6} & \WTB{30}{-10.1} & \WTB{21}{-7.2}
  & \WTB{30}{-2.1} & \WTO{7}{+0.5} & \WTB{3}{-0.2} & \WTB{9}{-0.6} & \WTB{30}{-2.1} \\

& N4
  & \WTB{22}{-7.6} & \WTB{56}{-18.9} & \WTB{65}{-22.1} & \WTB{56}{-19.0} & \WTB{56}{-18.9}
  & \WTB{27}{-9.0} & \WTB{33}{-11.2} & \WTB{39}{-13.2} & \WTB{30}{-10.0} & \WTB{17}{-5.8}
  & \WTB{9}{-0.6} & \WTB{25}{-1.7} & \WTB{22}{-1.5} & \WTO{14}{+1.0} & \WTO{27}{+1.9} \\

& N5
  & \WTB{20}{-6.9} & \WTB{43}{-14.7} & \WTB{64}{-21.5} & \WTB{56}{-18.9} & \WTB{56}{-18.8}
  & \WTB{17}{-5.6} & \WTB{30}{-10.2} & \WTB{45}{-15.1} & \WTB{28}{-9.4} & \WTB{32}{-10.7}
  & \WTB{33}{-2.3} & $0.0$ & \WTB{17}{-1.2} & \WTO{14}{+1.0} & \WTO{1}{+0.1} \\

\bottomrule
\end{tabular}
\end{table}

We next examine the effect of shorter training histories by training and evaluating N3 policies at D300 with $C_{\mathrm{train}}=C_{\mathrm{test}}\in\{1,2,5\}$. Training with these shorter histories recovers much of the SR lost when history is shortened only at evaluation (Tables~\ref{tab:truncation} and~\ref{tab:matched-context}). At $C=1$, these policies achieve 47.22\% mean SR, compared with 27.87\% for policies trained with $C_{\mathrm{train}}=10$ and evaluated with $C_{\mathrm{test}}=1$. This recovery suggests that training can change how much a policy relies on history, even when the task itself may not require it.

\begin{table}[t]
\centering
\small
\caption{RMSE and SR at different history lengths, with $C_{\mathrm{train}}=C_{\mathrm{test}}=C$.}

\label{tab:matched-context}
\begin{tabular}{ccccc}
\toprule
Context $C$ & RMSE, seed 0 & RMSE, seed 1 & SR (\%) , seed 0 & SR (\%) , seed 1 \\
\midrule
1  & 0.150260 & 0.147767 & 46.75 $\pm$ 1.02 & 47.68 $\pm$ 1.50 \\
2  & 0.149383 & 0.147528 & 46.87 $\pm$ 0.84 & 48.52 $\pm$ 1.36 \\
5  & 0.137165 & 0.139603 & \textbf{51.48 $\pm$ 0.26} & \textbf{50.00 $\pm$ 0.92} \\
10 & 0.131710 & 0.133098 & 48.12\% $\pm$ 1.70 pp & 48.38 $\pm$ 1.09 \\
\bottomrule
\end{tabular}
\end{table}

\begin{figure}[!t]
\centering
\includegraphics[width=0.9\linewidth]{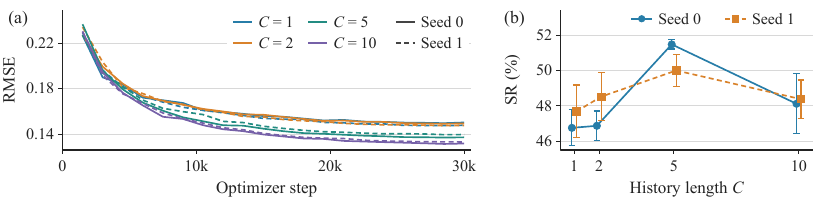}
\caption{History length for N3 on D300 ($C_{\mathrm{train}}=C_{\mathrm{test}}=C$). (a) Holdout action RMSE during training. (b) Mean final SR over three fixed-seed repeats per checkpoint. Error bars show sample standard deviations.}

\label{fig:matched-context}
\end{figure}

Although RMSE decreases with longer training and evaluation contexts, SR peaks at $C=5$ and falls at $C=10$ (Table~\ref{tab:matched-context} and Figure~\ref{fig:matched-context}). Longer histories may help the policy imitate expert actions, but also make it more sensitive to differences between expert demonstrations and its own execution.

\section{Can WorldToken use longer histories?}
\label{sec:ranking}

The RoboCasa results suggest that a policy's reliance on recent history can reflect how it was trained, rather than a requirement of the task itself. We next turn to RMBench Blocks Ranking, where the current observation alone does not determine the next swap in the demonstrated sequence, which allows us to examine whether WorldToken can use longer histories to complete successive swaps.

\begin{table}[!t]
\centering
\small
\caption{Blocks Ranking outcomes stratified by the number of swaps $S$ in the
reference solution. Each entry gives evaluator successes / strict behavioral
successes / evaluator-only successes.}
\label{tab:k-stratified}
\begin{tabular}{ccccccc}
\toprule
$C_{\mathrm{test}} \backslash S$ (episodes) & 1 (24) & 2 (14) & 3 (28) & 4 (18) & 5 (16) & Total (100) \\
\midrule
608 & 24/24/0 & 14/14/0 & 25/25/0 & 17/16/1 & 15/15/0 & 95/94/1 \\
288 & 24/24/0 & 14/14/0 & 23/23/0 & 16/16/0 & 15/15/0 & 92/92/0 \\
128 & 24/24/0 & 14/14/0 & 13/3/10 & 5/5/0 & 3/3/0 & 59/49/10 \\
64 & 24/24/0 & 3/3/0 & 2/0/2 & 9/0/9 & 0/0/0 & 38/27/11 \\
32 & 24/24/0 & 3/0/3 & 0/0/0 & 1/0/1 & 0/0/0 & 28/24/4 \\
\bottomrule
\end{tabular}
\end{table}

\begin{figure}[!t]
\centering
\includegraphics[width=\linewidth]{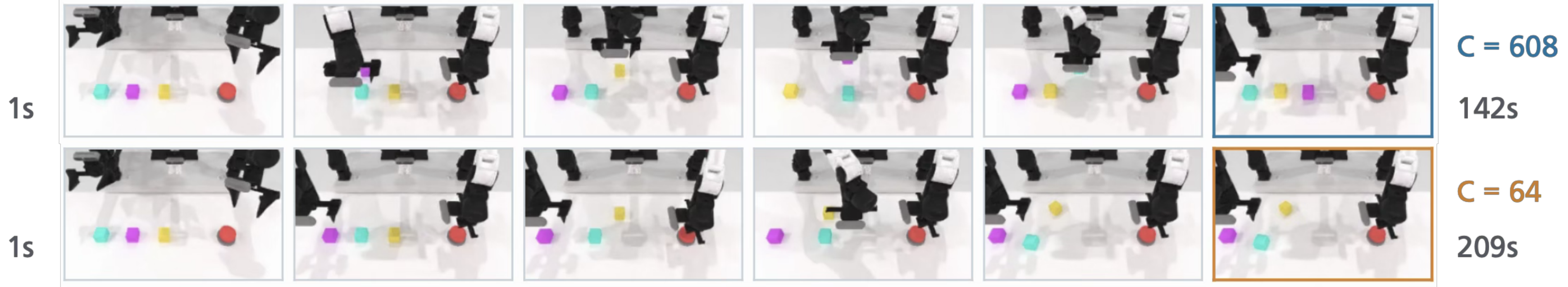}
\caption{One episode evaluated with two history lengths. The checkpoint, initial state, and evaluation protocol are fixed, only $C_{\mathrm{test}}$ changes. With $C_{\mathrm{test}}=608$, the policy completes the five-swap reference sequence and succeeds at 142.6 seconds. With $C_{\mathrm{test}}=64$, it completes the first swap but fails to finish the task within the 210-second horizon. Frame timestamps are simulated seconds.}
\label{fig:ranking-storyboard}
\end{figure}

Three colored blocks occupy left, middle, and right slots $(L,M,R)$, with target order $(1,2,3)$. In the demonstrations, the robot first presses the button, then follows the reference swap sequence
\begin{equation}
\mathrm{swap}(M,R)
\rightarrow \mathrm{swap}(L,R)
\rightarrow \mathrm{swap}(L,M)
\rightarrow \mathrm{swap}(M,R)
\rightarrow \mathrm{swap}(L,R),
\label{eq:swap-seq}
\end{equation}
pressing after each swap and stopping at the target. The five non-target orders require 1--5 swaps.

We report evaluator success when the official checks for final block positions, an open right gripper, and a button press are satisfied. These checks do not require following the demonstrated swap sequence. We therefore also report \emph{strict behavioral success}, which additionally requires following the reference swap sequence and placing the blocks accurately in the target slots. Evaluator successes that fail either additional criterion are counted as \emph{evaluator-only successes} (Appendix~\ref{app:ranking-behavior-protocol}).

The policy uses the N2 temporal-backbone configuration from Section~\ref{sec:evaluation-framework} and a maximum training history of $C_{\mathrm{train}}=608$. Model and training details are provided in Appendices~\ref{app:ranking-training} and~\ref{app:ranking-objective}. We evaluate the same checkpoint on 100 initial conditions at $C_{\mathrm{test}}\in\{608,288,128,64,32\}$, keeping other evaluation settings fixed. Each policy timestep spans 0.24 seconds, so these history windows cover approximately 8--146 seconds. A complete swap followed by a button press takes about 28 seconds. The tested windows therefore cover less than one to about five such cycles.

\subsection{Longer histories help complete multiple swaps}
\label{sec:ranking-truncation}

Increasing the evaluation history from 32 to 608 timesteps raises evaluator success from 28\% to 95\% and strict behavioral success from 24\% to 94\% (Table~\ref{tab:k-stratified}). All 24 episodes requiring one swap succeed at every tested history length, while shorter histories reduce success in episodes requiring multiple swaps. Figure~\ref{fig:ranking-storyboard} illustrates this difference. With $C_{\mathrm{test}}=64$, the policy completes the first swap but fails to finish the task, whereas $C_{\mathrm{test}}=608$ supports all five swaps. The main failures arise from errors in carrying out the swaps rather than an incorrect swap order. A block may be placed away from its intended position, after which the policy stalls without correcting the placement. These results suggest that longer histories help the policy complete successive swaps more reliably. %

\subsection{Continuing beyond demonstrations and the history window}
\label{sec:ranking-stress}

We also disable termination at first success to examine whether the policy can continue beyond the demonstrated sequences. This exploratory test uses nine initial conditions covering all five non-target initial arrangements. With $C_{\mathrm{test}}=608$ fixed, the history window begins to slide after 145.92 seconds. %

\begin{table}[t]
\caption{Nine exploratory Blocks Ranking stress trajectories
with rollouts allowed to continue after first success.
$S$ is the number of swaps required to first reach the target.
``Swaps'' counts correctly ordered reference-sequence swaps.
``Last'' is the time of the final correct swap.}
\label{tab:stress}
\centering
\small
\setlength{\tabcolsep}{4pt}
\renewcommand{\arraystretch}{1.0}
\begin{tabular}{@{}crrr|crrr|crrr@{}}
\toprule
Env.\ seed & $S$ & Swaps & Last (s) &
Env.\ seed & $S$ & Swaps & Last (s) &
Env.\ seed & $S$ & Swaps & Last (s) \\
\midrule
100000 & 2 & 13 & 350.52 & 100004 & 3 & 8 & 222.12 & 100015 & 5 &  5 & 139.38 \\
100001 & 3 &  5 & 139.20 & 100003 & 2 & 9 & 250.14 & 100016 & 1 &  5 & 139.32 \\
100002 & 4 & 17 & 470.28 & 100007 & 1 & 5 & 139.08 & 100008 & 1 & 31 & \textbf{856.44} \\
\bottomrule
\end{tabular}
\end{table}

Table~\ref{tab:stress} shows that five of the nine trajectories continue the reference sequence after the history window begins to slide. The longest completes 31 correctly ordered swaps, with the last correct swap occurring at 856.44 seconds. These results suggest that the policy has learned a repeating cycle of three swaps, $\mathrm{swap}(M,R)\rightarrow\mathrm{swap}(L,R)\rightarrow\mathrm{swap}(L,M)$. By repeating this cycle, it can continue beyond the maximum of five swaps in the demonstrations, even after the initial observations have left the history window and the policy encounters history prefixes not seen during training.

\section{Related Work}
\label{sec:related-main}

\paragraph{Sequence modeling and vision-language-action policies.} Sequence modeling is well established in decision making and generalist
agents \citep{chen2021decision,janner2021trajectory,zheng2022online,reed2022gato}. Robotic Transformer policies further explore multimodal conditioning
and different ways to organize observations and actions \citep{brohan2023rt1,jiang2023vima,octo2024,wang2024hpt,fu2024icrt}. More recently, VLA models have adopted pretrained vision-language
backbones for manipulation, combining transferred perceptual representations with discrete or continuous action generation \citep{brohan2023rt2,kim2025openvla,black2025pi0,li2024cogact,bjorck2025groot}. For history-conditioned control, inheriting a perceptual representation
leaves an additional design choice: how observations should enter cross-timestep sequence modeling. We argue that the basic unit of
temporal context should be examined as an explicit architectural choice, separately from the perceptual tokenization used within each timestep.

\paragraph{Temporal modeling and memory in robotics.}
Incorporating interaction history is a longstanding approach to control under partial observability, spanning recurrent imitation policies and attention-based models of observation histories \citep{mandlekar2022whatmatters,guhur2023hiveformer,li2024roboflamingo}. Recent memory-augmented VLAs extend this direction through historical compression, temporal aggregation, and memory retrieval and consolidation \citep{koo2026hamlet,shi2026memoryvla,torne2026mem,wang2026nativemem}. Related approaches encode past interactions as visual traces \citep{zheng2025tracevla} or organize longer histories through hierarchical and recurrent architectures \citep{shah2026aprism,zhou2026chronos}. Beyond architectural design, research also examines history relevance, training across context lengths, and the evaluation of memory-dependent behavior \citep{shah2026bhalo,agarwal2026longcontext,chen2026rmbench,dai2026robomme}. Within this literature, WorldToken studies time-first sequence modeling for robotic imitation learning, focusing on the relationship between within-timestep perceptual granularity and the organization of temporal context. See Appendix~\ref{sec:related}.

\section{Conclusion}
In this work, we propose WorldToken, which provides a time-first answer to how robot interaction histories should be organized. WorldToken encodes each policy timestep into one world token and separates within-timestep perception from cross-timestep history modeling. In this way, observations can be organized just like word tokens in an LLM, and hence benefit from scaling and support long-history tasks. Extensive experiments demonstrate the feasibility of this organization for multitask robotic manipulation and extended use of history. Beyond feasibility, our analysis indicates that a single world token per policy timestep preserves most decision-relevant information, since retaining all fused observation tokens improves closed-loop success only marginally at substantially higher backbone cost, and that history dependence is shaped largely by the training context, which provide a reference for future work. Although we only test our method on manipulation tasks, our method provides a clear, general modeling principle and a modular architecture that facilitates ablation studies, attribution of performance changes, and targeted improvements. We plan to compare our method against other baselines using larger datasets with greater data diversity in both simulated and real-world tasks in the future.

\section*{Reproducibility statement}
The main paper specifies the model architectures, training setups, and experimental protocols, and the appendix provides additional details on implementations, training, rollout, aggregation, context analyses, task specifications, and qualitative results. The planned public release includes training code, evaluation code, logs, and models.

\section*{AI use statement}
Large language models were used throughout this work under author direction: for literature collection, experimental statistics, code development, and manuscript drafting, with additional assistance in experimental design. All core ideas, the experimental program, and the conclusions originate from the authors. The authors verified the paper's numerical claims against the underlying configurations, logs, evaluation records, and analysis artifacts, reviewed all AI-assisted text, code, and figures, and take full responsibility for the paper's content.

\bibliography{refs}

@article{agarwal2026longcontext,
  author  = {Agarwal, Abhinav and Wei, Adam and Kargin, Taylan and Zeng, Michael and Becker, Cole and Dayi, Arif Kerem and Parrilo, Pablo and Ozdaglar, Asuman and Tedrake, Russ},
  title   = {{Training and evaluating diffusion policies with long context lengths}},
  journal = {arXiv preprint arXiv:2606.16447},
  year    = {2026},
  url     = {https://arxiv.org/abs/2606.16447}
}

@misc{arise2024robomimic,
  author       = {{ARISE Initiative}},
  title        = {{RoboCasa BC-Transformer implementation in the robocasa branch of robomimic}},
  howpublished = {Official software repository},
  year         = {2024},
  url          = {https://github.com/ARISE-Initiative/robomimic/tree/robocasa}
}

@inproceedings{assran2023jepa,
  author    = {Assran, Mahmoud and Duval, Quentin and Misra, Ishan and Bojanowski, Piotr and Vincent, Pascal and Rabbat, Michael and LeCun, Yann and Ballas, Nicolas},
  title     = {{Self-supervised learning from images with a joint-embedding predictive architecture}},
  booktitle = {IEEE/CVF Conference on Computer Vision and Pattern Recognition},
  year      = {2023},
  url       = {https://arxiv.org/abs/2301.08243}
}

@article{assran2025vjepa2,
  author  = {Assran, Mido and Bardes, Adrien and Fan, David and Garrido, Quentin and Howes, Russell and Komeili, Mojtaba and Muckley, Matthew and Rizvi, Ammar and Roberts, Claire and Sinha, Koustuv and others},
  title   = {{V-JEPA 2: Self-supervised video models enable understanding, prediction and planning}},
  journal = {arXiv preprint arXiv:2506.09985},
  year    = {2025},
  url     = {https://arxiv.org/abs/2506.09985}
}

@article{bjorck2025groot,
  author  = {Bjorck, Johan and Casta{\~n}eda, Fernando and Cherniadev, Nikita and Da, Xingye and Ding, Runyu and Fan, Linxi and Fang, Yu and Fox, Dieter and Hu, Fengyuan and Huang, Spencer and others},
  title   = {{GR00T N1: An open foundation model for generalist humanoid robots}},
  journal = {arXiv preprint arXiv:2503.14734},
  year    = {2025},
  url     = {https://arxiv.org/abs/2503.14734}
}

@inproceedings{black2025pi0,
  author    = {Black, Kevin and Brown, Noah and Driess, Danny and Esmail, Adnan and Equi, Michael Robert and Finn, Chelsea and Fusai, Niccolo and Groom, Lachy and Hausman, Karol and Ichter, Brian and Jakubczak, Szymon and Jones, Tim and Ke, Liyiming and Levine, Sergey and Li-Bell, Adrian and Mothukuri, Mohith and Nair, Suraj and Pertsch, Karl and Shi, Lucy Xiaoyang and Smith, Laura and Tanner, James and Vuong, Quan and Walling, Anna and Wang, Haohuan and Zhilinsky, Ury},
  title     = {{$\pi_0$: A Vision-Language-Action Flow Model for General Robot Control}},
  booktitle = {Robotics: Science and Systems},
  year      = {2025},
  doi       = {10.15607/RSS.2025.XXI.010},
  url       = {https://arxiv.org/abs/2410.24164}
}

@inproceedings{brohan2023rt1,
  author    = {Brohan, Anthony and Brown, Noah and Carbajal, Justice and Chebotar, Yevgen and Dabis, Joseph and Finn, Chelsea and Gopalakrishnan, Keerthana and Hausman, Karol and Herzog, Alex and Hsu, Jasmine and Ibarz, Julian and Ichter, Brian and Irpan, Alex and Jackson, Tomas and Jesmonth, Sally and Joshi, Nikhil J. and Julian, Ryan and Kalashnikov, Dmitry and Kuang, Yuheng and Leal, Isabel and Lee, Kuang-Huei and Levine, Sergey and Lu, Yao and Malla, Utsav and Manjunath, Deeksha and Mordatch, Igor and Nachum, Ofir and Parada, Carolina and Peralta, Jodilyn and Perez, Emily and Pertsch, Karl and Quiambao, Jornell and Rao, Kanishka and Ryoo, Michael S. and Salazar, Grecia and Sanketi, Pannag R. and Sayed, Kevin and Singh, Jaspiar and Sontakke, Sumedh and Stone, Austin and Tan, Clayton and Tran, Huong and Vanhoucke, Vincent and Vega, Steve and Vuong, Quan H. and Xia, Fei and Xiao, Ted and Xu, Peng and Xu, Sichun and Yu, Tianhe and Zitkovich, Brianna},
  title     = {{RT-1: Robotics transformer for real-world control at scale}},
  booktitle = {Robotics: Science and Systems},
  year      = {2023},
  doi       = {10.15607/RSS.2023.XIX.025},
  url       = {https://arxiv.org/abs/2212.06817}
}

@inproceedings{brohan2023rt2,
  author    = {Brohan, Anthony and Brown, Noah and Carbajal, Justice and Chebotar, Yevgen and Chen, Xi and Choromanski, Krzysztof and Ding, Tianli and Driess, Danny and Dubey, Avinava and Finn, Chelsea and Florence, Pete and Fu, Chuyuan and Gonzalez Arenas, Montse and Gopalakrishnan, Keerthana and Han, Kehang and Hausman, Karol and Herzog, Alexander and Hsu, Jasmine and Ichter, Brian and Irpan, Alex and Joshi, Nikhil and Julian, Ryan and Kalashnikov, Dmitry and Kuang, Yuheng and Leal, Isabel and Lee, Lisa and Lee, Tsang-Wei Edward and Levine, Sergey and Lu, Yao and Michalewski, Henryk and Mordatch, Igor and Pertsch, Karl and Rao, Kanishka and Reymann, Krista and Ryoo, Michael and Salazar, Grecia and Sanketi, Pannag and Sermanet, Pierre and Singh, Jaspiar and Singh, Anikait and Soricut, Radu and Tran, Huong and Vanhoucke, Vincent and Vuong, Quan and Wahid, Ayzaan and Welker, Stefan and Wohlhart, Paul and Wu, Jialin and Xia, Fei and Xiao, Ted and Xu, Peng and Xu, Sichun and Yu, Tianhe and Zitkovich, Brianna},
  title     = {{RT-2}: Vision-Language-Action Models Transfer Web Knowledge to Robotic Control},
  booktitle = {Proceedings of the 7th Conference on Robot Learning},
  series    = {Proceedings of Machine Learning Research},
  volume    = {229},
  pages     = {2165--2183},
  year      = {2023},
  publisher = {PMLR},
  url       = {https://proceedings.mlr.press/v229/zitkovich23a.html}
}

@inproceedings{chen2021decision,
  author    = {Chen, Lili and Lu, Kevin and Rajeswaran, Aravind and Lee, Kimin and Grover, Aditya and Laskin, Michael and Abbeel, Pieter and Srinivas, Aravind and Mordatch, Igor},
  title     = {{Decision transformer: Reinforcement learning via sequence modeling}},
  booktitle = {Advances in Neural Information Processing Systems},
  volume    = {34},
  year      = {2021},
  url       = {https://arxiv.org/abs/2106.01345}
}

@article{chen2026rmbench,
  author  = {Chen, Tianxing and Wang, Yuran and Li, Mingleyang and Qin, Yan and Shi, Hao and Li, Zixuan and Hu, Yifan and Zhang, Yingsheng and Wang, Kaixuan and Chen, Yue and others},
  title   = {{RMBench: Memory-dependent robotic manipulation benchmark with insights into policy design}},
  journal = {arXiv preprint arXiv:2603.01229},
  year    = {2026},
  url     = {https://arxiv.org/abs/2603.01229}
}

@inproceedings{chi2023diffusion,
  author    = {Chi, Cheng and Feng, Siyuan and Du, Yilun and Xu, Zhenjia and Cousineau, Eric and Burchfiel, Benjamin C. M. and Song, Shuran},
  title     = {{Diffusion policy: Visuomotor policy learning via action diffusion}},
  booktitle = {Robotics: Science and Systems},
  year      = {2023},
  doi       = {10.15607/RSS.2023.XIX.026},
  url       = {https://arxiv.org/abs/2303.04137}
}

@inproceedings{dai2026robomme,
  author    = {Dai, Yinpei and Fu, Hongze and Lee, Jayjun and Liu, Yuejiang and Zhang, Haoran and Yang, Jianing and Finn, Chelsea and Fazeli, Nima and Chai, Joyce},
  title     = {{RoboMME: Benchmarking and understanding memory for robotic generalist policies}},
  booktitle = {International Conference on Machine Learning},
  year      = {2026},
  url       = {https://arxiv.org/abs/2603.04639}
}

@article{fu2024icrt,
  author  = {Fu, Letian and Huang, Huang and Datta, Gaurav and Chen, Lawrence Yunliang and Panitch, William Chung-Ho and Liu, Fangchen and Li, Hui and Goldberg, Ken},
  title   = {{In-context imitation learning via next-token prediction}},
  journal = {arXiv preprint arXiv:2408.15980},
  year    = {2024},
  url     = {https://arxiv.org/abs/2408.15980}
}

@inproceedings{guhur2023hiveformer,
  author    = {Guhur, Pierre-Louis and Chen, Shizhe and Pinel, Ricardo Garcia and Tapaswi, Makarand and Laptev, Ivan and Schmid, Cordelia},
  title     = {{Instruction-driven history-aware policies for robotic manipulations}},
  booktitle = {Conference on Robot Learning},
  series    = {Proceedings of Machine Learning Research},
  volume    = {205},
  pages     = {175--187},
  year      = {2023},
  url       = {https://arxiv.org/abs/2209.04899},
  note      = {CoRL 2022 proceedings}
}

@inproceedings{hafner2019planet,
  author    = {Hafner, Danijar and Lillicrap, Timothy and Fischer, Ian and Villegas, Ruben and Ha, David and Lee, Honglak and Davidson, James},
  title     = {{Learning latent dynamics for planning from pixels}},
  booktitle = {International Conference on Machine Learning},
  series    = {Proceedings of Machine Learning Research},
  volume    = {97},
  year      = {2019},
  url       = {https://arxiv.org/abs/1811.04551}
}

@inproceedings{hafner2020dreamer,
  author    = {Hafner, Danijar and Lillicrap, Timothy and Ba, Jimmy and Norouzi, Mohammad},
  title     = {{Dream to control: Learning behaviors by latent imagination}},
  booktitle = {International Conference on Learning Representations},
  year      = {2020},
  url       = {https://arxiv.org/abs/1912.01603}
}

@article{hafner2025dreamerv3,
  author  = {Hafner, Danijar and Pasukonis, Jurgis and Ba, Jimmy and Lillicrap, Timothy},
  title   = {{Mastering diverse control tasks through world models}},
  journal = {Nature},
  volume  = {640},
  pages   = {647--653},
  year    = {2025},
  doi     = {10.1038/s41586-025-08744-2},
  url     = {https://arxiv.org/abs/2301.04104}
}

@inproceedings{ho2020ddpm,
  author    = {Ho, Jonathan and Jain, Ajay N. and Abbeel, Pieter},
  title     = {{Denoising diffusion probabilistic models}},
  booktitle = {Advances in Neural Information Processing Systems},
  volume    = {33},
  pages     = {6840--6851},
  year      = {2020},
  url       = {https://proceedings.neurips.cc/paper/2020/hash/4c5bcfec8584af0d967f1ab10179ca4b-Abstract.html}
}

@article{hoffmann2022chinchilla,
  author  = {Hoffmann, Jordan and Borgeaud, Sebastian and Mensch, Arthur and Buchatskaya, Elena and Cai, Trevor and Rutherford, Eliza and de Las Casas, Diego and Hendricks, Lisa Anne and Welbl, Johannes and Clark, Aidan and others},
  title   = {{Training compute-optimal large language models}},
  journal = {arXiv preprint arXiv:2203.15556},
  year    = {2022},
  url     = {https://arxiv.org/abs/2203.15556}
}

@inproceedings{jiang2023vima,
  author    = {Jiang, Yunfan and Gupta, Agrim and Zhang, Zichen and Wang, Guanzhi and Dou, Yongqiang and Chen, Yanjun and Fei-Fei, Li and Anandkumar, Anima and Zhu, Yuke and Fan, Linxi},
  title     = {{VIMA}: Robot Manipulation with Multimodal Prompts},
  booktitle = {Proceedings of the 40th International Conference on Machine Learning},
  series    = {Proceedings of Machine Learning Research},
  volume    = {202},
  pages     = {14975--15022},
  year      = {2023},
  publisher = {PMLR},
  url       = {https://proceedings.mlr.press/v202/jiang23b.html}
}

@article{kaplan2020scaling,
  author  = {Kaplan, Jared and McCandlish, Sam and Henighan, Tom and Brown, Tom B. and Chess, Benjamin and Child, Rewon and Gray, Scott and Radford, Alec and Wu, Jeffrey and Amodei, Dario},
  title   = {{Scaling laws for neural language models}},
  journal = {arXiv preprint arXiv:2001.08361},
  year    = {2020},
  url     = {https://arxiv.org/abs/2001.08361}
}

@article{kim2026a_rldx,
  author  = {Kim, Dongyoung and Jang, Huiwon and Koo, Myungkyu and Jang, Suhyeok and Kim, Taeyoung and Kim, Beomjun and Yoon, Byungjun and Jang, Changsung and Choi, Daewon and Han, Dongsu and others},
  title   = {{RLDX-1 technical report}},
  journal = {arXiv preprint arXiv:2605.03269},
  year    = {2026},
  url     = {https://arxiv.org/abs/2605.03269}
}

@inproceedings{kim2025openvla,
  author    = {Kim, Moo Jin and Pertsch, Karl and Karamcheti, Siddharth and Xiao, Ted and Balakrishna, Ashwin and Nair, Suraj and Rafailov, Rafael and Foster, Ethan P. and Sanketi, Pannag R. and Vuong, Quan and Kollar, Thomas and Burchfiel, Benjamin and Tedrake, Russ and Sadigh, Dorsa and Levine, Sergey and Liang, Percy and Finn, Chelsea},
  title     = {{OpenVLA: An open-source vision-language-action model}},
  booktitle = {Proceedings of the 8th Conference on Robot Learning},
  series    = {Proceedings of Machine Learning Research},
  volume    = {270},
  pages     = {2679--2713},
  year      = {2025},
  url       = {https://proceedings.mlr.press/v270/kim25c.html}
}

@inproceedings{koo2026hamlet,
  author    = {Koo, Myungkyu and Choi, Daewon and Kim, Taeyoung and Lee, Kyungmin and Kim, Changyeon and Seo, Younggyo and Shin, Jinwoo},
  title     = {{HAMLET: Switch your vision-language-action model into a history-aware policy}},
  booktitle = {International Conference on Learning Representations},
  year      = {2026},
  url       = {https://arxiv.org/abs/2510.00695}
}

@misc{torne2026mem,
  author = {Torne, Marcel and Pertsch, Karl and Walke, Homer and Vedder, Kyle and Nair, Suraj and Ichter, Brian and Ren, Allen Z. and Wang, Haohuan and Tang, Jiaming and Stachowicz, Kyle and Dhabalia, Karan and Equi, Michael and Vuong, Quan and Springenberg, Jost Tobias and Levine, Sergey and Finn, Chelsea and Driess, Danny},
  title  = {{MEM: Multi-Scale Embodied Memory for Vision Language Action Models}},
  year   = {2026},
  url    = {https://www.pi.website/download/Mem.pdf}
}

@misc{lecun2022path,
  author       = {LeCun, Yann},
  title        = {{A path towards autonomous machine intelligence}},
  howpublished = {OpenReview position paper},
  year         = {2022},
  url          = {https://openreview.net/forum?id=BZ5a1r-kVsf}
}

@inproceedings{lin2025datascaling,
  author    = {Lin, Fanqi and Hu, Yingdong and Sheng, Pingyue and Wen, Chuan and You, Jiacheng and Gao, Yang},
  title     = {{Data scaling laws in imitation learning for robotic manipulation}},
  booktitle = {International Conference on Learning Representations},
  year      = {2025},
  url       = {https://arxiv.org/abs/2410.18647}
}

@inproceedings{mandlekar2022whatmatters,
  author    = {Mandlekar, Ajay and Xu, Danfei and Wong, Josiah and Nasiriany, Soroush and Wang, Chen and Kulkarni, Rohun and Fei-Fei, Li and Savarese, Silvio and Zhu, Yuke and Mart{\'i}n-Mart{\'i}n, Roberto},
  title     = {{What matters in learning from offline human demonstrations for robot manipulation}},
  booktitle = {Conference on Robot Learning},
  series    = {Proceedings of Machine Learning Research},
  volume    = {164},
  pages     = {1678--1690},
  year      = {2022},
  url       = {https://arxiv.org/abs/2108.03298},
  note      = {Conference held in 2021; introduces the robomimic framework}
}

@inproceedings{nasiriany2024robocasa,
  author    = {Nasiriany, Soroush and Maddukuri, Abhiram and Zhang, Lance and Parikh, Adeet and Lo, Aaron and Joshi, Abhishek and Mandlekar, Ajay and Zhu, Yuke},
  title     = {{RoboCasa: Large-scale simulation of everyday tasks for generalist robots}},
  booktitle = {Robotics: Science and Systems},
  year      = {2024},
  url       = {https://arxiv.org/abs/2406.02523}
}

@inproceedings{octo2024,
  author    = {Ghosh, Dibya and Walke, Homer Rich and Pertsch, Karl and Black, Kevin and Mees, Oier and Dasari, Sudeep and Hejna, Joey and Kreiman, Tobias and Xu, Charles and Luo, Jianlan and Tan, You Liang and Chen, Lawrence Yunliang and Vuong, Quan and Xiao, Ted and Sanketi, Pannag R. and Sadigh, Dorsa and Finn, Chelsea and Levine, Sergey},
  title     = {{Octo: An open-source generalist robot policy}},
  booktitle = {Robotics: Science and Systems},
  year      = {2024},
  doi       = {10.15607/RSS.2024.XX.090},
  url       = {https://arxiv.org/abs/2405.12213}
}

@inproceedings{openx2024,
  author    = {{Open X-Embodiment Collaboration} and others},
  title     = {{Open x-embodiment: Robotic learning datasets and RT-X models}},
  booktitle = {IEEE International Conference on Robotics and Automation},
  year      = {2024},
  url       = {https://arxiv.org/abs/2310.08864}
}

@inproceedings{pearce2025scaling,
  author    = {Pearce, Tim and Rashid, Tabish and Bignell, David and Georgescu, Raluca and Devlin, Sam and Hofmann, Katja},
  title     = {{Scaling laws for pre-training agents and world models}},
  booktitle = {International Conference on Machine Learning},
  series    = {Proceedings of Machine Learning Research},
  volume    = {267},
  pages     = {48542--48562},
  year      = {2025},
  url       = {https://arxiv.org/abs/2411.04434}
}

@inproceedings{peebles2023dit,
  author    = {Peebles, William and Xie, Saining},
  title     = {{Scalable diffusion models with Transformers}},
  booktitle = {IEEE/CVF International Conference on Computer Vision},
  pages     = {4195--4205},
  year      = {2023},
  url       = {https://openaccess.thecvf.com/content/ICCV2023/html/Peebles_Scalable_Diffusion_Models_with_Transformers_ICCV_2023_paper.html}
}

@inproceedings{radford2021clip,
  author    = {Radford, Alec and Kim, Jong Wook and Hallacy, Chris and Ramesh, Aditya and Goh, Gabriel and Agarwal, Sandhini and Sastry, Girish and Askell, Amanda and Mishkin, Pamela and Clark, Jack and Krueger, Gretchen and Sutskever, Ilya},
  title     = {{Learning transferable visual models from natural language supervision}},
  booktitle = {International Conference on Machine Learning},
  series    = {Proceedings of Machine Learning Research},
  volume    = {139},
  pages     = {8748--8763},
  year      = {2021},
  url       = {https://proceedings.mlr.press/v139/radford21a.html}
}

@article{reed2022gato,
  author  = {Reed, Scott and Zolna, Konrad and Parisotto, Emilio and Gomez Colmenarejo, Sergio and Novikov, Alexander and Barth-Maron, Gabriel and Gimenez, Mai and Sulsky, Yury and Kay, Jackie and Springenberg, Jost Tobias and Eccles, Tom and Bruce, Jake and Razavi, Ali and Edwards, Ashley and Heess, Nicolas and Chen, Yutian and Hadsell, Raia and Vinyals, Oriol and Bordbar, Mahyar and de Freitas, Nando},
  title   = {{A generalist agent}},
  journal = {Transactions on Machine Learning Research},
  year    = {2022},
  url     = {https://arxiv.org/abs/2205.06175}
}

@inproceedings{ross2011dagger,
  author    = {Ross, St{\'e}phane and Gordon, Geoffrey and Bagnell, Drew},
  title     = {{A reduction of imitation learning and structured prediction to no-regret online learning}},
  booktitle = {International Conference on Artificial Intelligence and Statistics},
  series    = {Proceedings of Machine Learning Research},
  volume    = {15},
  pages     = {627--635},
  year      = {2011},
  url       = {https://proceedings.mlr.press/v15/ross11a.html}
}

@article{shah2026aprism,
  author  = {Shah, Rutav and Jenamani, Rajat Kumar and Zhang, Xiaohan and Sun, Lingfeng and Mart{\'i}n-Mart{\'i}n, Roberto and Zhu, Yuke and Ramanan, Deva and Schmeckpeper, Karl},
  title   = {{Scaling short-term memory of visuomotor policies for long-horizon tasks}},
  journal = {arXiv preprint arXiv:2606.16178},
  year    = {2026},
  url     = {https://arxiv.org/abs/2606.16178}
}

@article{shah2026bhalo,
  author  = {Shah, Rutav and Li, Yisu and Bello, Femi and Zhu, Yuke and Mart{\'i}n-Mart{\'i}n, Roberto},
  title   = {{Memory retrieval in visuomotor policies for long-horizon robot control}},
  journal = {arXiv preprint arXiv:2606.25136},
  year    = {2026},
  url     = {https://arxiv.org/abs/2606.25136}
}

@inproceedings{shi2026memoryvla,
  author    = {Shi, Hao and Xie, Bin and Liu, Yingfei and Sun, Lin and Liu, Fengrong and Wang, Tiancai and Zhou, Erjin and Fan, Haoqiang and Zhang, Xiangyu and Huang, Gao},
  title     = {{MemoryVLA: Perceptual-cognitive memory in vision-language-action models for robotic manipulation}},
  booktitle = {International Conference on Learning Representations},
  year      = {2026},
  url       = {https://arxiv.org/abs/2508.19236}
}

@article{su2021roformer,
  author  = {Su, Jianlin and Lu, Yu and Pan, Shengfeng and Murtadha, Ahmed and Wen, Bo and Liu, Yunfeng},
  title   = {{RoFormer: Enhanced Transformer with rotary position embedding}},
  journal = {arXiv preprint arXiv:2104.09864},
  year    = {2021},
  url     = {https://arxiv.org/abs/2104.09864}
}

@article{vujinovic2026actjepa,
  author  = {Vujinovic, Aleksandar and Kovacevic, Aleksandar},
  title   = {{ACT-JEPA: Novel joint-embedding predictive architecture for efficient policy representation learning}},
  journal = {IEEE Access},
  volume  = {14},
  pages   = {78895--78906},
  year    = {2026},
  doi     = {10.1109/ACCESS.2026.3696039},
  url     = {https://arxiv.org/abs/2501.14622}
}

@inproceedings{wu2024gr1,
  author    = {Wu, Hongtao and Jing, Ya and Cheang, Chilam and Chen, Guangzeng and Xu, Jiafeng and Li, Xinghang and Liu, Minghuan and Li, Hang and Kong, Tao},
  title     = {{Unleashing large-scale video generative pre-training for visual robot manipulation}},
  booktitle = {International Conference on Learning Representations},
  year      = {2024},
  url       = {https://arxiv.org/abs/2312.13139}
}

@article{yang2024qwen2,
  author  = {Yang, An and Yang, Baosong and Hui, Binyuan and Zheng, Bo and Yu, Bowen and Zhou, Chang and Li, Chengpeng and Li, Chengyuan and Liu, Dayiheng and Huang, Fei and others},
  title   = {{Qwen2 technical report}},
  journal = {arXiv preprint arXiv:2407.10671},
  year    = {2024},
  url     = {https://arxiv.org/abs/2407.10671}
}

@inproceedings{zhao2023act,
  author    = {Zhao, Tony Z. and Kumar, Vikash and Levine, Sergey and Finn, Chelsea},
  title     = {{Learning fine-grained bimanual manipulation with low-cost hardware}},
  booktitle = {Robotics: Science and Systems},
  year      = {2023},
  url       = {https://arxiv.org/abs/2304.13705}
}

@article{zhou2026chronos,
  author  = {Zhou, Yulin and Wang, Yimeng and Wang, Nengyu and Xing, Shaojia and Tu, Shiyun and Li, Xiang and Zhang, Jingkai and Jiang, Ningbo and Lin, Yuankai and Yang, Hua and others},
  title   = {{Chronos: A physics-informed full-history framework for non-markovian long-horizon manipulation}},
  journal = {arXiv preprint arXiv:2606.30318},
  year    = {2026},
  url     = {https://arxiv.org/abs/2606.30318},
  note    = {Submitted to IEEE Transactions on Robotics}
}

@article{zhu2024scaledp,
  author  = {Zhu, Minjie and Zhu, Yichen and Li, Jinming and Wen, Junjie and Xu, Zhiyuan and Liu, Ning and Cheng, Ran and Shen, Chaomin and Peng, Yaxin and Feng, Feifei and Tang, Jian},
  title   = {{Scaling diffusion policy in transformer to 1 billion parameters for robotic manipulation}},
  journal = {arXiv preprint arXiv:2409.14411},
  year    = {2024},
  url     = {https://arxiv.org/abs/2409.14411}
}

@inproceedings{liu2023visual,
  author    = {Liu, Haotian and Li, Chunyuan and Wu, Qingyang and Lee, Yong Jae},
  title     = {{Visual Instruction Tuning}},
  booktitle = {Advances in Neural Information Processing Systems},
  volume    = {36},
  year      = {2023},
  url       = {https://proceedings.neurips.cc/paper_files/paper/2023/hash/6dcf277ea32ce3288914faf369fe6de0-Abstract-Conference.html}
}

@inproceedings{janner2021trajectory,
  author    = {Janner, Michael and Li, Qiyang and Levine, Sergey},
  title     = {{Offline Reinforcement Learning as One Big Sequence Modeling Problem}},
  booktitle = {Advances in Neural Information Processing Systems},
  volume    = {34},
  year      = {2021},
  url       = {https://arxiv.org/abs/2106.02039}
}

@inproceedings{zheng2022online,
  author    = {Zheng, Qinqing and Zhang, Amy and Grover, Aditya},
  title     = {{Online Decision Transformer}},
  booktitle = {Proceedings of the 39th International Conference on Machine Learning},
  series    = {Proceedings of Machine Learning Research},
  volume    = {162},
  pages     = {27042--27059},
  year      = {2022},
  publisher = {PMLR},
  url       = {https://proceedings.mlr.press/v162/zheng22c.html}
}

@inproceedings{wang2024hpt,
  author    = {Wang, Lirui and Chen, Xinlei and Zhao, Jialiang and He, Kaiming},
  title     = {{Scaling Proprioceptive-Visual Learning with Heterogeneous Pre-trained Transformers}},
  booktitle = {Advances in Neural Information Processing Systems},
  volume    = {37},
  year      = {2024},
  url       = {https://arxiv.org/abs/2409.20537}
}

@article{li2024cogact,
  author  = {Li, Qixiu and Liang, Yaobo and Wang, Zeyu and Luo, Lin and Chen, Xi and Liao, Mozheng and Wei, Fangyun and Deng, Yu and Xu, Sicheng and Zhang, Yizhong and others},
  title   = {{CogACT: A Foundational Vision-Language-Action Model for Synergizing Cognition and Action in Robotic Manipulation}},
  journal = {arXiv preprint arXiv:2411.19650},
  year    = {2024},
  url     = {https://arxiv.org/abs/2411.19650}
}

@inproceedings{li2024roboflamingo,
  author    = {Li, Xinghang and Liu, Minghuan and Zhang, Hanbo and Yu, Cunjun and Xu, Jie and Wu, Hongtao and Cheang, Chilam and Jing, Ya and Zhang, Weinan and Liu, Huaping and Li, Hang and Kong, Tao},
  title     = {{Vision-Language Foundation Models as Effective Robot Imitators}},
  booktitle = {International Conference on Learning Representations},
  year      = {2024},
  url       = {https://arxiv.org/abs/2311.01378}
}

@article{wang2026nativemem,
  author  = {Wang, Ziye and Shi, Modi and Ni, Chaojun and Yang, Jiazhi and Li, Mengdi and Su, Zhizhong and Lin, Tianwei and Li, Hongyang},
  title   = {{NativeMEM: Native Memory Compression for Long-Horizon Robotic Manipulation}},
  journal = {arXiv preprint arXiv:2607.06678},
  year    = {2026},
  url     = {https://arxiv.org/abs/2607.06678}
}

@inproceedings{zheng2025tracevla,
  author    = {Zheng, Ruijie and Liang, Yongyuan and Huang, Shuaiyi and Gao, Jianfeng and Daum{\'e}, III, Hal and Kolobov, Andrey and Huang, Furong and Yang, Jianwei},
  title     = {{TraceVLA: Visual Trace Prompting Enhances Spatial-Temporal Awareness for Generalist Robotic Policies}},
  booktitle = {International Conference on Learning Representations},
  year      = {2025},
  url       = {https://arxiv.org/abs/2412.10345}
}
\bibliographystyle{iclr2027_conference}

\clearpage
\addtocontents{toc}{\protect\setcounter{tocdepth}{2}}
\appendix
\raggedbottom
\makeatletter
\setlength{\@fptop}{0pt}
\makeatother
\begingroup
\renewcommand{\contentsname}{Appendix contents}
\setcounter{tocdepth}{-1}
\tableofcontents
\endgroup
\clearpage
\section{Detailed WorldToken Architecture}
\label{app:method-details}

This appendix details the attention mask, history window, and diffusion objective of the three modules in Section~\ref{sec:method}.

\subsection{Within-Timestep Multimodal Encoding}
\label{app:method-encoder}

The within-timestep encoder fuses multiview images, proprioception, and the task condition into a fixed-width temporal representation $z_t$.

For each camera, a visual stem maps $I_t^i$ to visual patch tokens, which are augmented with embeddings that identify spatial position, camera view, and modality before multimodal fusion. If camera $i$ produces $M_i$ tokens,
\begin{equation}
V_t^i = \bigl(v_{t,1}^i, \ldots, v_{t,M_i}^i\bigr).
\label{eq:visual}
\end{equation}
Proprioception and task conditioning are projected into $u_t^{\mathrm{prop}}$ and $u_t^{\mathrm{task}}$. The within-timestep sequence is
\begin{equation}
X_t = \bigl[V_t^1; \ldots; V_t^{K_{\mathrm{view}}}; u_t^{\mathrm{prop}}; u_t^{\mathrm{task}}\bigr].
\label{eq:xt}
\end{equation}
We append $R$ learned readout tokens, denoted by $Q$, with $R=4$ by default. Their output states are concatenated and projected into one temporal token. As shown in Figure~\ref{fig:architecture}(a), observation tokens attend only to observation tokens, while readout tokens attend to both observation and readout tokens. The readouts therefore aggregate observation features without feeding information back to the observation tokens.

Let $r_t^1, \ldots, r_t^R$ denote the final outputs at the readout positions:
\begin{equation}
(r_t^1, \ldots, r_t^R) = F_{\mathrm{enc}}([X_t; Q])_{\mathrm{readout}}.
\label{eq:readout}
\end{equation}
Concatenation, RMSNorm, and a linear projection produce the only token passed to the temporal backbone:
\begin{equation}
z_t = W_z\,\mathrm{RMSNorm}([r_t^1; \ldots; r_t^R]).
\label{eq:zt}
\end{equation}

\subsection{Temporal Backbone and History Window}
\label{app:method-backbone}

We define the world-token trajectory as
\begin{equation}
Z_{1:T} = (z_1, z_2, \ldots, z_T).
\label{eq:ztraj}
\end{equation}
The causal temporal backbone $T_\phi$ uses the Qwen2 decoder architecture \citep{yang2024qwen2} to produce history-conditioned states:
\begin{equation}
(h_1, \ldots, h_T) = T_\phi(z_1, \ldots, z_T), \qquad h_t = T_\phi(z_{\le t})_t.
\label{eq:ht}
\end{equation}
Equation~\eqref{eq:ht} gives the full-prefix form; finite-window evaluation uses $z_{s_t:t}$ as in Eq.~\eqref{eq:decomp}. One-dimensional RoPE \citep{su2021roformer} is applied along the world-token sequence to encode policy time. Causal self-attention ensures that $h_t$ depends only on $z_{\le t}$.

Within each episode, each policy timestep adds one world token to the history. The action decoder reads only the last valid backbone output $h_t$.

Let $C$ be the maximum visible-history length. At policy timestep $t$ the policy reads
\begin{equation}
z_{\max(1,t-C+1):t}.
\label{eq:window}
\end{equation}
If $C$ covers the episode so far, the policy receives full-episode context; once the episode exceeds $C$, the earliest tokens are evicted.

When the window slides, we reindex the retained tokens contiguously and recompute the backbone outputs. The backbone has no absolute position embeddings. Subtracting the same offset from all retained positions preserves the relative position differences used by standard RoPE attention.

\subsection{DiT Action Head}
\label{app:method-decoder}

At policy timestep $t$, WorldToken generates an $H$-step action chunk
\begin{equation}
\hat{A}_t = (\hat{a}_{t,0}, \hat{a}_{t,1}, \ldots, \hat{a}_{t,H-1}) \in \R^{H \times d_a}.
\label{eq:chunk}
\end{equation}
Each action dimension is normalized to $[-1,1]$ using statistics from the training demonstrations and unnormalized before execution.

We model the conditional action distribution with diffusion. For a normalized ground-truth chunk $A_t$, training samples diffusion step $k$ and Gaussian noise $\epsilon$:
\begin{equation}
A_t^{(k)} =
\sqrt{\bar{\alpha}_k}\,A_t
+
\sqrt{1-\bar{\alpha}_k}\,\epsilon,
\qquad
\epsilon \sim \mathcal{N}(0,I).
\label{eq:noisy}
\end{equation}
Here $\bar{\alpha}_k$ is the cumulative product of the diffusion schedule's per-step signal-retention factors. A DiT denoiser \citep{peebles2023dit} predicts the injected noise conditioned on the current history representation $h_t$:
\begin{equation}
\hat{\epsilon}
=
\epsilon_\psi\bigl(A_t^{(k)}, k, h_t\bigr),
\qquad
L_{\mathrm{act}}
=
\E_{t,k,\epsilon}
\left[
\left\|
\epsilon -
\epsilon_\psi\bigl(A_t^{(k)}, k, h_t\bigr)
\right\|_2^2
\right].
\label{eq:lact}
\end{equation}
Equation~\eqref{eq:lact} is the default diffusion objective; Appendix~\ref{app:ranking-objective} specifies the task-specific modification for Blocks Ranking. The diffusion timestep and $h_t$ condition each DiT block through adaLN-Zero modulation. At inference, action chunks are generated from Gaussian noise using stochastic DDPM sampling \citep{ho2020ddpm}.

Control follows a receding-horizon scheme. Each query generates $H$ actions but executes only the first $H_{\mathrm{exec}}$:
\begin{equation}
\hat{A}_t \sim \mathcal{D}_\psi(\cdot \mid h_t),
\qquad
\hat{A}^{\mathrm{exec}}_t
=
(\hat{a}_{t,0},\ldots,\hat{a}_{t,H_{\mathrm{exec}}-1}).
\label{eq:receding}
\end{equation}
After executing this prefix, the policy observes the environment again, appends the next world token, and replans.

\section{Detailed Experimental Setup for RoboCasa}
\label{app:protocol}

This appendix specifies the RoboCasa data, models, training, and evaluation. Settings specific to the token and history comparisons are given in Appendices~\ref{app:granularity} and~\ref{app:robocasa-history}.

\subsection{Tasks and Datasets}
The RoboCasa sweep uses a fixed set of 23 tasks. We exclude \texttt{OpenDoubleDoor} because its official image-based generated-demonstration dataset has only 1,500 demonstrations, fewer than the common $D=2900$ endpoint. For each retained task, we reserve 100 demonstrations for a shared holdout set, giving 2,300 holdout demonstrations in total. None appear in any training set.

The $D=300$ training set is the official \texttt{300\_demos} subset. The $D\in\{50,100,1000\}$ sets are independently fixed, hash-recorded samples from the non-holdout pool; they may overlap but are not nested. The $D=2900$ set uses the full non-holdout pool. At each $D$, all model capacities and training seeds use the same demonstrations.

\subsection{Model Configurations and Training}
\label{app:robocasa-training}

Each data scale is trained for approximately 100 loader epochs, using the optimizer-step budgets in Table~\ref{tab:optimizer-steps}. Each epoch visits eight sample slots per training demonstration, drawing a random ten-timestep observation window for each slot. This samples windows rather than enumerating all valid windows. We evaluate the final scheduled checkpoint from every run, without rollout-based checkpoint selection. Appendix~\ref{app:optimization-audit} reports holdout RMSE at 10\% intervals of the training budget.

All training and evaluation used two single-node servers: one with eight NVIDIA A100-80GB GPUs and one with eight NVIDIA H100-80GB GPUs.

Table~\ref{tab:model-configs} specifies the five model configurations, including the within-timestep fusion encoder, causal temporal backbone, and DiT action decoder. Each model uses a separate CNN stem for each camera and the frozen \texttt{openai/clip-vit-large-patch14} text tower with projection \citep{radford2021clip} for task conditioning. The shared visual and training settings appear in Table~\ref{tab:protocol}. The learning rates in Table~\ref{tab:model-configs} were selected at $D=300$ and reused across the data scales for each capacity.

\begin{table}[h!]
\centering
\caption{RoboCasa N1--N5 architecture and peak learning rates. Parameters include all visual stems and policy modules, excluding the frozen CLIP text encoder. FF denotes feed-forward hidden width; Q and KV denote query and key/value attention heads.}
\label{tab:model-configs}
\small
\setlength{\tabcolsep}{4pt}
\renewcommand{\arraystretch}{1.05}
\begin{tabular*}{\linewidth}{@{\extracolsep{\fill}}lrrrrr@{}}
\toprule
Setting & N1 & N2 & N3 & N4 & N5 \\
\midrule
Parameters & 44.3M & 85.3M & 218.8M & 648.9M & 1490.3M \\
Encoder/temporal LR ($10^{-4}$) & 6.00 & 4.25 & 3.00 & 1.50 & 0.50 \\
Action-decoder LR ($10^{-4}$) & 3.00 & 3.00 & 3.00 & 3.00 & 3.00 \\
\midrule
\multicolumn{6}{@{}l}{\textit{Within-timestep fusion encoder (SwiGLU)}} \\
Width & 512 & 768 & 1024 & 1536 & 2048 \\
Layers & 1 & 2 & 4 & 6 & 8 \\
Q heads & 4 & 6 & 8 & 12 & 16 \\
KV heads & 4 & 6 & 8 & 12 & 16 \\
FFN width & 2048 & 3072 & 4096 & 6144 & 8192 \\
\midrule
\multicolumn{6}{@{}l}{\textit{Causal temporal backbone (Qwen2, SwiGLU)}} \\
Width & 512 & 768 & 1024 & 1536 & 2048 \\
Layers & 2 & 4 & 8 & 12 & 16 \\
Q heads & 4 & 6 & 8 & 12 & 16 \\
KV heads & 4 & 6 & 8 & 12 & 16 \\
FFN width & 1408 & 2048 & 2816 & 4096 & 5504 \\
\midrule
\multicolumn{6}{@{}l}{\textit{DiT action decoder (GELU, adaLN-Zero)}} \\
Width & 128 & 192 & 256 & 384 & 512 \\
Layers & 1 & 2 & 4 & 6 & 8 \\
Attention heads & 4 & 6 & 8 & 12 & 16 \\
FFN width & 512 & 768 & 1024 & 1536 & 2048 \\
\bottomrule
\end{tabular*}
\end{table}

\begin{table}[h!]
\centering
\caption{RoboCasa training and closed-loop execution recipe. The main sweep uses $C_{\mathrm{train}} = 10$; separately trained context variants are identified where they are analyzed.}
\label{tab:protocol}
\label{tab:optimizer-steps}
\small
\begin{tabular}{@{}>{\raggedright\arraybackslash}p{0.22\linewidth}>{\raggedright\arraybackslash}p{0.72\linewidth}@{}}
\toprule
Component & Protocol \\
\midrule
Inputs & Three $128 \times 128$ RGB views, 16-D proprioception, and a frozen 768-D CLIP task embedding \\
Visual stems & One five-stage CNN per camera; $5\times5$ convolutions with channels $(96,144,192,384,768)$, each followed by $2\times2$ max pooling, group normalization, and SiLU; output grid $4\times4\times768$ per view \\
Token readout & 48 visual tokens plus proprioception and task tokens; four learned readouts are concatenated, RMS-normalized, and projected to one temporal world token \\
Temporal sampling & 20~Hz environment control; one observation, world token, and replanning decision every four control steps (5~Hz) \\
Actions & Same-frame-aligned 12-D commands; $H = 10$ predicted, $H_{\mathrm{exec}} = 4$ executed before replanning \\
Context and diffusion & $C_{\mathrm{train}} = 10$ world tokens; 20-stage cosine schedule and stochastic 20-step DDPM evaluation \\
Optimization & AdamW, $\beta = (0.9, 0.95)$, $\epsilon = 10^{-8}$, zero weight decay, global gradient-norm clipping at 1.0, BF16, and global batch size 192 observation windows \\
Training budget & $D=50/100/300/1000/2900$: 5k/10k/30k/100k/280k optimizer steps, respectively \\
LR schedule & Encoder/temporal rates in Table~\ref{tab:model-configs}; action-decoder rate $3.0 \times 10^{-4}$; 5\% linear warmup, then cosine decay to 10\% of each peak rate \\
\bottomrule
\end{tabular}
\end{table}

Width denotes the hidden dimension $d$ of each module.
The feed-forward network (FFN) width denotes its intermediate dimension.
The fusion encoder and action decoder use an FFN width of $4d$.
The temporal backbone uses $8d/3$, rounded up to a multiple of 128.
Temporal attention dropout is 0.1.
Dropout is disabled in the fusion encoder and action decoder.

\subsection{Experimental Configurations}
\label{app:robocasa-configurations}

Table~\ref{tab:robocasa-experiment-design} lists training seeds separately from repeated evaluations of a fixed checkpoint. Comparisons share demonstration identities, holdout crops, and evaluation episode identities where these are specified as controlled.

\begin{table}[!htbp]
\centering
\caption{RoboCasa experiment configurations. ``Executions'' counts complete 1,150-episode evaluations per checkpoint and stated inference setting, not independent training seeds.}
\label{tab:robocasa-experiment-design}
\scriptsize
\setlength{\tabcolsep}{3pt}
\begin{tabular}{@{}p{0.21\linewidth}p{0.48\linewidth}cc@{}}
\toprule
Experiment & Changed / held fixed & Training seeds & Executions \\
\midrule
Data and model scaling & Five $D$ values $\times$ five capacities; $C_{\rm train}=C_{\rm test}=10$ & 0, 1 & 3 \\
Token count & $K=1,4,50$ at all five $D$; $d=768$, $C=10$ & 0 & 3 \\
Same-checkpoint history & $C_{\rm test}=1,2,5$ for every scaling checkpoint; $C_{\rm train}=10$ & 0, 1 & 1 \\
Matched training history & $C_{\rm train}=C_{\rm test}=1,2,5,10$; 218.8M, $D=300$ & 0, 1 & 3 \\
\bottomrule
\end{tabular}
\end{table}

\subsection{Evaluation Details}
\label{app:robocasa-evaluation}

\paragraph{Closed-loop evaluation.}
A low-level control step is one environment command, distinct from a simulator integration step. Each closed-loop evaluation contains 50 episodes for each of the same 23 tasks, or 1,150 episodes in total. At episode startup, we left-pad the history with copies of the first observation to reach $C_{\rm test}$ positions. Only RoboCasa uses this padding; RMBench does not.

\paragraph{Repeated evaluations and reporting.}
Each main-sweep final checkpoint is evaluated three times at $C_{\rm test}=10$, using the same episode identities, environment seeds, rollout seed, and protocol. Reported standard deviations are sample standard deviations across the stated repeats. Because stochastic diffusion and simulation are not bitwise deterministic, these values measure variation across fixed-seed executions, not uncertainty over independently sampled environments. To average across training seeds, we first average each checkpoint's repeated evaluations and then give each trained policy equal weight.

\paragraph{Action alignment.}
Let $j_t$ be the stored control-frame index for policy timestep $t$. Policy timesteps are four control frames apart, but each target chunk contains $H$ consecutive low-level commands: \texttt{action[$j_t$:$j_t+H$]}. The first command is applied after conditioning on the observation at $j_t$. Padded terminal slots repeat the final action and are masked. A query contributes to the loss only when all $H$ targets are real, so no padded terminal targets enter training. Appendix~\ref{app:ranking-training} gives the RMBench indexing convention.

\paragraph{Holdout RMSE.}
For the main sweep, \texttt{eval\_seed=0} selects eight fixed ten-timestep crops from each of the 2,300 holdout demonstrations. Evaluation uses these crops rather than all possible trajectory windows. Separately trained context variants use the same procedure with crops of their own training length $C$. At each query with a complete target, we evaluate one $H=10$ action chunk. Query $i$ sees only the first $i$ observations of its crop, so the first two queries see one and two observations, respectively. No query reads observations before the crop begins. We compute RMSE in the unnormalized 12-D command space and average the 23 taskwise RMSE values equally.

Deterministic and stochastic DDPM evaluations start from the same fixed-seed Gaussian $x_T$ for each query. The deterministic sampler follows the DDPM posterior mean without reverse-step noise. The stochastic sampler adds Gaussian noise with the variance specified by the DDPM schedule.

\paragraph{Observation fields and action normalization.}
The RGB views are ordered as left agent view, right agent view, and eye-in-hand. Proprioception concatenates end-effector position (3) and quaternion (4) relative to the base, base position (3) and quaternion (4), and gripper joint positions (2), preserving the stored coordinate order. With zero-based action indices, 0--2 are arm-translation commands, 3--5 arm-rotation commands, 6 the gripper command, 7--10 the base/torso commands, and 11 the base-mode command. These controller inputs have different meanings, so the full 12-D RMSE has no single physical unit.

For each action dimension, the normalizer scans all actions in the selected training demonstrations once. It sets $\ell=(a_{\max}+a_{\min})/2$ and $r=(a_{\max}-a_{\min})/2$, then applies $(a-\ell)/r$, replacing $r$ with 1 when $r<10^{-4}$. These fixed statistics are stored in the checkpoint. Generated actions are transformed back before RMSE computation and execution.

\paragraph{Rollout settings.}
Task horizons are measured in low-level control steps: 300 for \texttt{CoffeePressButton}; 600 for \texttt{CoffeeServeMug}, \texttt{CoffeeSetupMug}, and \texttt{PnPCounterToMicrowave}; 700 for \texttt{CloseDoubleDoor} and \texttt{PnPCounterToSink}; and 500 for each of the other 17 tasks. The reporting runtime uses robosuite 1.5.0 at commit \texttt{dc7fcf9fa6cd}, environment-bound action clipping with margin $10^{-4}$, and unit action scale.

\section{Data and Model Scaling on RoboCasa}
\label{app:robocasa-results}

This appendix provides detailed results supporting the RoboCasa analysis in Section~\ref{sec:robocasa}, including the full 50-policy sweep, holdout RMSE throughout training, task-level success rates, and baseline comparisons. These results allow the aggregate trends reported in the main text to be examined across training seeds, repeated evaluations, training progress, and individual tasks. We also document the settings and limitations of the baseline comparisons to clarify how their results should be interpreted.

\subsection{Success Rates and RMSE for All 50 Policies}
\label{app:complete-sr-grids}
\begin{table}[!htbp]
\centering
\caption{Closed-loop SR (\%) for all 50 scaling policies trained with $C_{\rm train}=10$. Each checkpoint is evaluated once at $h=1,2,5$ and three times at $h=10$, where $h=C_{\rm test}$ counts policy timesteps. Each evaluation contains 1,150 episodes; mean $\pm$ sample SD is reported only for $h=10$. Final-checkpoint stochastic full-chunk RMSE follows Appendix~\ref{app:robocasa-evaluation}. N1--N5 are defined in Table~\ref{tab:model-configs}.}
\label{tab:scaling-complete}
\label{tab:scaling-sr-seed0}
\label{tab:scaling-sr-seed1}
\label{tab:history-complete}
\label{fig:app-history-absolute}
\label{tab:truncation-sr-h1}
\label{tab:truncation-sr-h2}
\label{tab:truncation-sr-h5}
\small
\setlength{\tabcolsep}{2.2pt}
\renewcommand{\arraystretch}{1.12}
\renewcommand{\arraystretch}{0.92}
\begin{tabular*}{\linewidth}{@{\extracolsep{\fill}}rlrrrrrrrrr@{}}
\toprule
& & & & \multicolumn{3}{c}{Single execution} & \multicolumn{3}{c}{$h=10$ repeats} & $h=10$ \\
\cmidrule(lr){5-7}\cmidrule(lr){8-10}
$D$ & Model & Seed & RMSE & $h=1$ & $h=2$ & $h=5$ & 1 & 2 & 3 & Mean $\pm$ SD \\
\midrule
50 & N1 & 0 & 0.20765 & 10.61 & 10.00 & 13.22 & 14.52 & 13.91 & 14.00 & $14.14\pm0.33$ \\
 &  & 1 & 0.20301 & 12.52 & 15.57 & 18.78 & 18.17 & 18.96 & 17.30 & $18.14\pm0.83$ \\
 & N2 & 0 & 0.19514 & 14.26 & 16.09 & 21.65 & 20.96 & 20.96 & 22.26 & $21.39\pm0.75$ \\
 &  & 1 & 0.19682 & 10.96 & 13.57 & 18.09 & 20.35 & 17.65 & 18.78 & $18.93\pm1.35$ \\
 & N3 & 0 & 0.19532 & 14.70 & 17.57 & 22.26 & 22.70 & 22.00 & 21.22 & $21.97\pm0.74$ \\
 &  & 1 & 0.19411 & 14.35 & 15.13 & 20.61 & 23.39 & 22.70 & 22.09 & $22.72\pm0.65$ \\
 & N4 & 0 & 0.19279 & 16.61 & 18.26 & 25.74 & 27.13 & 28.17 & 26.09 & $27.13\pm1.04$ \\
 &  & 1 & 0.19518 & 16.43 & 15.04 & 23.48 & 24.17 & 23.74 & 24.17 & $24.03\pm0.25$ \\
 & N5 & 0 & 0.18934 & 16.70 & 17.39 & 23.57 & 22.52 & 23.22 & 23.48 & $23.07\pm0.49$ \\
 &  & 1 & 0.18610 & 14.78 & 16.09 & 19.39 & 21.48 & 22.00 & 21.48 & $21.65\pm0.30$ \\
\midrule
100 & N1 & 0 & 0.17477 & 19.83 & 26.70 & 30.35 & 29.48 & 29.57 & 30.87 & $29.97\pm0.78$ \\
 &  & 1 & 0.17701 & 21.65 & 22.52 & 27.13 & 27.30 & 26.87 & 27.91 & $27.36\pm0.52$ \\
 & N2 & 0 & 0.17212 & 22.52 & 24.52 & 31.13 & 30.17 & 31.74 & 31.65 & $31.19\pm0.88$ \\
 &  & 1 & 0.16947 & 21.65 & 24.96 & 32.35 & 34.43 & 31.83 & 30.87 & $32.38\pm1.85$ \\
 & N3 & 0 & 0.16618 & 24.70 & 25.91 & 35.48 & 37.30 & 34.87 & 35.30 & $35.83\pm1.30$ \\
 &  & 1 & 0.16491 & 21.74 & 25.91 & 33.74 & 31.91 & 34.00 & 33.83 & $33.25\pm1.16$ \\
 & N4 & 0 & 0.16913 & 21.48 & 25.13 & 34.61 & 36.09 & 36.35 & 36.96 & $36.46\pm0.45$ \\
 &  & 1 & 0.17028 & 17.39 & 25.04 & 34.61 & 36.43 & 35.74 & 36.70 & $36.29\pm0.49$ \\
 & N5 & 0 & 0.17117 & 20.09 & 25.22 & 33.30 & 33.48 & 33.30 & 34.35 & $33.71\pm0.56$ \\
 &  & 1 & 0.17098 & 19.91 & 24.43 & 34.61 & 33.13 & 34.35 & 36.35 & $34.61\pm1.62$ \\
\midrule
300 & N1 & 0 & 0.15265 & 22.96 & 30.09 & 40.52 & 41.04 & 40.87 & 41.22 & $41.04\pm0.17$ \\
 &  & 1 & 0.15193 & 30.96 & 39.91 & 42.17 & 41.13 & 41.74 & 41.22 & $41.36\pm0.33$ \\
 & N2 & 0 & 0.13680 & 29.65 & 35.22 & 46.35 & 48.00 & 46.17 & 48.26 & $47.48\pm1.14$ \\
 &  & 1 & 0.13544 & 29.57 & 36.43 & 46.35 & 46.52 & 46.35 & 45.65 & $46.17\pm0.46$ \\
 & N3 & 0 & 0.13171 & 27.91 & 34.96 & 47.04 & 46.70 & 50.00 & 47.65 & $48.12\pm1.70$ \\
 &  & 1 & 0.13310 & 27.83 & 35.74 & 48.17 & 48.87 & 49.13 & 47.13 & $48.38\pm1.09$ \\
 & N4 & 0 & 0.13550 & 29.91 & 39.30 & 49.91 & 50.78 & 51.22 & 51.74 & $51.25\pm0.48$ \\
 &  & 1 & 0.13487 & 29.30 & 38.17 & 49.91 & 50.52 & 51.91 & 51.74 & $51.39\pm0.76$ \\
 & N5 & 0 & 0.14404 & 30.17 & 37.13 & 48.35 & 50.00 & 49.39 & 50.09 & $49.83\pm0.38$ \\
 &  & 1 & 0.14376 & 28.35 & 34.78 & 48.61 & 49.83 & 49.83 & 49.91 & $49.86\pm0.05$ \\
\midrule
1000 & N1 & 0 & 0.12933 & 34.17 & 40.52 & 48.96 & 48.96 & 49.91 & 48.43 & $49.10\pm0.75$ \\
 &  & 1 & 0.12888 & 35.83 & 42.43 & 49.74 & 50.26 & 49.91 & 49.39 & $49.86\pm0.44$ \\
 & N2 & 0 & 0.10873 & 37.65 & 49.22 & 56.43 & 55.83 & 56.35 & 54.87 & $55.68\pm0.75$ \\
 &  & 1 & 0.10998 & 34.43 & 45.57 & 52.70 & 52.87 & 52.78 & 50.96 & $52.20\pm1.08$ \\
 & N3 & 0 & 0.09937 & 38.17 & 46.78 & 57.91 & 59.39 & 58.26 & 57.74 & $58.46\pm0.84$ \\
 &  & 1 & 0.10051 & 37.57 & 47.74 & 57.22 & 58.35 & 56.96 & 58.09 & $57.80\pm0.74$ \\
 & N4 & 0 & 0.10117 & 38.78 & 47.57 & 57.04 & 56.78 & 58.43 & 57.74 & $57.65\pm0.83$ \\
 &  & 1 & 0.10044 & 37.65 & 46.70 & 57.74 & 56.78 & 56.26 & 57.04 & $56.70\pm0.40$ \\
 & N5 & 0 & 0.10853 & 38.87 & 46.78 & 57.74 & 56.43 & 56.70 & 55.48 & $56.20\pm0.64$ \\
 &  & 1 & 0.10806 & 37.57 & 47.04 & 57.48 & 56.87 & 56.52 & 55.91 & $56.43\pm0.48$ \\
\midrule
2900 & N1 & 0 & 0.10905 & 41.39 & 47.57 & 53.83 & 55.04 & 55.48 & 52.70 & $54.41\pm1.50$ \\
 &  & 1 & 0.10851 & 40.61 & 49.04 & 55.22 & 54.09 & 56.00 & 53.91 & $54.67\pm1.16$ \\
 & N2 & 0 & 0.09083 & 43.74 & 52.87 & 60.61 & 58.96 & 58.17 & 60.09 & $59.07\pm0.96$ \\
 &  & 1 & 0.09069 & 42.61 & 53.22 & 58.70 & 59.83 & 59.74 & 59.91 & $59.83\pm0.09$ \\
 & N3 & 0 & 0.08363 & 40.43 & 53.74 & 59.30 & 61.04 & 56.96 & 60.00 & $59.33\pm2.12$ \\
 &  & 1 & 0.08447 & 42.00 & 54.00 & 59.13 & 61.74 & 62.00 & 59.83 & $61.19\pm1.19$ \\
 & N4 & 0 & 0.08370 & 36.61 & 49.65 & 57.83 & 57.65 & 60.52 & 58.26 & $58.81\pm1.51$ \\
 &  & 1 & 0.08436 & 38.87 & 52.00 & 59.74 & 58.09 & 57.83 & 57.48 & $57.80\pm0.31$ \\
 & N5 & 0 & 0.08662 & 41.04 & 50.87 & 58.70 & 59.48 & 60.43 & 59.04 & $59.65\pm0.71$ \\
 &  & 1 & 0.08615 & 40.43 & 48.61 & 59.39 & 58.61 & 60.87 & 58.35 & $59.28\pm1.39$ \\
\bottomrule
\end{tabular*}
\end{table}

\paragraph{Descriptive power-law fits.}
For each capacity, the five data-scale points are the arithmetic means of the two training-seed RMSE values. We fit $\log(\mathrm{RMSE})=b-\alpha\log D$ by unweighted ordinary least squares over these five points and compute $R^2$ in the same log-RMSE space.

\clearpage
\subsection{RMSE at 10\% Training Intervals}
\label{app:optimization-audit}
The final checkpoints lie in the flatter part of the recorded training curves.
Across all 50 runs, the relative RMSE reduction over the last 20\% of the
scheduled budget (80\% to 100\%) has a median of 0.53\% and a maximum of
2.21\%. Over the final 10\% (90\% to 100\%), the median is 0.08\%.
\begin{table}[!htbp]
\centering
\caption{RoboCasa holdout stochastic full-chunk RMSE at 10\% training intervals, seed 0, using the metric in Appendix~\ref{app:robocasa-evaluation}. Progress is optimizer steps divided by the scheduled budget: 5k/10k/30k/100k/280k for $D=50/100/300/1000/2900$. Entries use exact logged steps, without interpolation.}
\label{tab:robocasa-training-progress-seed0}
\label{tab:robocasa-training-progress}
\label{fig:app-robocasa-training}
\label{fig:app-robocasa-optimization}
\small
\setlength{\tabcolsep}{1.5pt}
\renewcommand{\arraystretch}{1.08}
\begin{tabular*}{\linewidth}{@{\extracolsep{\fill}}rl*{10}{r}@{}}
\toprule
$D$ & Params & 10\% & 20\% & 30\% & 40\% & 50\% & 60\% & 70\% & 80\% & 90\% & 100\% \\
\midrule
50 & 44.3M & 0.41821 & 0.31303 & 0.27148 & 0.25973 & 0.23128 & 0.21949 & 0.21252 & 0.21222 & 0.20983 & 0.20765 \\
 & 85.3M & 0.41246 & 0.35694 & 0.24800 & 0.22547 & 0.21357 & 0.20352 & 0.20087 & 0.19598 & 0.19501 & 0.19514 \\
 & 218.8M & 0.40335 & 0.34023 & 0.27017 & 0.23318 & 0.21947 & 0.20875 & 0.19826 & 0.19802 & 0.19648 & 0.19532 \\
 & 648.9M & 0.33931 & 0.23841 & 0.22386 & 0.20937 & 0.20262 & 0.19702 & 0.19359 & 0.19386 & 0.19313 & 0.19279 \\
 & 1.49B & 0.27697 & 0.21628 & 0.20776 & 0.19756 & 0.19285 & 0.19263 & 0.19124 & 0.19109 & 0.19044 & 0.18934 \\
\addlinespace[3pt]
100 & 44.3M & 0.39791 & 0.27210 & 0.22569 & 0.20335 & 0.19464 & 0.18634 & 0.18098 & 0.17686 & 0.17595 & 0.17477 \\
 & 85.3M & 0.30601 & 0.22293 & 0.20872 & 0.19259 & 0.18493 & 0.17955 & 0.17515 & 0.17277 & 0.17237 & 0.17212 \\
 & 218.8M & 0.26028 & 0.21184 & 0.19617 & 0.18517 & 0.17863 & 0.17552 & 0.16948 & 0.16763 & 0.16724 & 0.16618 \\
 & 648.9M & 0.23599 & 0.20562 & 0.19045 & 0.18188 & 0.17736 & 0.17374 & 0.17132 & 0.16982 & 0.16926 & 0.16913 \\
 & 1.49B & 0.22077 & 0.18924 & 0.18223 & 0.17932 & 0.17597 & 0.17370 & 0.17270 & 0.17192 & 0.17148 & 0.17117 \\
\addlinespace[3pt]
300 & 44.3M & 0.21602 & 0.18548 & 0.16730 & 0.16072 & 0.15789 & 0.15515 & 0.15375 & 0.15248 & 0.15204 & 0.15265 \\
 & 85.3M & 0.20004 & 0.17128 & 0.15749 & 0.14874 & 0.14546 & 0.14227 & 0.13900 & 0.13765 & 0.13699 & 0.13680 \\
 & 218.8M & 0.19457 & 0.16529 & 0.15336 & 0.14426 & 0.13967 & 0.13615 & 0.13402 & 0.13292 & 0.13191 & 0.13171 \\
 & 648.9M & 0.18732 & 0.16346 & 0.15164 & 0.14691 & 0.14215 & 0.13926 & 0.13726 & 0.13627 & 0.13541 & 0.13550 \\
 & 1.49B & 0.17897 & 0.16058 & 0.15385 & 0.15074 & 0.14796 & 0.14645 & 0.14523 & 0.14438 & 0.14394 & 0.14404 \\
\addlinespace[3pt]
1000 & 44.3M & 0.16689 & 0.14403 & 0.13747 & 0.13414 & 0.13272 & 0.13176 & 0.13019 & 0.12972 & 0.12950 & 0.12933 \\
 & 85.3M & 0.15205 & 0.13051 & 0.12170 & 0.11857 & 0.11469 & 0.11205 & 0.11029 & 0.10969 & 0.10879 & 0.10873 \\
 & 218.8M & 0.15245 & 0.12478 & 0.11399 & 0.10965 & 0.10525 & 0.10286 & 0.10128 & 0.09993 & 0.09921 & 0.09937 \\
 & 648.9M & 0.14708 & 0.12308 & 0.11645 & 0.10929 & 0.10659 & 0.10402 & 0.10254 & 0.10128 & 0.10092 & 0.10117 \\
 & 1.49B & 0.14283 & 0.12695 & 0.11988 & 0.11528 & 0.11266 & 0.11105 & 0.10942 & 0.10916 & 0.10836 & 0.10853 \\
\addlinespace[3pt]
2900 & 44.3M & 0.13699 & 0.12244 & 0.11685 & 0.11388 & 0.11233 & 0.11033 & 0.10989 & 0.10918 & 0.10921 & 0.10905 \\
 & 85.3M & 0.12267 & 0.10732 & 0.10211 & 0.09790 & 0.09498 & 0.09335 & 0.09205 & 0.09155 & 0.09076 & 0.09083 \\
 & 218.8M & 0.11636 & 0.09999 & 0.09447 & 0.08980 & 0.08727 & 0.08615 & 0.08465 & 0.08417 & 0.08348 & 0.08363 \\
 & 648.9M & 0.11615 & 0.10074 & 0.09301 & 0.09021 & 0.08774 & 0.08586 & 0.08458 & 0.08391 & 0.08351 & 0.08370 \\
 & 1.49B & 0.11378 & 0.10132 & 0.09558 & 0.09188 & 0.09055 & 0.08857 & 0.08758 & 0.08646 & 0.08636 & 0.08662 \\
\bottomrule
\end{tabular*}
\end{table}

\begin{table}[!htbp]
\centering
\caption{RoboCasa holdout stochastic full-chunk RMSE at 10\% training intervals, seed 1, using the metric in Appendix~\ref{app:robocasa-evaluation}. Progress is optimizer steps divided by the scheduled budget: 5k/10k/30k/100k/280k for $D=50/100/300/1000/2900$. Entries use exact logged steps, without interpolation.}
\label{tab:robocasa-training-progress-seed1}
\small
\setlength{\tabcolsep}{1.5pt}
\renewcommand{\arraystretch}{1.08}
\begin{tabular*}{\linewidth}{@{\extracolsep{\fill}}rl*{10}{r}@{}}
\toprule
$D$ & Params & 10\% & 20\% & 30\% & 40\% & 50\% & 60\% & 70\% & 80\% & 90\% & 100\% \\
\midrule
50 & 44.3M & 0.41265 & 0.36703 & 0.30551 & 0.24915 & 0.24223 & 0.22526 & 0.21346 & 0.20761 & 0.20684 & 0.20301 \\
 & 85.3M & 0.41460 & 0.31348 & 0.25889 & 0.23951 & 0.22562 & 0.21208 & 0.20414 & 0.19972 & 0.19819 & 0.19682 \\
 & 218.8M & 0.39254 & 0.28892 & 0.23680 & 0.21865 & 0.20694 & 0.20339 & 0.19930 & 0.19517 & 0.19357 & 0.19411 \\
 & 648.9M & 0.37310 & 0.25907 & 0.22671 & 0.21701 & 0.20702 & 0.19907 & 0.20138 & 0.19804 & 0.19506 & 0.19518 \\
 & 1.49B & 0.26374 & 0.21731 & 0.20156 & 0.19317 & 0.19170 & 0.18972 & 0.18823 & 0.18731 & 0.18720 & 0.18610 \\
\addlinespace[3pt]
100 & 44.3M & 0.28741 & 0.23559 & 0.20988 & 0.19770 & 0.18795 & 0.18145 & 0.17921 & 0.17774 & 0.17699 & 0.17701 \\
 & 85.3M & 0.31299 & 0.21829 & 0.19867 & 0.18964 & 0.18305 & 0.17601 & 0.17273 & 0.17065 & 0.16928 & 0.16947 \\
 & 218.8M & 0.25743 & 0.21477 & 0.19513 & 0.18202 & 0.17580 & 0.16951 & 0.16723 & 0.16519 & 0.16534 & 0.16491 \\
 & 648.9M & 0.22760 & 0.19918 & 0.18997 & 0.18003 & 0.17849 & 0.17363 & 0.17196 & 0.17081 & 0.16950 & 0.17028 \\
 & 1.49B & 0.21579 & 0.18919 & 0.18121 & 0.18160 & 0.17432 & 0.17193 & 0.17251 & 0.17158 & 0.17115 & 0.17098 \\
\addlinespace[3pt]
300 & 44.3M & 0.21696 & 0.17824 & 0.16642 & 0.15908 & 0.15538 & 0.15390 & 0.15282 & 0.15188 & 0.15173 & 0.15193 \\
 & 85.3M & 0.20141 & 0.16884 & 0.15676 & 0.14888 & 0.14439 & 0.14156 & 0.13848 & 0.13707 & 0.13578 & 0.13544 \\
 & 218.8M & 0.19170 & 0.16711 & 0.15352 & 0.14558 & 0.14062 & 0.13714 & 0.13610 & 0.13394 & 0.13323 & 0.13310 \\
 & 648.9M & 0.19037 & 0.16538 & 0.15334 & 0.14652 & 0.14100 & 0.13936 & 0.13678 & 0.13567 & 0.13489 & 0.13487 \\
 & 1.49B & 0.17863 & 0.15991 & 0.15371 & 0.14901 & 0.14836 & 0.14607 & 0.14560 & 0.14432 & 0.14388 & 0.14376 \\
\addlinespace[3pt]
1000 & 44.3M & 0.16650 & 0.14604 & 0.13806 & 0.13510 & 0.13208 & 0.13021 & 0.12925 & 0.12887 & 0.12872 & 0.12888 \\
 & 85.3M & 0.15905 & 0.13433 & 0.12191 & 0.11875 & 0.11418 & 0.11301 & 0.11119 & 0.11089 & 0.11019 & 0.10998 \\
 & 218.8M & 0.14750 & 0.12548 & 0.11522 & 0.11001 & 0.10641 & 0.10479 & 0.10232 & 0.10158 & 0.10057 & 0.10051 \\
 & 648.9M & 0.14540 & 0.12131 & 0.11420 & 0.11018 & 0.10559 & 0.10327 & 0.10179 & 0.10062 & 0.10044 & 0.10044 \\
 & 1.49B & 0.14269 & 0.12519 & 0.11899 & 0.11489 & 0.11216 & 0.11054 & 0.10963 & 0.10837 & 0.10810 & 0.10806 \\
\addlinespace[3pt]
2900 & 44.3M & 0.13605 & 0.12162 & 0.11687 & 0.11390 & 0.11160 & 0.10998 & 0.10947 & 0.10889 & 0.10885 & 0.10851 \\
 & 85.3M & 0.12315 & 0.10721 & 0.10169 & 0.09778 & 0.09553 & 0.09331 & 0.09166 & 0.09082 & 0.09084 & 0.09069 \\
 & 218.8M & 0.11918 & 0.10115 & 0.09508 & 0.09166 & 0.08923 & 0.08669 & 0.08583 & 0.08471 & 0.08454 & 0.08447 \\
 & 648.9M & 0.11520 & 0.10078 & 0.09431 & 0.08996 & 0.08768 & 0.08639 & 0.08498 & 0.08463 & 0.08463 & 0.08436 \\
 & 1.49B & 0.11576 & 0.09991 & 0.09588 & 0.09259 & 0.08945 & 0.08781 & 0.08729 & 0.08659 & 0.08630 & 0.08615 \\
\bottomrule
\end{tabular*}
\end{table}

\clearpage
\subsection{Success Rates for Individual Tasks}
\begin{table}[!htbp]
\centering
\caption{Task-level RoboCasa SR (\%) for the 218.8M policy. Each data-scale entry averages three executions of 50 episodes per task, with training seeds separate. The final column records successes in the selected single execution: seed 1, $D=2900$, repeat 02 (713/1,150 overall).}
\label{tab:taskwise-scaling-complete}
\label{fig:app-taskwise-scaling}
\label{tab:best-n3-taskwise}
\footnotesize
\setlength{\tabcolsep}{1.7pt}
\renewcommand{\arraystretch}{1.12}
\begin{tabular*}{\linewidth}{@{\extracolsep{\fill}}lrrrrr rrrrr r@{}}
\toprule
& \multicolumn{5}{c}{Training seed 0: $D$} & \multicolumn{5}{c}{Training seed 1: $D$} & Selected run \\
\cmidrule(lr){2-6}\cmidrule(lr){7-11}
Task & 50 & 100 & 300 & 1000 & 2900 & 50 & 100 & 300 & 1000 & 2900 & Successes \\
\midrule
CloseDoubleDoor & 23.33 & 67.33 & 88.00 & 91.33 & 94.67 & 36.67 & 68.67 & 92.67 & 90.67 & 91.33 & 45/50 \\
CloseDrawer & 88.00 & 98.67 & 100.00 & 100.00 & 98.67 & 87.33 & 96.00 & 99.33 & 100.00 & 97.33 & 48/50 \\
CloseSingleDoor & 57.33 & 67.33 & 82.67 & 83.33 & 87.33 & 50.67 & 76.00 & 76.00 & 84.00 & 84.00 & 45/50 \\
CoffeePressButton & 44.67 & 56.67 & 78.00 & 91.33 & 94.67 & 38.00 & 51.33 & 84.67 & 86.00 & 94.67 & 49/50 \\
CoffeeServeMug & 12.67 & 20.00 & 54.00 & 62.00 & 64.67 & 6.00 & 24.00 & 61.33 & 67.33 & 69.33 & 38/50 \\
CoffeeSetupMug & 1.33 & 3.33 & 19.33 & 28.00 & 32.67 & 0.00 & 8.67 & 22.67 & 32.67 & 42.00 & 23/50 \\
OpenDrawer & 24.00 & 52.67 & 74.00 & 92.00 & 79.33 & 22.67 & 57.33 & 79.33 & 86.00 & 86.67 & 43/50 \\
OpenSingleDoor & 28.67 & 32.67 & 46.00 & 61.33 & 62.00 & 21.33 & 26.67 & 50.67 & 57.33 & 66.00 & 30/50 \\
PnPCabToCounter & 0.67 & 4.00 & 11.33 & 12.00 & 14.67 & 1.33 & 7.33 & 14.67 & 14.00 & 16.67 & 9/50 \\
PnPCounterToCab & 0.00 & 10.67 & 32.67 & 28.67 & 42.67 & 3.33 & 5.33 & 26.67 & 39.33 & 42.67 & 23/50 \\
PnPCounterToMicrowave & 0.67 & 4.00 & 3.33 & 17.33 & 20.00 & 0.00 & 4.67 & 8.67 & 14.67 & 32.00 & 16/50 \\
PnPCounterToSink & 1.33 & 6.67 & 16.67 & 26.67 & 28.00 & 3.33 & 5.33 & 6.00 & 26.00 & 32.00 & 18/50 \\
PnPCounterToStove & 0.67 & 2.00 & 4.00 & 21.33 & 22.00 & 0.00 & 0.67 & 1.33 & 20.67 & 32.00 & 16/50 \\
PnPMicrowaveToCounter & 0.67 & 3.33 & 10.67 & 18.67 & 18.00 & 0.00 & 0.67 & 9.33 & 21.33 & 18.00 & 10/50 \\
PnPSinkToCounter & 1.33 & 4.00 & 22.67 & 47.33 & 52.67 & 4.67 & 8.00 & 14.00 & 34.00 & 52.00 & 25/50 \\
PnPStoveToCounter & 0.00 & 1.33 & 8.00 & 45.33 & 42.00 & 0.00 & 0.67 & 10.00 & 41.33 & 48.67 & 22/50 \\
TurnOffMicrowave & 39.33 & 66.00 & 93.33 & 98.67 & 99.33 & 63.33 & 65.33 & 95.33 & 94.00 & 100.00 & 50/50 \\
TurnOffSinkFaucet & 54.00 & 85.33 & 83.33 & 84.67 & 84.00 & 65.33 & 58.67 & 72.67 & 89.33 & 88.00 & 41/50 \\
TurnOffStove & 6.67 & 6.67 & 16.67 & 20.67 & 21.33 & 1.33 & 6.00 & 18.00 & 18.67 & 14.00 & 9/50 \\
TurnOnMicrowave & 32.67 & 84.67 & 86.67 & 90.00 & 95.33 & 42.00 & 47.33 & 89.33 & 93.33 & 92.00 & 46/50 \\
TurnOnSinkFaucet & 27.33 & 43.33 & 62.00 & 76.00 & 70.00 & 18.00 & 60.67 & 51.33 & 69.33 & 65.33 & 32/50 \\
TurnOnStove & 14.67 & 30.67 & 48.00 & 69.33 & 62.00 & 12.67 & 21.33 & 47.33 & 66.67 & 66.00 & 36/50 \\
TurnSinkSpout & 45.33 & 72.67 & 65.33 & 78.67 & 78.67 & 44.67 & 64.00 & 81.33 & 82.67 & 76.67 & 39/50 \\
\midrule
All tasks & 21.97 & 35.83 & 48.12 & 58.46 & 59.33 & 22.72 & 33.25 & 48.38 & 57.80 & 61.19 & 713/1,150 \\
\bottomrule
\end{tabular*}
\end{table}

\clearpage
\subsection{Local BC-Transformer Comparison}
\label{app:local-bc-transformer}
The local BC-Transformer comparison uses the same 23 tasks, $D=300$ demonstrations, and final 1,150 evaluation episodes as WorldToken. It follows the official native training recipe: one training seed (123), batch size 16, 500,000 optimizer steps, and AdamW weight decay 0.01. It observes and replans after every action, whereas WorldToken observes every four control steps and executes four actions per query. The comparison therefore matches the data and evaluation episodes but not the optimizer settings, training compute, seed count, observation cadence, or action-execution cadence.

Table~\ref{tab:local-bc-transformer} lists all repeated evaluations underlying the 31.28\% local reference in the main text.
\begin{table}[!htbp]
\centering
\caption{Local BC-Transformer trained with the official native recipe (Appendix~\ref{app:local-bc-transformer}): training seed 123, $D=300$, final 500k-step checkpoint. All three repeats use the same 1,150 episodes across 23 tasks. SR is in percent; SD is the sample standard deviation across repeats, in percentage points.}
\label{tab:local-bc-transformer}
\small
\setlength{\tabcolsep}{3pt}
\renewcommand{\arraystretch}{1.12}
\begin{tabular*}{\linewidth}{@{\extracolsep{\fill}}lrrrrr@{}}
\toprule
Policy & Training seed & Repeat 1 & Repeat 2 & Repeat 3 & Mean $\pm$ SD \\
\midrule
BC-Transformer & 123 & 32.00 & 30.78 & 31.04 & $31.28\pm0.64$ \\
\bottomrule
\end{tabular*}
\end{table}

\subsection{Comparison with the Reported $\pi_{0.5}$ Result}
\label{app:public-scale}

Table~\ref{tab:public-scale} gives the public comparison used for context in the main text. WorldToken uses a frozen CLIP text encoder and trains its visual and policy modules from scratch. The reported $\pi_{0.5}$ policy uses multi-source robot and Web pretraining. \textbf{Although our 23-task evaluation excludes \texttt{OpenDoubleDoor}, which is included in the 24-task benchmark underlying the reported $\pi_{0.5}$ result \citep{kim2026a_rldx}, WorldToken N2 at $D=2900$ achieves over 80\% mean SR across the other three tasks in RoboCasa's official ``Opening and closing doors'' family: \texttt{OpenSingleDoor}, \texttt{CloseSingleDoor}, and \texttt{CloseDoubleDoor} (training seed 0, averaged over three evaluations). WorldToken strictly follows the official RoboCasa
evaluation protocol for all 23 evaluated tasks
\citep{nasiriany2024robocasa}.}

\begin{table}[!htbp]
\centering
\small
\caption{WorldToken and the reported $\pi_{0.5}$ result on RoboCasa Kitchen.
Training data counts are generated demonstrations per task. The WorldToken
result averages two training seeds, each with three evaluations. The
$\pi_{0.5}$ result is reported by \citet{kim2026a_rldx}.}
\label{tab:public-scale}
\begin{tabular*}{\linewidth}{@{\extracolsep{\fill}}lrrrr@{}}
\toprule
Policy & Training data & Parameters & Tasks & SR (\%) \\
\midrule
WorldToken & 2,900 & 85.3M & 23 & 59.4 \\
$\pi_{0.5}$ & 300 & $\sim$3.35B & 24 & 62.1 \\
\bottomrule
\end{tabular*}
\end{table}

\section{Token Interfaces, Policy Performance, and Computational Cost}
\label{app:granularity}

\subsection{Single- and Multi-Token Interfaces}

The 85.3M-parameter reference passes one learned world token per timestep. For $K=4$, we concatenate the same four internal readout states, jointly project them to $4d$ dimensions, and split the result into four $d$-dimensional tokens. The number of internal readouts remains four. For $K=50$, we pass all 48 post-fusion CNN feature tokens, followed by the proprioception and task tokens. These features are already fused, rather than raw pixels. In every variant, tokens are grouped by timestep, and the decoder reads the final contextualized token of the current timestep. The history always spans ten policy timesteps, giving $N=10K$ temporal positions.

The temporal backbone applies a standard lower-triangular causal mask over the flattened token sequence, including within each timestep. RoPE assigns consecutive positions to all $CK$ tokens, so tokens from the same timestep have distinct positions. The final token can attend to all earlier tokens in its timestep and all earlier timesteps. For $K=4$, this final token is the last block of the joint projection and has no predefined modality. For $K=50$, it is the contextualized task-language token.

At each $D$, all variants use the same training demonstrations, holdout demonstrations, and final 1,150 evaluation episodes. The added variants use only training seed 0. Because parameter count and temporal compute are not jointly matched, the results describe these token interfaces; they do not isolate the causal effect of token count or establish an optimal interface.

\subsection{Policy Performance across Data Scales}

Table~\ref{tab:token-allocation-complete} reports results across the five
data scales. Relative to $K=1$, $K=50$ reduces full-chunk RMSE by
1.7--4.2\%. Its SR gain is larger at
$D=50$ and $100$, at 5.10 and 6.38 percentage points, respectively,
and is 2.09, 2.06, and 1.59 points at $D=300$, $1000$, and $2900$.
These differences use the exact success counts before rounding the
displayed means.

Increasing the learned interface to $K=4$ gives less uniform results:
RMSE improves at two of five data scales, and the SR change ranges from
$-1.13$ to $+6.96$ points. The largest gain again occurs at $D=100$.
These seed-0 results show that both the interface construction and data
scale matter; increasing token count alone does not give a uniform gain.
Panel (b) reports full-window temporal-backbone FLOPs.
Appendix~\ref{app:token-history-compute} separately reports per-query costs
with cached history, including the encoder and action decoder.
\begin{table}[!htbp]
\centering
\caption{Token-count experiments at $C=10$, training seed 0. (a) Final-checkpoint results for all 15 $(D,K)$ settings. Stochastic full-chunk RMSE follows Appendix~\ref{app:robocasa-evaluation}. SR entries give three evaluation percentages (1,150 episodes each) and their mean $\pm$ sample SD. The $K=1,4,50$ interfaces are defined in Appendix~\ref{app:granularity}. (b) Temporal dimensions and analytic full-prefix FLOPs (Equation~\ref{eq:temporal-flops}), excluding encoder and decoder. Relative FLOPs use the 85.3M, $K=1$ reference; parameter count and compute are not jointly matched.}
\label{tab:token-allocation-complete}
\label{tab:token-granularity}
\label{tab:token-granularity-scales}
\label{tab:token-flops}
\label{tab:token-width-flops}
\label{fig:app-token-allocation}
\footnotesize
\setlength{\tabcolsep}{3pt}
\renewcommand{\arraystretch}{1.12}
\renewcommand{\arraystretch}{1.0}
\begin{tabular*}{\linewidth}{@{\extracolsep{\fill}}rrrrrrcr@{}}
\toprule
\multicolumn{8}{l}{\textbf{(a) Complete token-interface results}} \\
$D$ & $K$ & Params & $N$ & Rel. FLOPs & RMSE & SR: repeats 1 / 2 / 3 & SR: mean $\pm$ SD \\
\midrule
50 & 1 & 85.3M & 10 & $1.00\times$ & 0.19514 & 20.96/20.96/22.26 & $21.39\pm0.75$ \\
 & 4 & 92.4M & 40 & $4.03\times$ & 0.20163 & 20.00/19.48/21.30 & $20.26\pm0.94$ \\
 & 50 & 83.0M & 500 & $55.31\times$ & 0.19086 & 25.83/27.04/26.61 & $26.49\pm0.62$ \\
\midrule
100 & 1 & 85.3M & 10 & $1.00\times$ & 0.17212 & 30.17/31.74/31.65 & $31.19\pm0.88$ \\
 & 4 & 92.4M & 40 & $4.03\times$ & 0.16586 & 38.17/38.00/38.26 & $38.14\pm0.13$ \\
 & 50 & 83.0M & 500 & $55.31\times$ & 0.16497 & 37.57/38.09/37.04 & $37.57\pm0.52$ \\
\midrule
300 & 1 & 85.3M & 10 & $1.00\times$ & 0.13680 & 48.00/46.17/48.26 & $47.48\pm1.14$ \\
 & 4 & 92.4M & 40 & $4.03\times$ & 0.13826 & 49.22/47.83/49.22 & $48.75\pm0.80$ \\
 & 50 & 83.0M & 500 & $55.31\times$ & 0.13230 & 49.74/49.13/49.83 & $49.57\pm0.38$ \\
\midrule
1000 & 1 & 85.3M & 10 & $1.00\times$ & 0.10873 & 55.83/56.35/54.87 & $55.68\pm0.75$ \\
 & 4 & 92.4M & 40 & $4.03\times$ & 0.10972 & 55.83/56.00/54.96 & $55.59\pm0.56$ \\
 & 50 & 83.0M & 500 & $55.31\times$ & 0.10686 & 57.57/58.09/57.57 & $57.74\pm0.30$ \\
\midrule
2900 & 1 & 85.3M & 10 & $1.00\times$ & 0.09083 & 58.96/58.17/60.09 & $59.07\pm0.96$ \\
 & 4 & 92.4M & 40 & $4.03\times$ & 0.08997 & 60.00/59.83/60.00 & $59.94\pm0.10$ \\
 & 50 & 83.0M & 500 & $55.31\times$ & 0.08851 & 60.26/59.83/61.91 & $60.67\pm1.10$ \\
\bottomrule
\end{tabular*}
\par\medskip
\begin{tabular*}{\linewidth}{@{\extracolsep{\fill}}lrrrrrr@{}}
\toprule
\multicolumn{7}{l}{\textbf{(b) Temporal dimensions and full-prefix FLOPs}} \\
Configuration & $d$ & $f$ & $L$ & $K$ & $N$ & Temporal FLOPs \\
\midrule
85.3M, one token & 768 & 2048 & 4 & 1 & 10 & 0.567G ($1.00\times$) \\
92.4M, four tokens & 768 & 2048 & 4 & 4 & 40 & 2.285G ($4.03\times$) \\
83.0M, fifty tokens & 768 & 2048 & 4 & 50 & 500 & 31.384G ($55.31\times$) \\
\bottomrule
\end{tabular*}
\end{table}

\subsection{Why Short-Context Cost Is Nearly Linear in Token Count}

\label{app:temporal-flops}
We first count temporal-backbone computation for a full window, without KV caching. For temporal width $d$, feed-forward width $f$, depth $L$, and sequence length $N$, the leading matrix-multiplication count is
\begin{equation}
F(N)=L(8d^2+6df)N+4LdN^2,
\label{eq:temporal-flops}
\end{equation}
using two FLOPs per multiply--accumulate. The $d=768$, $f=2048$, $L=4$ backbone gives $56{,}623{,}104N+12{,}288N^2$, with $N=10K$ at $C=10$. Table~\ref{tab:token-flops} lists the resulting counts. They exclude the within-timestep encoder, action decoder, and temporal wrapper's output projection. They also assume full rectangular attention products; causal kernels may skip masked work. These FLOPs are not end-to-end latency measurements. The cached operator counts in Appendix~\ref{app:token-history-compute} include the output projection and measure the incremental work per policy query.

The linear term includes the attention projections and feed-forward layers. The quadratic term comes from attention between token positions. The two terms are equal when $N=2d+1.5f=4{,}608$ for this backbone. At $C=10$, the $K=1,4,50$ interfaces contain $N=10,40,500$ tokens. The linear term therefore dominates at these lengths, giving the approximately $4.03$- and $55.31$-fold costs in Table~\ref{tab:token-flops}.

\subsection{Module Costs with Cached History}
\label{app:history-compute}
\label{app:token-history-compute}

We next count incremental computation for the N2 $K=1,4,50$ interfaces with a growing KV cache and no eviction. A context of $C$ observations, including the current one, contains $N=KC$ tokens. Each query encodes only the current observation and processes its $K$ new tokens alongside $(C-1)K$ cached tokens. At $C=1$, there is no prior history. This cached calculation differs from the RoboCasa success-rate evaluations, which recompute the full window.

\begin{table}[!htbp]
\centering
\small
\setlength{\tabcolsep}{3pt}
\caption{Cached computation in GFLOPs per timestamp, where one timestamp denotes one policy query. (a) Baseline with one observation, $C=1$. Fusion/projections include encoder readouts, and encoder total is a subtotal. Action-head costs include all 20 diffusion steps; totals are computed before rounding. (b) Additional cost per query for 100 more retained observations. Relative slopes compare the cost of added history, not total policy costs.}
\label{tab:history-baseline-growth}
\label{tab:history-compute-baselines}
\label{tab:token-history-components}
\begin{tabular*}{\linewidth}{@{\extracolsep{\fill}}rrrrrrr@{}}
\toprule
\multicolumn{7}{l}{\textbf{(a) One-observation baseline: $C=1$}} \\
$K$ & CNN & \shortstack{Fusion +\\projections} & \shortstack{Encoder\\total} & \shortstack{Temporal\\backbone} & \shortstack{Action\\head} & Total \\
\midrule
1 & 19.1103 & 2.1189 & 21.2292 & 0.0578 & 0.4502 & 21.7372 \\
4 & 19.1103 & 2.1330 & 21.2433 & 0.2314 & 0.4502 & 21.9250 \\
50 & 19.1103 & 1.9606 & 21.0709 & 2.9209 & 0.4502 & 24.4420 \\
\bottomrule
\end{tabular*}
\par\medskip
\begin{tabular*}{\linewidth}{@{\extracolsep{\fill}}rrr@{}}
\toprule
\multicolumn{3}{l}{\textbf{(b) Cost of 100 additional historical observations}} \\
$K$ & $\Delta_{100,K}$ (GFLOPs/timestamp) & Slope relative to $K=1$ \\
\midrule
1 & 0.0012288 & $1\times$ \\
4 & 0.0196608 & $16\times$ \\
50 & 3.0720000 & $2500\times$ \\
\bottomrule
\end{tabular*}
\end{table}

The three CNNs dominate encoder computation. Fusion/projections cover the remaining encoder operations and readouts; the temporal and action-head architectures are shared across variants. At $C=10$, total cached computation is 21.7373, 21.9268, and 24.7185 GFLOPs/query for $K=1,4,50$, respectively. Thus, including all modules, $K=50$ costs $1.14\times$ as much as $K=1$ at this history length.

\subsection{Computational Growth with History Length}
\label{app:history-growth}

The encoder and action sampler have fixed input shapes; only temporal computation grows with $C$. Per 100 additional observations,
\begin{equation}
F_K(C)=F_K(1)+\frac{C-1}{100}\Delta_{100,K},
\label{eq:cached-history-flops}
\end{equation}
where $F_K$ is total GFLOPs/query and Table~\ref{tab:history-baseline-growth}(b) gives $\Delta_{100,K}$. With four temporal layers of width 768, the $K\times CK$ attention products make the cost of added history proportional to $K^2$. For equal increases in observation count, the added costs have ratio $1:16:2500$. This ratio describes the growth with history, not the total policy cost.

\paragraph{FLOPs counting.}
\label{app:cached-flops-protocol}

Counts use archived architectures with random weights, synthetic batch-one inputs, CPU FP32, and no gradients in evaluation mode. Inputs are three $128\times128$ RGB views, 16-D proprioception, and a precomputed 768-D language embedding; outputs are ten-step, 12-D action chunks. \texttt{FlopCounterMode} counts convolution and dense matrix products at two FLOPs per multiply--accumulate, including temporal projections and all 20 denoising steps. Normalization, softmax, pooling, other elementwise operations, and data movement are excluded.

\section{History Truncation and Short-Context Training on RoboCasa}
\label{app:robocasa-history}

\subsection{Inference-Time History Truncation}
For each scaling checkpoint, we vary only $C_{\rm test}\in\{1,2,5,10\}$, keeping the initial conditions, environment seeds, rollout seed, and execution settings fixed. Each shorter-history setting uses one evaluation; the $C_{\rm test}=10$ reference is the mean of three evaluations.

Table~\ref{tab:scaling-complete} in Appendix~\ref{app:complete-sr-grids} reports the complete history-truncation results for all 50 scaling policies.

\subsection{Training and Evaluation with Matched Context Lengths}
\label{app:matched-training-history}
We also train separate 218.8M policies at $D=300$ with matching training and evaluation contexts, $C\in\{1,2,5,10\}$. Each setting uses two training seeds and three evaluations per final checkpoint.

The global batch sizes are $1920,960,384,192$ windows for $C=1,2,5,10$, respectively, with 30k optimizer steps. Multiplying batch size by $C$ gives 1,920 observation-query positions per update and 57.6M over training. These budgets are matched before excluding queries without all ten action targets. All variants sample crops from the same training demonstrations, using their own context length $C$. The exact queries and the number of valid action chunks after masking can therefore differ across variants.

\begin{table}[!htbp]
\centering
\caption{D300 matched-context training with N3 (218.8M), $C_{\rm train}=C_{\rm test}=C$. Each final 30k-step checkpoint is evaluated three times on 1,150 episodes. Stochastic full-chunk RMSE uses the context-specific crops in Appendix~\ref{app:robocasa-evaluation}. SR entries give the repeated evaluation percentages and their mean $\pm$ sample SD. $C=10$ is the scaling reference.}
\label{tab:matched-context-complete}
\small
\setlength{\tabcolsep}{4pt}
\renewcommand{\arraystretch}{1.12}
\renewcommand{\arraystretch}{1.05}
\begin{tabular*}{\linewidth}{@{\extracolsep{\fill}}rrrrrrr@{}}
\toprule
& & & \multicolumn{3}{c}{SR (\%): repeats} & SR (\%) \\
\cmidrule(lr){4-6}
$C$ & Seed & RMSE & 1 & 2 & 3 & Mean $\pm$ SD \\
\midrule
1 & 0 & 0.150260 & 45.65 & 47.65 & 46.96 & $46.75\pm1.02$ \\
 & 1 & 0.147767 & 48.17 & 46.00 & 48.87 & $47.68\pm1.50$ \\
\midrule
2 & 0 & 0.149383 & 47.83 & 46.52 & 46.26 & $46.87\pm0.84$ \\
 & 1 & 0.147528 & 47.74 & 50.09 & 47.74 & $48.52\pm1.36$ \\
\midrule
5 & 0 & 0.137165 & 51.74 & 51.22 & 51.48 & $51.48\pm0.26$ \\
 & 1 & 0.139603 & 50.87 & 49.04 & 50.09 & $50.00\pm0.92$ \\
\midrule
10 & 0 & 0.131710 & 46.70 & 50.00 & 47.65 & $48.12\pm1.70$ \\
 & 1 & 0.133098 & 48.87 & 49.13 & 47.13 & $48.38\pm1.09$ \\
\bottomrule
\end{tabular*}
\end{table}

The truncation experiment measures a fixed policy's reliance on history; the separately trained policies show how training with shorter contexts changes that reliance.

Separately trained short-context policies recover much of the success
lost when long-context checkpoints are truncated. Across the two training
seeds, mean SR is 47.22\%, 47.70\%, 50.74\%, and 48.25\% at
$C=1,2,5,10$, respectively. The $C=5$ policies have the highest observed
SR, while reported RMSE decreases with $C$ under the context-specific
crop protocol in Appendix~\ref{app:robocasa-evaluation}.

\addtocontents{toc}{\protect\newpage}
\section{History Use and Extended Execution in RMBench Blocks Ranking}
\label{app:ranking-protocol}
\label{app:ranking-loss}
\label{app:ranking-results}

\subsection{Task and Demonstrations}

Blocks Ranking uses three colored blocks in left, middle, and right slots, with target order $(1,2,3)$. The reference procedure begins with a button press, then follows the positional swaps in Eq.~\eqref{eq:swap-seq}, pressing after each swap and stopping at the target. The five non-target initial permutations require one through five swaps.

Initial training uses 50 \texttt{demo\_clean} demonstrations for each of \texttt{battery\_try}, \texttt{blocks\_ranking\_try}, \texttt{cover\_blocks}, \texttt{observe\_and\_pickup}, \texttt{press\_button}, \texttt{put\_back\_block}, \texttt{rearrange\_blocks}, \texttt{swap\_T}, and \texttt{swap\_blocks}: 450 demonstrations in total, with no initial holdout.

We continue from the seed-1 nine-task checkpoint at step 5,000 using 45 Blocks Ranking demonstrations. Episode IDs 6, 15, 19, 25, and 46 are held out from this continuation (split seed 4). All five were seen during initial training, so their offline loss monitors further training on previously seen demonstrations.

\subsection{Detailed Experimental Setup}
\label{app:ranking-training}

\paragraph{Model architecture.}
The RMBench model has 54,300,238 trainable parameters and a 768-D world token. Its within-timestep encoder has two 768-D layers, six attention heads, and SwiGLU FFN width 3,072. The temporal backbone has four 768-D layers, six heads, and FFN width 2,048. The diffusion action decoder has two 192-D layers, six heads, and FFN width 768. Four readout tokens pass through both fusion layers together with the observation tokens, using the asymmetric mask and shared FFNs in Appendix~\ref{app:method-encoder} (Eq.~\eqref{eq:readout}).

The encoder shares a $20\times20$ patch projection across the three native $240\times320$ RGB camera fields \texttt{head\_camera}, \texttt{left\_camera}, and \texttt{right\_camera}. It also receives the 14-D \texttt{joint\_action/vector} proprioceptive field and a nine-way task one-hot condition. Each action chunk contains $H=8$ absolute 14-D joint-position commands. Reconstruction, predictive auxiliary losses, and action-decoder cross-attention to observation tokens are disabled.

\paragraph{Sequence construction and supervision.}
Initial training uses a maximum history of $C_{\mathrm{train}}=288$ and samples one observation sequence per episode in each dataset pass. Observations are four raw frames apart. For an episode of length $L$, let $j_{\max}=L-H-1$. If $j_{\max}\leq4(C_{\mathrm{train}}-1)$, we uniformly sample a phase from $\{0,1,2,3\}$ and take its complete subsequence through $j_{\max}$. Otherwise, we uniformly sample a legal start and take a capped window.

Continuation increases the cap to $C_{\mathrm{train}}=608$ and constructs four sequences per episode, starting at raw frames 0, 1, 2, and 3 with stride four. All 50 Blocks Ranking demonstrations fit within this cap: the maximum length is $L=2,436$, giving at most 607 queries with complete future action chunks per phase. Sequences are shuffled during training. The five continuation holdout demonstrations use the same four phases and are evaluated with the ordinary diffusion objective.

Every sampled query with a complete future chunk contributes loss and sees only the causal prefix of its sampled sequence. Each optimizer update averages eight single-window microbatches; the batch count therefore refers to windows, not individual action queries.

\begin{table}[!htbp]
\centering
\small
\caption{Recorded Blocks Ranking continuation and action-generation settings. The same step-5,500 checkpoint is used for every visible-history intervention.}
\label{tab:ranking-training-protocol}
\begin{tabular}{@{}p{0.32\linewidth}p{0.62\linewidth}@{}}
\toprule
Setting & Value \\
\midrule
Initial / continuation budget & 5,000 / 500 optimizer steps; continuation uses 45 training / 5 continuation-holdout demonstrations \\
Training context cap & Initial $C_{\mathrm{train}}=288$; continuation $C_{\mathrm{train}}=608$ \\
Microbatch / accumulation & 1 / 8 steps \\
Optimizer & AdamW, $\beta=(0.9,0.95)$, $\epsilon=10^{-8}$, zero weight decay \\
Gradient clipping / precision & Unit gradient norm / BF16 \\
Reference learning rates & Within-timestep encoder and temporal backbone: $4.25\times10^{-4}$; action decoder: $3.0\times10^{-4}$ \\
Schedule & Initial 250-step warmup, then cosine decay to 10\% of reference; continuation restores optimizer/scheduler state at global step 5,000 \\
Action normalization & Inherited per-dimension min--max statistics from the nine-task training split; nondegenerate dimensions mapped to $[-1,1]$ \\
Diffusion head & Epsilon prediction; 20 squared-cosine DDPM steps; fixed-small variance; denoised clipping to $[-1,1]$ \\
Rollout sampling & Stochastic ancestral reverse steps; four executed actions per replanning query \\
\bottomrule
\end{tabular}
\end{table}

AdamW uses separate parameter groups for the within-timestep encoder, temporal backbone, and action decoder. Continuation preserves optimizer and action-normalization state and extends the cosine schedule to step 5,500 without restarting warmup.

We construct a fixed set of 100 initial conditions using RMBench's \texttt{demo\_clean} setup, accepting candidates when the scripted expert passes both planning and official success checks. The first 100 candidate seeds, 100000--100099, all passed without rejection.

All history interventions keep the same final checkpoint, the initial conditions, success predicate, and horizon fixed to measure the policy's use of visible history.

\paragraph{Action alignment.}
Let $j_t$ be the raw data-frame index of policy query $t$. The observation is \texttt{vector[$j_t$]}, and the target contains the next $H$ absolute 14-D joint-position commands, \texttt{vector[$j_t+1$:$j_t+H+1$]}. The first target is the first command applied after the current observation. The one-index offset comes from the dataset's storage convention and adds no open-loop delay. We sample only queries with complete future target chunks, so training uses no padded terminal targets.

\subsection{Task-Specific Training Objective}
\label{app:ranking-objective}

During Blocks Ranking continuation, a training-only loss modification favors a small additional offset along the expert's local joint-space descent direction, motivated by the button press. Orthogonal components and targets without a usable descent direction retain the ordinary diffusion loss.

Let $i\in\{0,\ldots,H-1\}$ be an action position, $r=j_t+i+1$ its target raw frame, $q_r=\texttt{vector}[r]$, and $z^{\mathrm{EE}}_r$ the recorded left-end-effector height. Define $\Delta z_i=z^{\mathrm{EE}}_r-z^{\mathrm{EE}}_{r-1}$ and the 14-D direction $d_i=(q_r^{1:6}-q_{r-1}^{1:6},0,\ldots,0)$, retaining only the six left-arm joints. A descent candidate requires $\Delta z_i<-0.2$~mm and $\|d_i\|_2>10^{-6}$. Both increments use adjacent raw frames, rather than the stride-four observations. The recorded end-effector displacement selects frames and scales the joint-space direction. The loss applies no forward kinematics to predicted actions, and end-effector pose is not a policy input.

For dimension $a$, action normalization is $x_a=(q_a-\mu_a)/s_a$, with $\mu_a=(q_a^{\max}+q_a^{\min})/2$ and $s_a=(q_a^{\max}-q_a^{\min})/2$ from initial nine-task training. A half-range below $10^{-4}$ is replaced by $s_a=1$. Directions use only this linear scale, without subtracting $\mu$:
\begin{equation}
v_i=d_i\oslash s,\qquad
u_i=\frac{v_i}{\|v_i\|_2},\qquad
c_i=\left\langle \frac{0.004}{-\Delta z_i}v_i,u_i\right\rangle,\qquad
p_i=\frac{1.5}{4}c_i .
\label{eq:ranking-descent-direction}
\end{equation}
Here $\oslash$ denotes componentwise division, and distances used in $c_i$ are in meters. A direction is usable when $\|v_i\|_2>10^{-8}$ and $c_i>0$. Otherwise, the original loss is retained; we set $u_i=0$ when $v_i=0$.

Let $k\in\{0,\ldots,19\}$ denote the diffusion step, sampled once per query chunk, and let $x_i^{(k)}$ be its noisy normalized target. The loss uses the \emph{unclipped} reconstruction
\begin{equation}
\hat{x}_{0,i}=\frac{x_i^{(k)}-\sqrt{1-\bar\alpha_k}\,\hat\epsilon_i}{\sqrt{\bar\alpha_k}},
\qquad e_i=\langle\hat{x}_{0,i}-x_{0,i},u_i\rangle .
\end{equation}
Clipping is used during sampling (Table~\ref{tab:ranking-training-protocol}), but not for this training reconstruction. We replace the DDPM prediction-loss component parallel to $u_i$ with
\begin{equation}
\ell_{\parallel}(e_i)
=
3\kappa_k
\begin{cases}
16(p_i-e_i)^2, & e_i < p_i, \\
0, & p_i \le e_i \le c_i, \\
(e_i-c_i)^2, & e_i > c_i,
\end{cases}
\label{eq:ranking-descent-loss}
\end{equation}
where $\kappa_k=\bar\alpha_k/\max(1-\bar\alpha_k,10^{-8})$ maps the denoised-action distance to epsilon-prediction space. Writing $\eta_i=\hat\epsilon_i-\epsilon_i$ and $m_i\in\{0,1\}$ for the usable-direction mask, the loss for one query is
\begin{equation}
\mathcal L_{\mathrm{query}}=
\frac{1}{14H}\sum_{i=0}^{H-1}\left[
\|\eta_i\|_2^2+m_i\left\{\ell_{\parallel}(e_i)-\langle\eta_i,u_i\rangle^2\right\}\right].
\label{eq:ranking-loss-reduction}
\end{equation}
The factors 3 and 16 apply only to the replaced parallel epsilon-error component. The preferred 1.5--4~mm range scales the expert's local joint-space direction as above, using no button-contact or press-stage label. We average query losses within each window, then average the eight microbatch window losses per optimizer update.

The standard-loss and modified-loss continuations share the nine-task step-5,000 checkpoint, 45/5 split, 500 additional optimizer steps, architecture, and evaluation settings. They use different recorded code revisions, so the comparison does not isolate the effect of the objective alone. All history interventions use the same modified-loss checkpoint across $C$.

\subsection{Comparison of Training Objectives}
\label{app:ranking-loss-results}

The standard-loss and modified-loss policies succeed in 13/100 and 95/100 evaluation episodes, respectively. Both frequently reach the evaluator's target block geometry, but success also requires the open-gripper and button conditions. The matched-seed example shows how a millimeter-scale difference in button position can change the binary outcome. As detailed in Appendix~\ref{app:ranking-objective}, the full success-rate gap compares two training procedures whose differences are not limited to the loss.

\paragraph{Diagnostic definitions and records.}
Target geometry means the evaluator's block-order and pairwise-position conditions, reached at any check regardless of the gripper and button conditions. For each failure that reaches this geometry, we record the minimum button position while it holds. There are 86 such standard-loss failures and three modified-loss failures. Button positions are in mm, with a $-5$\,mm threshold.

Standard-loss summaries and paired outcome counts come from saved evaluation records and were not recomputed from raw rollouts. Modified-loss summaries and the matched-seed example use retained rollout and diagnostic records. Paired outcome counts compare separate stochastic rollouts. The last row uses seed 100084 and a separate standard-loss diagnostic execution.

\begin{table}[!htbp]
\centering
\small
\renewcommand{\arraystretch}{1.10}
\caption{Standard-loss and modified-loss results at $C=608$ on the same fixed evaluation set of 100 initial conditions. Panels report episode outcomes, paired outcome counts, and button-position diagnostics (mm). Definitions and source records are given in Appendix~\ref{app:ranking-loss-results}.}
\label{tab:ranking-loss-results}
\begin{tabular*}{\linewidth}{@{\extracolsep{\fill}}p{0.57\linewidth}cc@{}}
\toprule
 & Standard loss & Modified loss \\
\midrule
\multicolumn{3}{@{}l}{\textbf{(a) Episode outcomes}} \\
Evaluator successes & $13$ & 95 \\
Evaluator failures & $87$ & 5 \\
Target block geometry reached & $99$ & 98 \\
Failures reaching target block geometry & $86$ & 3 \\
Failures without target block geometry & $1$ & 2 \\
\midrule
\multicolumn{3}{@{}l}{\textbf{(b) Paired outcome counts}} \\
Both policies succeed & \multicolumn{2}{c}{$13$} \\
Only standard-loss policy succeeds & \multicolumn{2}{c}{$0$} \\
Only modified-loss policy succeeds & \multicolumn{2}{c}{$82$} \\
Both policies fail & \multicolumn{2}{c}{$5$} \\
\midrule
\multicolumn{3}{@{}l}{\textbf{(c) Button-position diagnostics (mm)}} \\
Deepest button position among geometry-reaching failures & $-4.65$ & $-0.004$ \\
Median of per-episode deepest positions in those failures & $-0.54$ & $-0.004$ \\
Matched seed 100084: deepest position with target geometry & $-1.485$ & $-5.020$ \\
\bottomrule
\end{tabular*}
\end{table}

\subsection{History Intervention and Success Criteria}
\label{app:ranking-behavior-protocol}

\textbf{Our standard 100-episode Blocks Ranking evaluations strictly follow the official RMBench rollout protocol \citep{chen2026rmbench}, using \texttt{demo\_clean}, expert-validated initial conditions, the unmodified official success checks, and the official 3,500-action
horizon. The official task initializer samples uniformly from the five non-target block permutations, corresponding to one through five swaps in the reference procedure. Our evaluation contains 24, 14, 28, 18, and 16 episodes in these categories, respectively: 76\% require multiple
reference swaps and 62\% require at least three. The complete categories, episode counts, and stratified success rates are reported in Table~\ref{tab:k-stratified} in the main text, and Figure~\ref{fig:ranking-storyboard} illustrates a five-swap episode.}

Each history condition evaluates the same 100 initial conditions using the official success predicate and a 3,500-action (210-second) horizon. Queries are separated by four 0.06-second actions. We report a nominal history budget of $0.24C$ seconds: 145.92, 69.12, 30.72, 15.36, and 7.68 seconds for $C=608,288,128,64,32$, respectively. This budget counts $C$ observations sampled at 0.24-second intervals.

The official evaluator checks final block positions, an open right gripper, and the button press, without checking the reference swap sequence. We group episodes by the number of reference swaps required by their initial permutation, determined before rollout. The five groups contain 24, 14, 28, 18, and 16 episodes requiring one through five swaps, respectively. Each group has a single initial permutation, so differences between groups also reflect initial geometry.

\paragraph{Stable-order reconstruction and strict success.}
We retain all official evaluator labels and add a stricter behavioral classification. Strict success also requires the observed stable-order path through its first target state to match the reference, and final placement to satisfy Table~\ref{tab:ranking-audit-thresholds}. An evaluator-only success passes the official checks but fails at least one of these additional criteria.

\begin{table}[!htbp]
\centering
\small
\caption{Criteria for stable block order and final placement. Stability criteria are shared with the extended-rollout sequence monitor. The final-placement cutoff is used only to classify strict behavioral success.}
\label{tab:ranking-audit-thresholds}
\begin{tabular}{@{}p{0.32\linewidth}p{0.62\linewidth}@{}}
\toprule
Quantity & Operational rule \\
\midrule
Stable order & Same left-to-right block order for eight consecutive action-level records satisfying the motion, row/height, and right-gripper criteria below \\
Motion & Maximum block displacement between consecutive sampled positions $\leq1.5$\,mm \\
Row and height & Each block within 6\,cm of $y=-0.10$\,m and $z\in[0.740,0.785]$\,m \\
Right gripper & Open-command value $>0.8$ \\
Final nominal-slot placement & For blocks $1,2,3$, $x^*=(0.04,0.16,0.28)$\,m and $y^*=-0.10$\,m; maximum absolute $x/y$ error $\leq4$\,cm at evaluator success \\
Sensitivity cutoffs & Recompute the final-placement classification at 2, 3, 4, and 5\,cm; keep sequence and evaluator labels fixed \\
\bottomrule
\end{tabular}
\end{table}

We reconstruct stable orders from the final position snapshot before each action's evaluator check. These action-level traces cover all 500 episodes. We visually reviewed flagged evaluator successes and two orderly control episodes.

\paragraph{Failures and transitions at the wrong phase.}

For each failed episode, we find the first action at which the target-geometry and open-gripper conditions hold simultaneously, if any. We then check for a button position below $-5$\,mm during that action or any later action. We flag a \emph{candidate button-only failure} if there is no such button event and, at that first action, the maximum absolute block-center $x/y$ error is at most 4\,cm and all block heights lie in $[0.740,0.785]$\,m. Separately, we count consecutive reference swaps from the initial state, stopping at the first mismatch.

A \emph{wrong-phase candidate} is the first stable-order transition that departs from the reference path but matches a transition used at another reference phase. It must also meet two conditions: each block is within 4\,cm in absolute $x/y$ error of the nominal slot assigned by its observed left-to-right order; and the button is below $-5$\,mm during the confirming action or a later action before the next stable transition. These candidate labels describe observed behavior, without identifying the policy's intended subgoal or the cause of an error.

\subsection{History Length, Block Placement, and Swap Order}
\label{app:ranking-context-results}
\label{app:ranking-audit-results}

All 24 one-swap episodes succeed at every tested history length (Table~\ref{tab:k-stratified}). Shorter histories produce more failures in episodes requiring multiple swaps. Table~\ref{tab:ranking-history-results} adds placement accuracy, progress along the reference sequence, and wrong-phase diagnostics for the same 100 evaluation episodes per $C$.

Placement error is the maximum absolute block-center $x/y$ error at first evaluator success. In the table, ``displaced'' means an error above 4\,cm, and ``signature'' denotes the wrong-phase flag. Panel (c) varies only the placement cutoff, keeping official labels fixed. Panel (d) counts path conditions among all 100 episodes per context length; panel (e) counts the listed conditions among evaluator failures. Panel (f) covers 266 cycles from the 94 strict successes at $C=608$. Each cycle runs from the first press associated with one stable order to the first press associated with the next.

\begin{table}[!htbp]
\centering
\small
\renewcommand{\arraystretch}{1.02}
\setlength{\tabcolsep}{4pt}
\caption{Placement, sequence, and failure diagnostics for the 100 evaluation episodes at each history length. Aggregate and swap-stratified outcomes appear in Table~\ref{tab:k-stratified}; success criteria are defined in Appendix~\ref{app:ranking-behavior-protocol}. Panels (a)--(e) describe history length, placement, and behavioral outcomes. Panel (f) reports swap-and-press cycle durations at $C=608$.}
\label{tab:ranking-history-results}
\label{fig:app-ranking-behavior}
\label{tab:ranking-behavior-audit}
\label{fig:app-ranking-audit}
\label{tab:ranking-failure-audit}
\begin{tabular*}{\linewidth}{@{\extracolsep{\fill}}p{0.48\linewidth}ccccc@{}}
\toprule
 & $C=608$ & $C=288$ & $C=128$ & $C=64$ & $C=32$ \\
\midrule
\multicolumn{6}{@{}l}{\textbf{(a) History windows and strict-success durations}} \\
Visible window (s) & 145.92 & 69.12 & 30.72 & 15.36 & 7.68 \\
Longest strict success (s) & 142.62 & 142.50 & 142.26 & 59.94 & 31.98 \\
\midrule
\multicolumn{6}{@{}l}{\textbf{(b) Block placement and swap order in successful episodes}} \\

Strict-success placement: minimum (cm) & 0.36 & 0.36 & 0.36 & 0.45 & 0.25 \\
Strict-success placement: median (cm) & 1.63 & 1.66 & 0.84 & 0.62 & 0.53 \\
Strict-success placement: maximum (cm) & 2.39 & 2.23 & 1.85 & 2.13 & 0.84 \\
Evaluator-only successes & 1 & 0 & 10 & 11 & 4 \\
\quad Expected path, displaced & 0 & 0 & 8 & 2 & 3 \\
\quad Displaced, unexpected path; signature absent & 0 & 0 & 0 & 9 & 1 \\
\quad Displaced, unexpected path; signature present & 0 & 0 & 2 & 0 & 0 \\
\quad Unexpected path, within 4\,cm & 1 & 0 & 0 & 0 & 0 \\
\midrule
\multicolumn{6}{@{}l}{\textbf{(c) Evaluator-only successes at each placement cutoff}} \\
2\,cm & 18 & 17 & 10 & 13 & 4 \\
3, 4, and 5\,cm (each) & 1 & 0 & 10 & 11 & 4 \\
\midrule
\multicolumn{6}{@{}l}{\textbf{(d) Signatures across all 100 episodes per history}} \\
Unexpected stable-order path & 4 & 4 & 29 & 40 & 9 \\
Wrong-phase candidate & 1 & 1 & 6 & 1 & 0 \\
\midrule
\multicolumn{6}{@{}l}{\textbf{(e) Block placement, button presses, and swap order in failed episodes}} \\
Evaluator failures & 5 & 8 & 41 & 62 & 72 \\
Target geometry and open gripper never co-occurred & 4 & 8 & 40 & 62 & 71 \\
Target geometry with open gripper reached; button never below $-5$\,mm from that action onward & 1 & 0 & 1 & 0 & 1 \\
Candidate button-only failure & 0 & 0 & 0 & 0 & 0 \\
Complete expected path, evaluator failure & 0 & 0 & 1 & 0 & 1 \\
Exactly one reference swap matched, evaluator failure & 0 & 0 & 0 & 53 & 61 \\
\bottomrule
\end{tabular*}

\medskip
\begin{tabular*}{\linewidth}{@{\extracolsep{\fill}}lrrrr@{}}
\toprule
\multicolumn{5}{@{}l}{\textbf{(f) Swap-and-press cycle duration, $C=608$}} \\
Unit & Minimum & Median & Mean & Maximum \\
\midrule
World-token equivalents & 112.50 & 116.50 & 116.07 & 118.75 \\
Simulated seconds & 27.00 & 27.96 & 27.86 & 28.50 \\
\bottomrule
\end{tabular*}
\end{table}

\subsection{Extended Execution beyond Demonstrations}
\label{app:ranking-stress-protocol}
\label{app:ranking-stress-results}

The exploratory stress test disables termination at first evaluator success while keeping the checkpoint, stochastic sampler, four-action execution interval, and $C=608$ window fixed. We record one trajectory for each of the nine seeds used for the exploratory stress test in Table~\ref{tab:stress}, covering all five non-target initial permutations. Rollouts have no fixed action-step limit; they end under the stopping conditions below.

The monitor identifies stable block orders using Table~\ref{tab:ranking-audit-thresholds} and checks transitions against the repeating sequence $\mathrm{swap}(M,R)$, $\mathrm{swap}(L,R)$, $\mathrm{swap}(L,M)$. A swap counts as correct when the next stable order matches the prescribed positional swap; a subsequent button press is not separately required for every counted swap. We record first evaluator success independently, so the target arrangement may be reached before the success event. The history window starts sliding at 145.92 seconds.

Each trajectory ends at the first confirmed stable-order deviation or after 1,500 actions (90 seconds) without a new confirmed reference transition. Event times equal action indices multiplied by 0.06 seconds. The test measures how long ordered behavior continues after removing the demonstrations' stopping rule.

The nine trajectories complete a median of eight confirmed swaps and a maximum of 31. Five continue the reference sequence after the $C=608$ window starts sliding. Ordered behavior can therefore persist beyond the initial episode prefix, although its duration varies substantially across seeds.

\section{Relationship between Offline Action RMSE and Closed-Loop Success}
\label{app:rmse-sr}

\subsection{Agreement and Differences between RMSE and Success Rate}
\label{app:rmse-sr-analysis}

\begin{figure}[!t]
\centering
\includegraphics[width=0.9\linewidth]{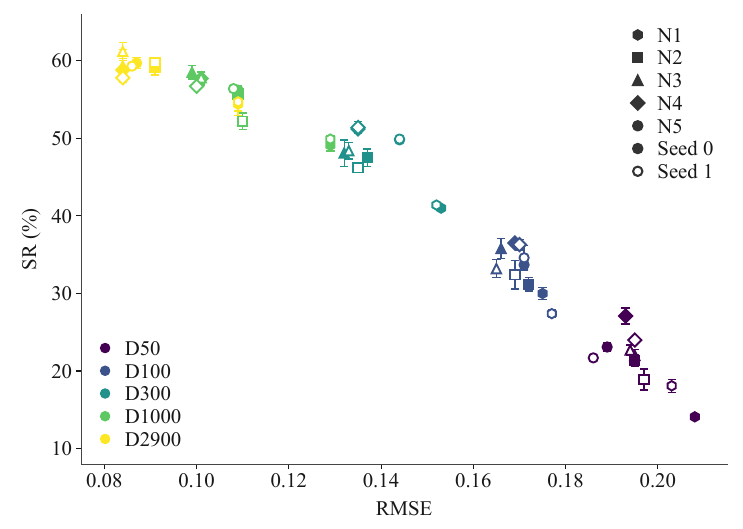}
\caption{Holdout stochastic RMSE versus closed-loop SR for all 50 trained policies. The $x$-axis is task-averaged RMSE over all 12 action dimensions and all ten steps in each sampled action chunk; the $y$-axis is mean SR over three complete evaluations with a ten-step context. Color, marker, and fill encode demonstrations per task $D$, capacity, and training seed. Error bars are fixed-seed repeatability sample standard deviations (Appendix~\ref{app:robocasa-evaluation}). Descriptive Spearman correlation is $\rho = -0.980$, summarizing the sweep-level trend rather than ranking nearby checkpoints.}
\label{fig:rmse-sr}
\end{figure}

Across the RoboCasa sweep, lower expert-action RMSE generally accompanies higher closed-loop success. All 40 comparisons between adjacent data sizes at fixed model capacity and training seed show lower full-chunk stochastic holdout RMSE and higher mean SR with a ten-step context. The same holds for all 10 same-data comparisons between the 44.3M- and 218.8M-parameter models. Figure~\ref{fig:rmse-sr} shows this association across all 50 policies.

This overall trend does not reliably rank nearby checkpoints. In the separately trained context comparison, increasing the training and evaluation context from five to ten steps lowers both RMSE and SR for both training seeds. Mean RMSE falls from 0.138384 to 0.132404, while mean SR falls from 50.74\% to 48.25\%. Across tasks, RMSE improves on 20/23; SR decreases on 15, is unchanged on one, and increases on seven. Thus, lower RMSE under these context-specific evaluation protocols does not imply higher closed-loop SR.

From $D=1000$ to $D=2900$, all 10 matched training-seed/model-capacity comparisons improve in both metrics. Using full-precision final measurements, the mean relative RMSE reduction is 17.10\% with a 1.76-percentage-point sample standard deviation. SR increases by 3.39 percentage points on average, with a 2.12-point sample standard deviation. RMSE also improves in all 230 task-level comparisons across ten seed/capacity pairs and 23 tasks. Task-level SR can increase or decrease, as the 218.8M results show (Table~\ref{tab:taskwise-scaling-complete}). Relative RMSE reductions and absolute SR changes use different units and aggregation, so their numerical magnitudes are not directly comparable.

Holdout RMSE measures sampled action-chunk similarity on fixed expert observations and histories. Closed-loop SR measures task completion along trajectories generated by the policy and environment. Two differences help explain why the metrics can disagree, although the present experiments do not isolate the cause of that disagreement.

First, expert demonstrations have limited state coverage. A small policy error can change later images, proprioception, object poses, and contacts. Better predictions on recorded trajectories therefore need not improve recovery after a deviation, as in the covariate-shift problem of behavior cloning \citep{ross2011dagger}. A recorded action may also be only one of several valid approaches, contact patterns, timings, or recovery paths. RoboCasa uses 50 human demonstrations per task to guide MimicGen in generating roughly 3,000 trajectories \citep{nasiriany2024robocasa}. Additional generated trajectories provide denser coverage of this expert distribution, but not necessarily recovery states induced by the policy. Identifying this mechanism would require failure-state data, recovery demonstrations, or on-policy data collection.

Second, success depends on receding-horizon execution, controllers, contact dynamics, termination, and the evaluator's discrete success predicate. In the Blocks Ranking diagnostic (Appendix~\ref{app:ranking-loss-results}), the standard-loss record associates a millimeter-scale difference in final button depression with a different binary outcome, while sorting remains largely intact. This example shows that SR reflects the full execution and evaluation process, rather than model outputs alone.

These observations concern the behavior-cloning policies trained on the benchmark data studied here. RMSE provides a dense diagnostic on fixed expert-trajectory crops and has a smaller relative range across training seeds than SR in this grid. Repeated rollout evaluation remains necessary to assess closed-loop capability. Large, consistent changes in both metrics provide complementary evidence. When the metrics disagree, training-state coverage, recovery behavior, controller effects, and evaluation rules deserve closer examination; small metric differences alone offer a weak basis for ranking models.

\subsection{Definitions of Offline RMSE and Closed-Loop Success}
\label{app:formal}

For the main RoboCasa sweep, let $p_{E}^{\mathrm{ho}}(\tau)$ denote the finite holdout expert-trajectory distribution. RMSE uses the fixed crops described in Appendix~\ref{app:robocasa-evaluation}: eight ten-timestep crops from each task's 100 holdout demonstrations, rather than all trajectory windows. For a crop starting at policy index $s$, query $t$ receives only $o_{s:t}$. Let $j_t$ be its stored control-frame index and $A_t=(a_{j_t},\ldots,a_{j_t+H-1})$ the target of consecutive low-level commands. The indexed collection $\mathcal{Q}_m$ contains all pairs $(o_{s:t},A_t)$ with a complete $H$-step target for task $m$. Then
\begin{equation}
\mathrm{RMSE}_{m}
=
\Biggl[
\frac{1}{\lvert \mathcal{Q}_m \rvert\, H d_{a}}
\sum_{(o_{s:t},A_t) \in \mathcal{Q}_m}
\left\| \hat{A}(o_{s:t}) - A_t \right\|_{2}^{2}
\Biggr]^{1/2},
\qquad
\mathrm{RMSE}_{E}
=
\frac{1}{M}
\sum_{m=1}^{M}
\mathrm{RMSE}_{m},
\label{eq:rmse-e}
\end{equation}
with $M = 23$ tasks and one stochastic ten-step, 12-dimensional chunk $\hat{A}(o_{s:t})$ per query under \texttt{eval\_seed=0}. Targets and predictions are in the original, unnormalized controller-command space. We average squared error over all scalar action elements for each task, take the square root, and then average equally across tasks. The deterministic variant uses the same initial $x_T$ and the posterior-mean reverse transitions in Appendix~\ref{app:robocasa-evaluation}. RMSE measures imitation on holdout expert observations and histories; the training objective is diffusion noise-prediction loss.

For closed-loop evaluation, the initial-state distribution $\rho_{0}$, policy, and environment/evaluation protocol $\mathcal{M}$ induce the trajectory distribution
\begin{equation}
p_{\pi,\mathcal{M}}(\tau \mid \rho_{0}).
\label{eq:policy-traj}
\end{equation}
Let $E_{\mathcal{M}}(\tau)$ be the evaluator's binary outcome for trajectory $\tau$. The set of successful trajectories is
\begin{equation}
T^{\mathrm{succ}}_{\mathcal{M}} = \{\tau \mid E_{\mathcal{M}}(\tau) = 1\}.
\label{eq:succ-set}
\end{equation}
Then
\begin{equation}
\mathrm{SR}(\pi; \mathcal{M}, \rho_{0})
=
\Pr_{\tau \sim p_{\pi,\mathcal{M}}(\cdot \mid \rho_{0})}
\bigl[\tau \in T^{\mathrm{succ}}_{\mathcal{M}}\bigr].
\label{eq:sr-def}
\end{equation}
Successful trajectories may follow expert paths, alternative solutions, or recovery paths. They may also be very short if the success criterion permits this. Offline RMSE on $p_{E}^{\mathrm{ho}}$ and closed-loop SR on $p_{\pi,\mathcal{M}}$ therefore evaluate different distributions and outcomes, as discussed in Appendix~\ref{app:rmse-sr-analysis}.

\subsection{Variation across Training Seeds and Repeated Evaluations}
\label{app:stability}

We summarize variation in metrics with different units using the relative range
\begin{equation}
R_{\mathrm{rel}}(x)
=
\frac{\max_{i} x_{i} - \min_{i} x_{i}}{\bar{x}}.
\label{eq:rel-range}
\end{equation}
For each of the 50 policies, we use three evaluations with a ten-step context, totaling 150 executions of 1,150 episodes. Each checkpoint's repeats use the same episodes, environment seeds, and rollout seed, as specified in Appendix~\ref{app:robocasa-evaluation}. They measure fixed-seed repeatability rather than variation across independently sampled rollout seeds. SR relative range has median 3.40\%, third quartile 5.09\%, and maximum 14.24\%; the corresponding absolute ranges are 1.39, 2.09, and 4.09 percentage points. Many small differences between nearby high-performing configurations fall within this range of variation.

Across training seeds, we apply the same formula to the two results for each of the 25 $(D,\text{capacity})$ pairs. RMSE relative range has median 0.76\%, third quartile 1.14\%, and maximum 2.26\%; the SR values are 1.74\%, 6.35\%, and 24.78\%, respectively. These summaries compare different trained policies, so they do not isolate the contributions of individual sources of variation. Together, the repeatability and cross-seed results support using RMSE and repeated SR as complementary evidence for broad trends, rather than selecting models by small local differences.

\section{Extended Related Work on Robotic Sequence Modeling}
\label{sec:related}

\subsection{Sequence organization in robot policies}
\label{sec:related-seq}

Sequence modeling is widely used in robot learning and generalist agents. Decision Transformer arranges returns, states, and actions in a causal sequence. Gato serializes data across tasks, modalities, and embodiments into one token stream, using a shared Transformer to generate text, actions, and other outputs from context \citep{chen2021decision, reed2022gato}. These works formulate decision making as sequence modeling while leaving open the choice of a basic unit for temporal context.

Robot Transformers use different sequence units. RT-1 compresses each visual frame into several tokens with TokenLearner. ICRT combines vision and proprioception into a state token and interleaves it with action features in a causal sensorimotor sequence. GR-1 places language, image sequences, robot state, and future-image prediction in one GPT-style sequence \citep{brohan2023rt1, fu2024icrt, wu2024gr1}. All three track physical time, but their top-level positions can differ in modality, input source, or generation role.

Another design choice separates history modeling from action generation. Diffusion Policy models continuous, multimodal action sequences with conditional diffusion; ACT uses a conditional VAE to generate action chunks; Octo summarizes context with learned readout tokens and uses a lightweight diffusion head; and $\pi_0$ adds a separately parameterized flow-matching action expert \citep{chi2023diffusion, zhao2023act, octo2024, black2025pi0}. These methods establish precedents for separate action-generation modules. Here, we focus on how temporal context is organized and accessed before action generation.

Compact timestep representations also have direct precedents. The BC-Transformer baseline in RoboCasa encodes each observation timestep, applies non-causal self-attention within a fixed ten-observation window, and lets every window position emit an action prediction through a Gaussian mixture head \citep{nasiriany2024robocasa, arise2024robomimic}. HAMLET introduces moment tokens that compress each timestep of a pretrained VLA and aggregates them with a lightweight memory module, and Chronos represents each control step with one state-representative token propagated through a selective state-space model \citep{koo2026hamlet, zhou2026chronos}.

WorldToken contributes one observation-derived token per policy timestep to a causal sequence. We study data and model scaling, vary the number of tokens per timestep across data scales, and test the effect of visible history. Token-count comparisons keep tokens grouped by timestep and preserve causal order.

\subsection{Temporal context, memory, and partial observability}
\label{sec:related-memory}

History-conditioned control predates modern Transformers: recurrent imitation policies such as BC-RNN maintain a hidden state across timesteps, and Hiveformer jointly models language, multiview observations, and full observation/action history for multitask manipulation \citep{mandlekar2022whatmatters, guhur2023hiveformer}.

History availability, necessity, and utilization are distinct. Availability concerns whether the architecture can receive past observations; necessity concerns whether the task is partially observable without them; utilization concerns whether the trained policy's behavior depends on history. A large input window establishes only availability. Truncating or perturbing history while fixing policy weights primarily tests utilization \citep{chen2026rmbench, shi2026memoryvla, zhou2026chronos}.

Existing designs retain history in three broad ways. Recurrent state compression summarizes the past in a continually updated hidden state, as in BC-RNN and Chronos's selective state-space model. Explicit sequence context keeps past interactions available for attention. Examples include Hiveformer's early full-history Transformer, HALO's VQA-supervised relevance and sparse top-$K$ attention to reduce spurious historical correlations, and PRISM's gated attention and hierarchical compression for minute-scale visuomotor memory. Controlled Diffusion Policy studies also show that context-length gains depend on conditioning, denoising architecture, variable-length training, and data conditions \citep{shah2026aprism, shah2026bhalo, agarwal2026longcontext}.

Dedicated memory subsystems define write, update, or retrieval rules. RMBench proposes Mem-0 as an explicit-memory reference, MemoryVLA maintains a Perceptual-Cognitive Memory Bank, and RoboMME evaluates memory-augmented VLA integration across several dimensions \citep{chen2026rmbench, shi2026memoryvla, dai2026robomme}. Across these approaches, longer context does not by itself ensure effective memory.

WorldToken uses explicit sequence context, without a separate memory bank or hand-designed rules for retrieval or updates. We measure history utilization through controlled context interventions on fixed trained checkpoints.

Memory design also affects computation. RoboMME compares memory budgets through performance--compute curves, while HAMLET and MEM compress observation history at different granularities \citep{dai2026robomme, koo2026hamlet, torne2026mem}. Appendix~\ref{app:history-growth} examines how the tokens contributed by each timestep affect computational growth with cached history.

\subsection{Latent representations and predictive world models}
\label{sec:related-world}

Compressing observations into latent state and predicting the environment there is a classic world-model strategy. PlaNet learns latent dynamics for planning, Dreamer learns through imagined trajectories, and JEPA-style methods predict abstract representations rather than reconstructing all inputs \citep{hafner2019planet, hafner2020dreamer, hafner2025dreamerv3, lecun2022path, assran2023jepa}. V-JEPA 2-AC adds an action-conditioned latent world model for robot planning, while GR-1 and ACT-JEPA combine behavior learning with future image or latent prediction \citep{assran2025vjepa2, wu2024gr1, vujinovic2026actjepa}.

WorldToken also forms a compact representation before decision making. Its tokens are policy representations aligned with physical time, rather than predictive world-model states. They are not defined through future-state prediction, though such objectives could be added.

\subsection{Scaling in robot learning}
\label{sec:related-scaling}

Scaling studies examine the effects of data and model size. RT-1 varies data size, model capacity, and diversity; Open X-Embodiment/RT-X studies cross-robot data aggregation; and OpenVLA combines vision--language pretraining with large robot datasets \citep{brohan2023rt1, openx2024, kim2025openvla}. Other studies show that demonstration count and the diversity of training environments and objects affect policy generalization \citep{lin2025datascaling}. ScaleDP and broader imitation-learning studies show that the benefits of model capacity depend on architecture, tokenization, data, and optimization \citep{zhu2024scaledp, pearce2025scaling}.

Our sweep evaluates all combinations of five dataset sizes and five model capacities within one WorldToken family, with visual and policy modules trained from scratch. Given the architecture dependence found in prior work, our conclusions characterize this family within the tested range.

\clearpage
\section{Qualitative RoboCasa Rollouts}
\label{app:robocasa-qualitative}

\begin{figure}[!htbp]
  \centering
  \includegraphics[page=1,width=\textwidth]{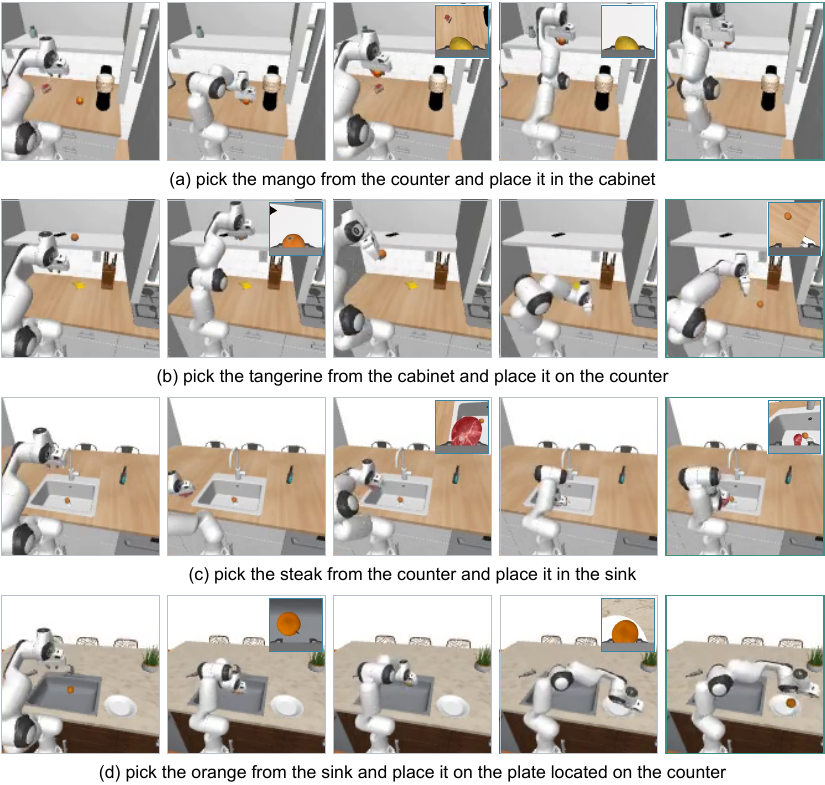}
  \caption{Pick and place (I): cabinet and sink transfers. The robot moves objects between the counter and a cabinet (a,b), and between the counter and a sink (c,d).}
  \label{fig:robocasa-appendix-1}
\end{figure}

\begin{figure}[!htbp]
  \centering
  \includegraphics[page=2,width=\textwidth]{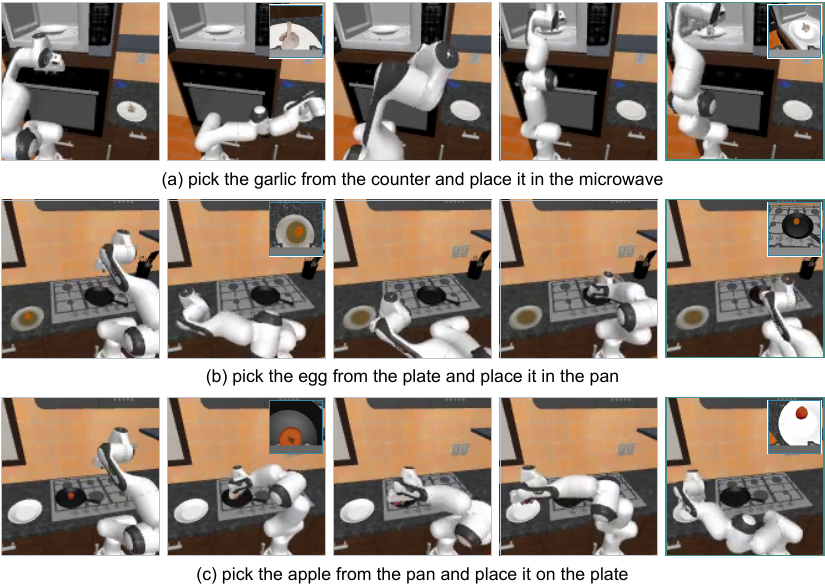}
  \caption{Pick and place (II): microwave and stove transfers. The robot places an object in the microwave (a), moves an egg from a plate to a pan (b), and moves an apple from a pan to a plate (c).}
  \label{fig:robocasa-appendix-2}
\end{figure}

\begin{figure}[!htbp]
  \centering
  \includegraphics[page=3,width=\textwidth]{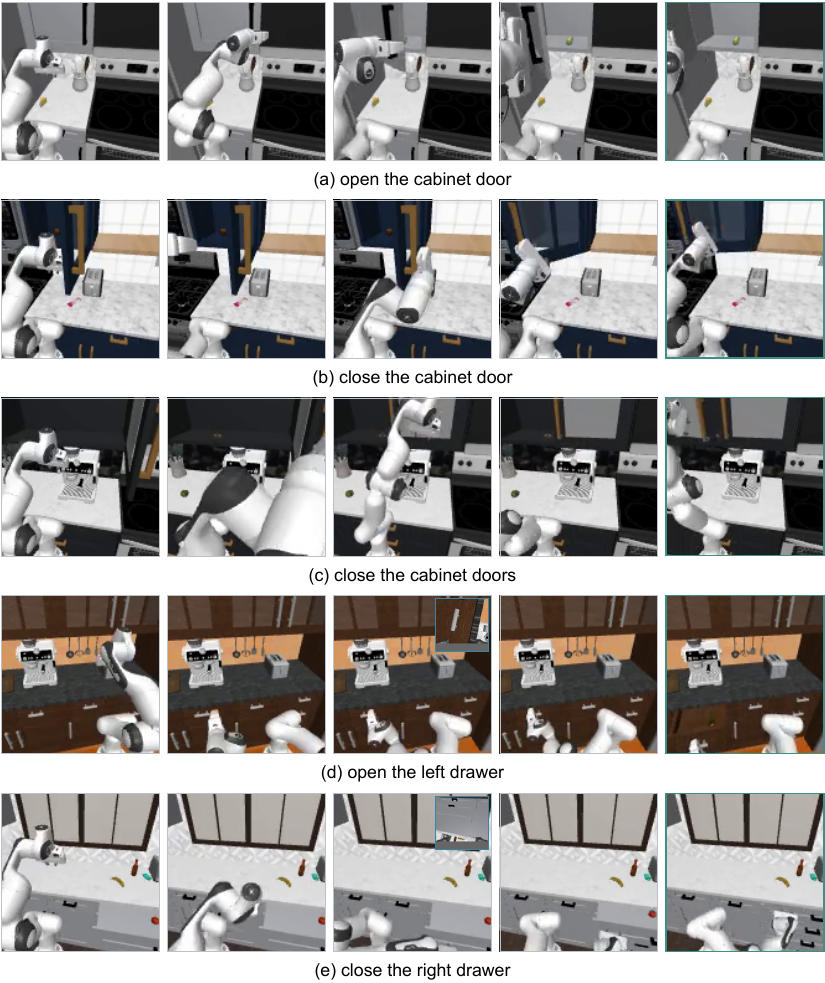}
  \caption{Opening and closing doors and drawers. Panels (a--c) show opening a cabinet door, closing a cabinet door, and closing both cabinet doors, respectively. Panels (d,e) show opening and closing a drawer.}
  \label{fig:robocasa-appendix-3}
\end{figure}

\begin{figure}[!htbp]
  \centering
  \includegraphics[page=4,width=\textwidth]{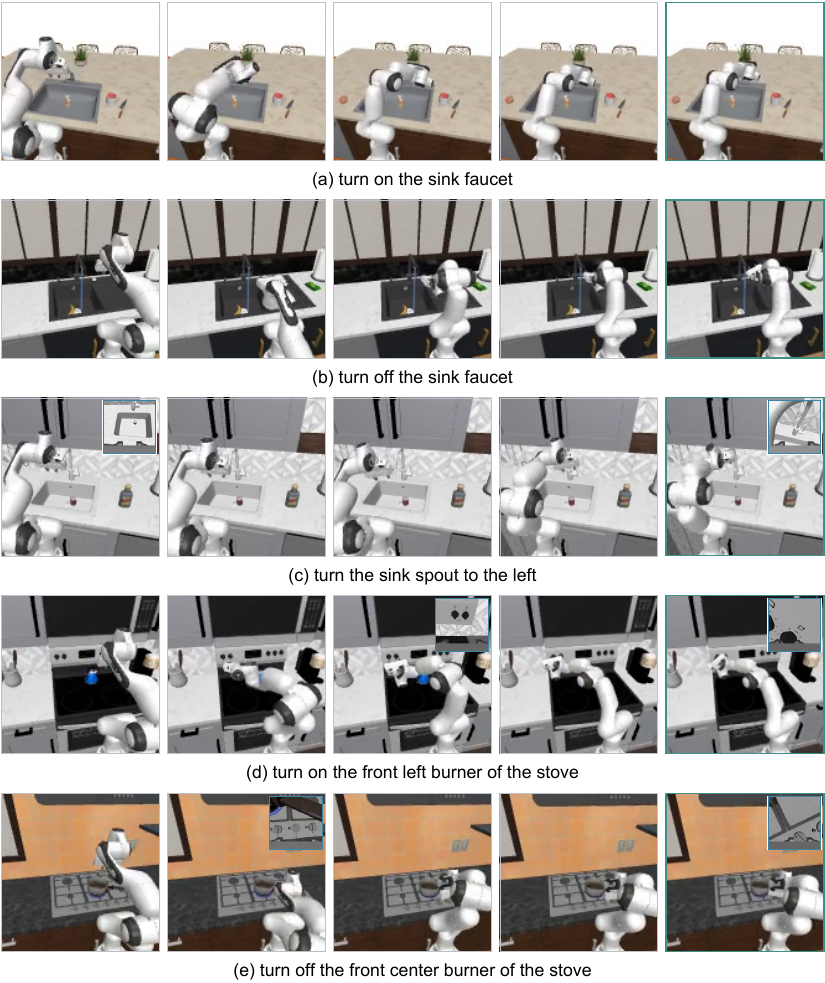}
  \caption{Sink and stove controls. Panels (a--c) show faucet and spout control in the Turning levers family. Panels (d,e) show switching stove burners on and off in the Twisting knobs family.}
  \label{fig:robocasa-appendix-4}
\end{figure}

\begin{figure}[!htbp]
  \centering
  \includegraphics[page=5,width=\textwidth]{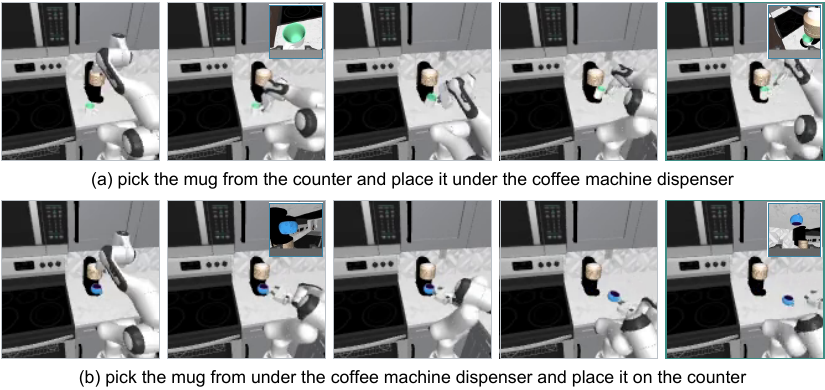}
  \caption{Mug placement and retrieval at the coffee machine. The two tasks belong to the official Insertion family: placing a mug under the dispenser (a) and moving it from the dispenser to the counter (b).}
  \label{fig:robocasa-appendix-5}
\end{figure}

\begin{figure}[!htbp]
  \centering
  \includegraphics[page=6,width=\textwidth]{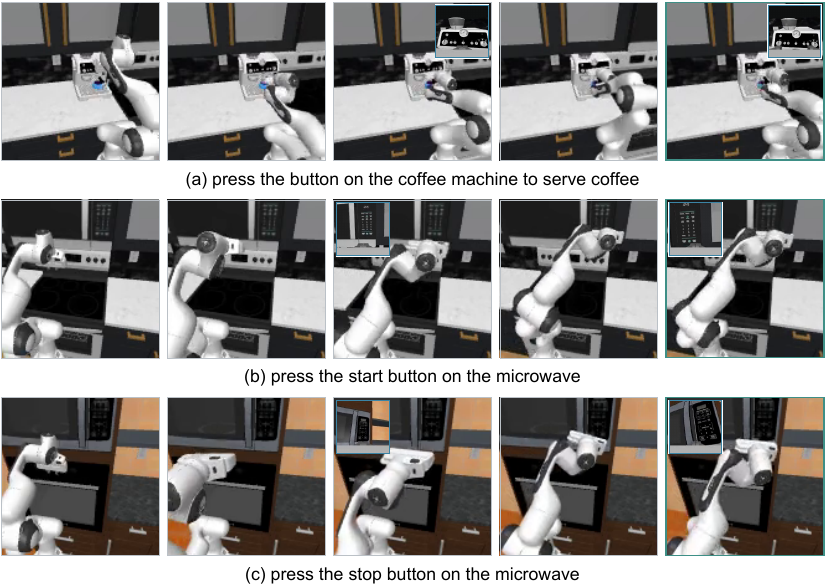}
  \caption{Pressing appliance buttons. The robot presses the button on the coffee machine (a), the microwave start button (b), and the microwave stop button (c).}
  \label{fig:robocasa-appendix-6}
\end{figure}

Figures~\ref{fig:robocasa-appendix-1}--\ref{fig:robocasa-appendix-6}
show 22 successful WorldToken rollouts on RoboCasa, with one trajectory per
task--scene configuration. The examples are organized into seven official
atomic-task families. Episodes are selected across training runs, checkpoints, and
evaluated history lengths, with success verified from evaluation logs.

Each row presents five frames together with the complete original language
instruction. Frames proceed from left to right at nonuniform intervals.
Blue-bordered insets show wrist-camera observations from the same timestep
as the corresponding main frame, and green borders mark the last displayed
frame of each trajectory.

\clearpage

\clearpage

\clearpage

\clearpage

\clearpage

\clearpage

\end{document}